\documentclass[11pt]{article}
\usepackage[usenames,dvipsnames]{color}
\usepackage{amsmath}
\usepackage{textcomp}
\usepackage{gensymb}
\usepackage{amsmath,amssymb,amsfonts,amsthm}
\usepackage{array}
\usepackage{cite}
\usepackage{lscape}
\usepackage{color}
\usepackage{epsfig}
\usepackage{graphics}
\usepackage{graphicx}
\usepackage{xcolor}
\usepackage{latexsym}
\usepackage{longtable}
\usepackage{upref}
\usepackage{bm}
\usepackage{algorithm}
\usepackage{algpseudocode}
\usepackage{enumerate}
\usepackage[mathscr]{eucal}
\usepackage{rotate, graphics,epsfig}
\usepackage{gensymb}
\usepackage[left=2.1cm,right=2.1cm,top=1.8cm,bottom=1.8cm]{geometry}
\usepackage{float}
\usepackage{subfigure}
\usepackage{subfig}
\usepackage{caption}
\usepackage{url}
\usepackage{fancyhdr}
\usepackage{float}
\usepackage{amsmath}
\usepackage[T1]{fontenc}
\usepackage[utf8]{inputenc}
\usepackage{authblk}
\usepackage{blindtext}
\usepackage{dutchcal}

\usepackage{amssymb}
\usepackage{mathtools}
\usepackage[toc,title,page]{appendix}

\newtheorem{theorem}{Theorem}[section]

\theoremstyle{definition}

\newtheorem{remark}{Remark}

\DeclareMathOperator*{\argmin}{argmin}
\newcommand{\ScrS}{\mathscr{S}}
\begin{document}
	\renewcommand{\thefootnote} {\fnsymbol{footnote}}
	\setcounter{page}{1}
	\date{}
	\title{
		\textbf{Discovery of interaction and diffusion kernels in particle-to-mean-field multi-agent systems}}
	
	\author[1]{Giacomo Albi}
	\author[2]{Alessandro Alla}
	\author[1]{Elisa Calzola}
	
	\affil[1]{Department of Computer Science, University of Verona, Verona, Italy (giacomo.albi@univr.it)}
	\affil[2]{Department of Mathematics, La Sapienza University of Rome, Rome, Italy (alessandro.alla@uniroma1.it)}
	\affil[3]{Department of Mathematics and Computer Science, University of Ferrara, Ferrara, Italy (elisa.calzola@unife.it)}

	\maketitle
	\begin{abstract}
		We propose a data-driven framework to learn interaction kernels in stochastic multi-agent systems. 
		Our approach aims at identifying the functional form of nonlocal interaction and diffusion terms directly from trajectory data, without any a priori knowledge of the underlying interaction structure. Starting from a discrete stochastic binary-interaction model, we formulate the inverse problem as a sequence of sparse regression tasks in structured finite-dimensional spaces spanned by compactly supported basis functions, such as piecewise linear polynomials. In particular, we assume that pairwise interactions between agents are not directly observed and that only limited trajectory data are available. To address these challenges, we propose two complementary identification strategies. The first based on random-batch sampling, which compensates for latent interactions while preserving the statistical structure of the full dynamics in expectation. The second based on a mean-field approximation, where the empirical particle density reconstructed from the data defines a continuous nonlocal regression problem. Numerical experiments demonstrate the effectiveness and robustness of the proposed framework, showing accurate reconstruction of both interaction and diffusion kernels even from partially observed. The method is validated on benchmark models, including bounded-confidence and attraction-repulsion dynamics, where the two proposed strategies achieve comparable levels of accuracy.
			\end{abstract}
	\begin{center}
	\end{center}
		\begin{center}
	\end{center}
\section{Introduction}
The increasing availability of high-resolution data has sti\-mu\-lated a major research effort toward the discovery of differential equations directly from observations. In this setting, one typically assumes little or no a priori knowledge of the governing laws, and the objective is to reconstruct interpretable mathematical models from trajectory data. Among the various approaches in this direction we mention the Sparse Identification of Nonlinear Dynamics (SINDy) framework \cite{BPK16,RBPK16,RABK19}, which promotes sparse representations over candidate libraries for interpretable and efficient equation recovery, with extensions improving finite-data accuracy for stochastic~\cite{WM24} and partial differential equations~\cite{messenger2021weak}. Neural-network-based methodologies, such as Physics-Informed Neural Networks (PINNs)~\cite{Karniadakis2021,RPK19}, provide an alternative paradigm in which physical structure is embedded directly into learning architectures. In contrast, the present work formulates a regression problem to recover the governing mechanisms from data within structured finite-dimensional function spaces, yielding interpretable analytical approximations of the underling system. In particular, we focus on the discovery of interaction laws within  multiscale agent systems similarly to recent approaches proposed in~\cite{bongini2017inferring,MillerTangZhongMaggioni2023}, and motivated by data-driven models in socio-economic settings \cite{albi2025data, bonafini2023data}.

Large systems of interacting agents constitute a fundamental modeling paradigm for describing hierarchical populations across a wide range of applications, including biological systems~\cite{BTDAC97,CKFL05}, epidemiological dynamics~\cite{DFP26,DPTZ21} and collective decision-making processes such as opinion formation~\cite{APZ14,during2009boltzmann,toscani2006kinetic}. In this setting, stochasticity and diffusion mechanisms are essential components of many interacting agent models, significantly affecting the emergence and large-scale organization of collective dynamics~\cite{erban16,bicego2025computation}. Kinetic and mean-field descriptions of collective dynamics provide a multiscale modeling framework to connect microscopic interactions to macroscopic evolution laws \cite{toscani2006kinetic,during2009boltzmann}, where Boltzmann-type and Fokker--Planck equations are derived from binary interaction rules, see for example~\cite{CFRT10,CCH14, pareschi2013interacting}. This hierarchy captures emergent collective behavior across particle, kinetic, and hydrodynamic levels. Extensions incorporating stochasticity, uncertainty, and heterogeneous interactions have been studied in several directions, e.g in \cite{tosin2018boltzmann,pareschi2022mean,carrillo2019particle}. 

Within this broad context, a rapidly growing research direction addresses the inverse problem of learning nonlocal interaction kernels from trajectory data, initiated in \cite{bongini2017inferring} and further developed into nonparametric learning frameworks with statistical guarantees in \cite{lu2019nonparametric,LMT21,LMT22,MillerTangZhongMaggioni2023} and numerical kernel-based reconstruction approaches \cite{fiedler2025recent}. These works establish consistency results and learning theory for the recovery of radial interaction kernels in deterministic and stochastic multi-agent systems, typically assuming access to multiple independent trajectories generated from different initial conditions. Related mean-field formulations with PDE-level stability estimates were proposed also in \cite{CEMT25,lang2022learning}.
Despite significant progress, most kernel-learning approaches assume full observability, focus primarily on drift identification, and rely on multiple trajectories for statistical stability. However, in practice interactions may be latent, dynamics inherently stochastic, and data limited, requiring for robust identification strategies with respect to partial observability and scarcity of data.

In this work, we address these challenges by developing a regression framework for the joint identification of nonlocal drift and diffusion interaction kernels in stochastic multi-agent systems with unobserved interaction structure. Inspired by particle-based simulation methods such as \cite{nanbu1980direct, albi2013binary, jin2020random}, we adopt a discrete stochastic binary-interaction model to describe the underlying agent dynamics. This modeling choice provides a natural bridge between microscopic interactions and macroscopic collective behavior, ensuring multiscale consistency \cite{borghi2025wasserstein, huang2025mean}.  Hence, the inverse problem is formulated as a regression task in finite-dimensional linear spaces spanned by compactly supported piecewise linear basis functions, where two main challenges arise due to limited availability of independent trajectories and the unobservability of pairwise interactions. To cope with these issues, we propose two different strategies. First, we use a random-batch regression approach to manage latent interactions via randomly sampled subsets, reducing computational complexity while preserving the statistical structure in expectation. Furthermore, we establish a connection with a mean-field formulation in which the empirical density is reconstructed from the data, and defining a continuous nonlocal regression problem. This macroscopic perspective is subsequently integrated into the reconstruction process, enhancing the robustness and stability of the overall identification framework. We also provide a priori error estimates that account for the fact that the reconstructed trajectories do not use information on the interaction between particles. Numerical experiments confirm that this approach accurately reconstructs finite-dimensional approximations of both drift and diffusion interaction kernels in noisy and data-limited regimes, demonstrating its effectiveness and scalability for inverse problems in stochastic interacting particle systems.

The manuscript is organized as follows. 
In Section~\ref{sec:int}, we introduce the modeling hierarchy and the binary interaction dynamics underlying the learning problem. 
Section~\ref{sec:discovery} presents the identification of interaction and diffusion kernels under full observability of the pairwise interactions. 
Sections~\ref{sec:batch} and~\ref{sec:mfadensity} address the case of partial observability via random-batch and mean-field approaches, respectively. 
Section~\ref{sec:stima} provides an a priori error estimate for the reconstructed trajectories.
Section~\ref{sec:num} presents numerical experiments assessing the performance of the proposed methods, and final conclusions are drawn in Section~\ref{sec:conclusions}. 
\section{Model hierarchy for pairwise interacting agent systems}\label{sec:int}

We consider a system of $N$ agents characterized at discrete time $t_n$ by
features $\{x_i^n\}_{i=1}^N$, with $x_i^n \in \mathbb{R}^d$.
Interactions occur pairwise, over a time step $[t_n,t_n+\Delta t]$
each agent $i \in \{1,\dots,N\}$ interacts with a selected agent
$j(i)$, and the post-interaction states are updated according to
\begin{equation}\label{eq:binary_disc}
	\begin{aligned}
		x_i^{n+1} &= x_i^n
		+ \Delta t\, P(x_i^n,x_{j(i)}^n)(x_{j(i)}^n - x_i^n)
		+ \sqrt{\Delta t}\, D(x_i^n,x_{j(i)}^n)\,\xi_{ij(i)}^n, \\
		x_{j(i)}^{n+1} &= x_{j(i)}^n
		+ \Delta t\, P(x_{j(i)}^n,x_i^n)(x_i^n - x_{j(i)}^n)
		+ \sqrt{\Delta t}\, D(x_{j(i)}^n,x_i^n)\,\xi_{j(i)i}^n,
	\end{aligned}
\end{equation}
where $P,D:\mathbb{R}^d\times\mathbb{R}^d \to \mathbb{R}$ weight the
compromise and stochastic fluctuation mechanisms.
The random vectors $\xi_{ij}^n$ are i.i.d.\ with zero mean and covariance
$\mathbb{E}[\xi_{ij}^n(\xi_{ij}^n)^\top]=\mathsf{I}_d$.

The microscopic rule \eqref{eq:binary_disc} defines a discrete-time
stochastic process on $\mathbb{R}^{dN}$. Randomness enters in the microscopic dynamics through two mechanisms:
the stochastic fluctuations $\xi_{ij}^n$ acting at each binary
interaction, and the random selection of partners $j(i)$ at every time
step. As a result, the system evolves through stochastic pairwise
interactions rather than deterministic averaged forces.  
Such binary random interactions constitute the elementary mechanism
underlying random-batch approximations of mean-field particle
systems and, in the large population limit, admit an equivalent
kinetic representation in terms of the one particle distribution
function. The following subsections detail these connections.

\subsection{Mean-field dynamics via random-batch approximation}

The binary scheme \eqref{eq:binary_disc} can be interpreted as the
building block of a stochastic {random-batch} approximation of a
fully interacting system.
At time step $n$, each agent $i$ interacts with a random subset
$J_i^n \subset \{1,\dots,N\}\setminus\{i\}$ of size $N_p \ll N$,
leading to
\begin{equation}\label{eq:rbm_disc}
	x_i^{n+1}
	= x_i^n
	+ \frac{\Delta t}{N_p}\sum_{j \in J_i^n}
	P(x_i^n,x_j^n)(x_j^n - x_i^n)
	+ \frac{\sqrt{\Delta t}}{\sqrt{N_p}}\sum_{j \in J_i^n}
	D(x_i^n,x_j^n)\,\xi_{ij}^n .
\end{equation}
Under the assumption that, for each fixed $i$ and $n$, the set
$J_i^n$ is sampled uniformly among all subsets of
$\{1,\dots,N\}\setminus\{i\}$ of cardinality $N_p$,
independently of the particle configuration and of previous time steps,
the random-batch approximation provides an unbiased estimator of the
full interaction operators. Hence, passing formally to the limit $\Delta t \to 0$ and setting $N_p=N$ yield the
continuous-time interacting particle system
\begin{equation}\label{eq:meanfield_sde}
	dx_i
	=
	\frac{1}{N}\sum_{j=1}^N
	P(x_i,x_j)(x_j - x_i)\,dt
	+
	\frac{1}{\sqrt{N}}\sum_{j=1}^N
	D(x_i,x_j)\,dB_{ij},
\end{equation}
where $(B_{ij})_{i,j}$ are independent Brownian motions.
This stochastic differential system represents the mean-field
interaction of agents driven by averaged compromise and cumulative
stochastic perturbations.

In the limit $N\to\infty$, we assume that the interaction kernels
$P$ and $D$ are globally Lipschitz continuous and of at most linear
growth, ensuring well-posedness of the interacting particle system.
We further assume that the initial data are independent identically
distributed with law $f_0\in\mathcal P_2(\mathbb R^d)$ and that the
driving Brownian motions are independent.
Under these assumptions, classical propagation of chaos results
yield convergence of the empirical measure toward a nonlinear
McKean-Vlasov process, whose law $f(x,t)$ solves the nonlocal
Fokker-Planck equation 
\begin{equation}\label{eq:mf_FP}
	\partial_t f(x,t)
	=
	-\,\nabla_x\!\cdot\!\left(f(x,t)\,\mathcal{P}[f](x,t)
	-\frac12\,\nabla_x\!\bigl(f(x,t)\,\mathcal{D}[f](x,t)\bigr)\right),
\end{equation}
where
\[
\begin{aligned}
	\mathcal{P}[f](x,t)
	&=
	\int_{\mathbb{R}^d} P(x,x_*)(x_*-x)\,f(x_*,t)\,dx_*,
	\cr
	\mathcal{D}[f](x,t)
	&=
	\int_{\mathbb{R}^d} D(x,x_*)^2\,f(x_*,t)\,dx_*,
\end{aligned}
\]
see for example \cite{frank2005nonlinear} for a general reference.

\subsection{Kinetic description via binary interactions}
In the mean-field regime $N\to\infty$, under exchangeability and
propagation of chaos, the interacting particle system admits a
mesoscopic description in terms of the one particle density
$f(x,t)$. Within the kinetic framework, interactions are modeled as
stochastic binary collisions between statistically independent agents
with joint distribution $f(x,t)f(x_*,t)$.

In this context, the binary interaction mechanism associated with
\eqref{eq:binary_disc} can be written in the equivalent form as
\begin{equation}\label{eq:binary}
	\begin{aligned}
		x' &= x + \delta\, P(x,x_*)(x_* - x)
		+ \sqrt{\delta}\, D(x,x_*)\,\xi, \\
		x_*' &= x_* + \delta\, P(x_*,x)(x - x_*)
		+ \sqrt{\delta}\, D(x_*,x)\,\xi_*,
	\end{aligned}
\end{equation}
where $(x,x_*)$ and $(x',x_*')$ denote the pre- and post-interaction
states, respectively, $\delta>0$ measures the interaction strength,
and $\xi,\xi_*$ are centered random vectors describing stochastic
fluctuations. The evolution of the particle density $f(\cdot)$ can be described by a Boltzmann-type equation of the following form
\begin{equation}\label{eq:boltzmann_eq}
	\partial_t f(x,t) = Q(f,f)(x,t),
\end{equation}
with interaction operator
\begin{equation}\label{eq:boltzmann_operator}
	Q(f,f)(x)
	=
	\mathbb{E}\!\left[
	\lambda \int_{\mathbb{R}^d}
	\left(
	\frac{1}{J'} f(x')f(x_*') - f(x)f(x_*)
	\right) dx_*
	\right],
\end{equation}
where $J'$ is the Jacobian of the transformation
$(x,x_*) \mapsto (x',x_*')$ and $\lambda>0$ is the interaction
frequency, see for example \cite{toscani2006kinetic,pareschi2013interacting}.

\paragraph{Quasi-invariant interaction limit and Fokker--Planck equation}
In weak form, for any test function $\varphi\in C_c^{2,\alpha}(\mathbb{R}^d)$,
with $\alpha\in(0,1]$, we can write
\begin{equation}\label{eq:kinetic_weak}
	\frac{d}{dt}\int_{\mathbb{R}^d} \varphi(x)f(x,t)\,dx
	=
	{\lambda}
	\iint_{\mathbb{R}^{2d}}
	\mathbb{E}\!\left[\varphi(x')-\varphi(x)\right]
	f(x,t)f(x_*,t)\,dx\,dx_*.
\end{equation}
To derive the macroscopic limit, we consider the quasi-invariant scaling
\begin{equation}\label{eq:scaling}
	\delta=\varepsilon,
	\qquad
	\lambda={1}/{\varepsilon},
	\qquad 0<\varepsilon\ll1,
\end{equation}
corresponding to high frequency but weak interactions.
Then the scaled dynamics reads
\[
x' - x
=
\varepsilon P(x,x_*)(x_* - x)
+ \sqrt{\varepsilon} D(x,x_*) \xi,
\]
and using Taylor's expansion up to second order for $\varphi\in C_c^{2,\alpha}(\mathbb{R}^d)$ we have
\[
\varphi(x')-\varphi(x)
=
\nabla\varphi(x)\cdot(x'-x)
+\frac12 (x'-x)^\top \nabla^2\varphi(x)(x'-x)
+\mathcal{O}(|x'-x|^{2+\alpha}).
\]
Then, the limiting Fokker-Planck equation \eqref{eq:mf_FP} is recovered in weak form
for every test function $\varphi\in C_c^{2,\alpha}(\mathbb{R}^d)$ for
$0<\alpha\le 1$. Indeed, using the previous Taylor expansion in \eqref{eq:kinetic_weak} and controlling the reminder, in the limit for $\varepsilon\to 0$ one obtains the following equation
\begin{equation}\label{eq:kinetic_fokkerplanck}
	\begin{aligned}
		&\frac{d}{dt}\int_{\mathbb{R}^d} \varphi(x)\,f(x,t)\,dx
		=\cr
		&\qquad\int_{\mathbb{R}^d}
		\nabla \varphi(x)\cdot \mathcal{P}[f](x,t)\, f(x,t)\,dx
		+
		\frac12
		\int_{\mathbb{R}^d}
		\Delta \varphi(x)\, \mathcal{D}[f](x,t)\, f(x,t)\,dx.
	\end{aligned}
\end{equation}
Integration by parts gives the strong nonlinear Fokker-Planck, which
coincides with the forward Kolmogorov equation \eqref{eq:mf_FP} associated
with the nonlinear mean-field stochastic dynamics
\eqref{eq:meanfield_sde}, providing consistency between the microscopic,
kinetic, and mean-field descriptions of the model hierarchy. We refer for further details on this limit to \cite{toscani2006kinetic,pareschi2013interacting}.

\section{Discovery of drift and diffusion interaction kernels}\label{sec:discovery}
The discrete dynamics \eqref{eq:binary_disc} of the ensemble of $N$ agents  can be conveniently written in a vectorized form, introducing the full vector $X^n\in\mathbb{R}^{dN}$ that collects the agent states in $d$-blocks, i.e., 
$X^n = \mathrm{vec}(x_1^n, \ldots, x_N^n) \in \mathbb{R}^{dN}$, where $x_i^n\in\mathbb{R}^d$ is the feature of $i$-agent at time $t_n:=n\Delta t$ with $n=0,\ldots, M$.
To identify the interacting agent associated to each $i$-agent, we consider the permutation matrix $S^{n}\in\mathbb{R}^{N\times N}$ encoding pairings at time step $n$, and define the block action on $\mathbb{R}^{dN}$ by
$$\mathsf{S}^{n} := S^{n}\otimes \mathsf{I}_d \in \mathbb{R}^{dN\times dN},$$ where the interacting $j(i)$-agent is given by $(\mathsf{S}^{n}X^n)_i = x_{j(i)}^n$. Then, the vectorized form  of the full dynamics \eqref{eq:binary_disc} reads as
\begin{equation}\label{eq:binary_matrix}
	X^{n+1}
	= X^n 
	+ \Delta t\, \mathbf{P}^n \odot \big( \mathsf{S}^{n}X^n - X^n \big)
	+ \sqrt{\Delta t}\, \mathbf{D}^n \odot \Xi^n,
\end{equation}
where $\odot$ stands for the Hadamard product. In \eqref{eq:binary_matrix} we introduce, through the Kronecker product $\otimes,$ the vectorized interaction and diffusion coefficients
\[
\mathbf{P}^n := P^n \otimes \mathbf{1}_d \in \mathbb{R}^{dN}, 
\qquad 
\mathbf{D}^n := D^n \otimes \mathbf{1}_d \in \mathbb{R}^{dN},
\]
where $P^n, D^n \in \mathbb{R}^N$ collect the scalar weights associated with each agent at time $t_n$, 
and $\mathbf{1}_d = (1,\ldots,1)^\top \in \mathbb{R}^d$ replicates them over the $d$ components of each block.
Specifically, the entries of $P^n$ and $D^n$ are given by
\begin{equation}\label{eq:ker_diff}
	P_i^n = P(x_i^n, x_{j(i)}^n), 
	\qquad 
	D_i^n = D(x_i^n, x_{j(i)}^n),
\end{equation}
corresponding to the values of the interaction and diffusion functions evaluated on the current pair $(x_i^n, x_{j(i)}^n)$.
Finally, we consider $\Xi^n\in\mathbb{R}^{dN}$ to be the stacked noise as a collection of each of the random variables $\xi_i^n\in\mathbb{R}^d$, i.e. $\Xi^n=(\xi^n_1,\ldots,\xi_N^n)^\top$. 
The stochastic discrete dynamics \eqref{eq:binary_matrix} describes the microscopic evolution of interacting agents driven by deterministic pairwise interactions and random perturbations. 
Due to the random pairing matrices $S^{n}$ and stochastic terms $\Xi^n$, the trajectories are random processes even when $P$ and $D$ are deterministic.

Our main goal is to reconstruct suitable approximations of the interaction kernel $P(\cdot)$ and diffusion function $D(\cdot)$ from observed data, following approaches similar to \cite{bongini2017inferring,LMT21,LMT22}. Hence, given the agent trajectories  samples $\{X^n\}_{n=0}^M$, we seek interaction and diffusion 
operators that best explain the observed dynamics, namely
\begin{equation}\label{eq:inverse_general}
	(P^*,D^*)
	=
	\argmin_{P \in \mathscr{A}_P,\; D \in \mathscr{A}_D}
	\mathcal{E}\big(P,D;\{X^n\}_{n=0}^M\big),
\end{equation}
where $\mathscr{A}_P$ and $\mathscr{A}_D$ are admissible classes of interaction and diffusion 
functions defined below, and $\mathcal{E}$ is a suitable empirical loss functional encoding the discrepancy 
between the model dynamics  \eqref{eq:binary_disc} and the observed trajectories.

In the remainder of this manuscript we will focus on the relevant case of {\it radial} interaction kernels and diffusions in \eqref{eq:ker_diff}, namely we assume 
$$P(x,y) = P(r),\qquad D(x,y) = D(r),\qquad r:=\lvert| x - y\rvert|,$$
which is standard framework in interacting particle systems such as opinion dynamics, aggregation models, or flocking systems. Nevertheless, the main difficulties of the reconstruction problem remain unchanged, since the stochasticity of the dynamics and the lack of observability of the interaction pairs still lead to ill-posed inverse formulations.

\begin{figure}[h]
	\centering
	\includegraphics[width=0.35\linewidth]{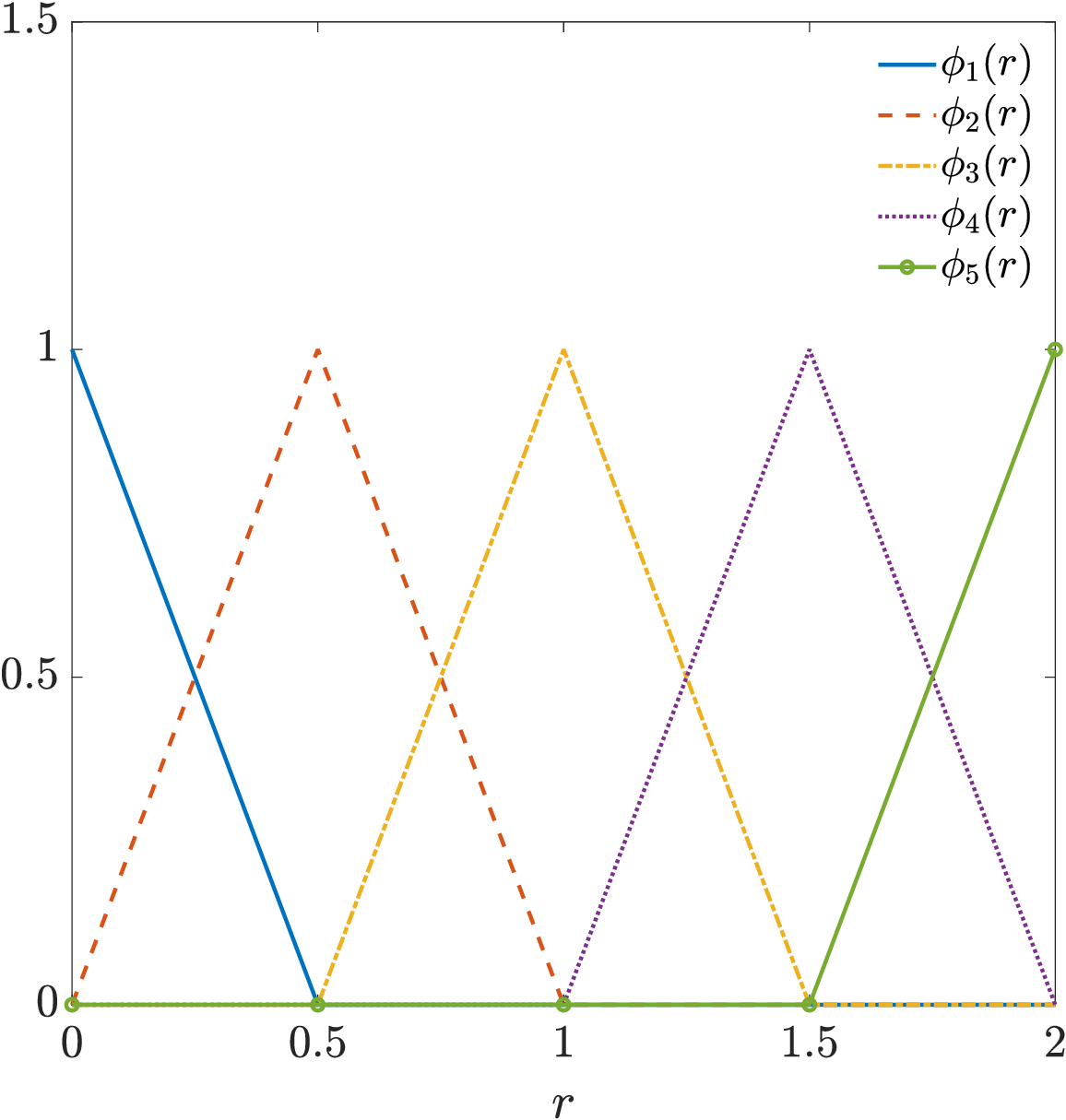}\quad\includegraphics[width=0.35\linewidth]{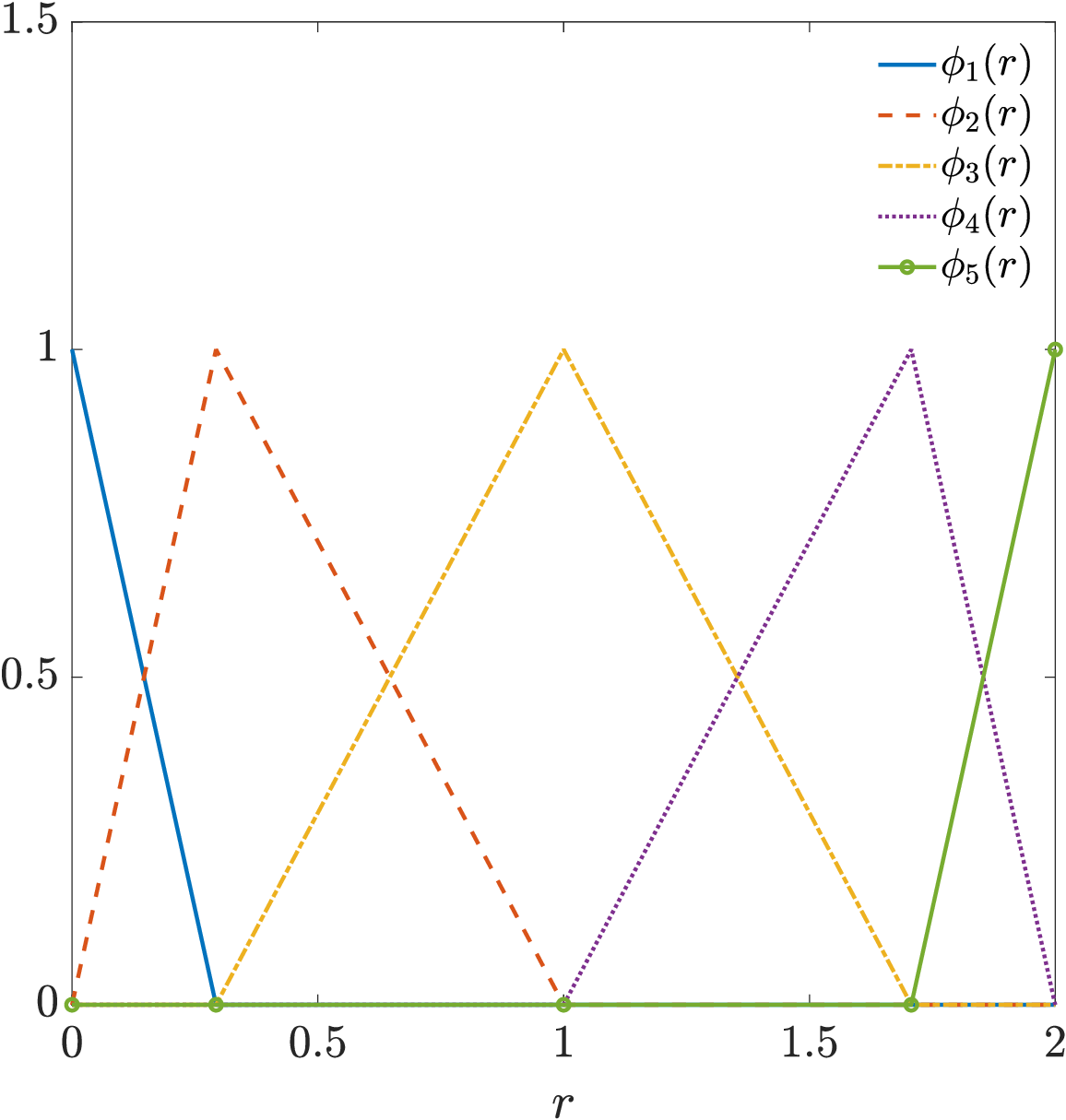}
	\caption{Case of domain $[0,2]$, example of $N_b^P=5$ piecewise linear basis functions on a uniform mesh (left) and on a Chebyshev mesh (right).}
	\label{fig:basis}
\end{figure}

To realize the inverse problem \eqref{eq:inverse_general} we introduce a finite-dimensional approximation in terms of a family of one-dimensional non-negative basis functions $\{\Phi_k(\cdot)\}_{k\ge 1}$ and $\{\Psi_k(\cdot)\}_{k\ge 1}$. Specifically, we selected a family of piecewise linear and compactly supported basis functions defined on suitable discretizations of the domains of $P$ and $D$ (see an example in Figure \ref{fig:basis}).

Thus, we can introduce the approximation spaces that we will use in \eqref{eq:inverse_general} as follows
\[
\mathscr{A}_P := \mathrm{span}\{\Phi_1,\ldots,\Phi_{N_b^P}\},
\qquad
\mathscr{A}_D := \mathrm{span}\{\Psi_1,\ldots,\Psi_{N_b^D}\},
\]
with respective dimensions $N_b^P$ and $N_b^D$ and the approximations of the drift and diffusion
kernels read as
\begin{equation}\label{eq:Phat}
	\widehat P(r):=\sum_{k=1}^{N_b^P} \varrho_k\, \phi_k(r),
	\qquad
	\widehat D^2(r) := 2 \sum_{k=1}^{N_b^D} \zeta_k\, \psi_k(r),
\end{equation}
where $\varrho \in \mathbb{R}^{N_b^P}$ and $\zeta \in \mathbb{R}^{N_b^D}$ are the vector coefficient to be determined by means of the optimization problem \eqref{eq:inverse_general}.

We introduce the design matrices
${\bf\Theta}^n=[\Theta_1^n,\ldots, \Theta_{N_b^P}^n]\in\mathbb{R}^{dN\times N_b^P}$ and ${\bf\Lambda}^n=[\Lambda_1^n,\ldots, \Lambda_{N_b^D}^n]\in\mathbb{R}^{dN\times N_b^D}$ such that their $k-$th columns at time $t_n$ are given by
\begin{equation}\label{eq:matrices_reg}
	\begin{aligned}
		\Theta^{n}_{k}
		&=
		\phi_k\!\left(\,\lVert \mathsf{S}^{n}X^n - X^n\rVert\,\right)
		\odot \bigl(\mathsf{S}^{n}X^n - X^n\bigr),
		\qquad &k = 1,\ldots,N_b^P,
		\\
		\Lambda^{n}_{k}
		&=
		\psi_k\!\left(\,\lVert \mathsf{S}^{n}X^n - X^n\rVert\,\right),
		\qquad &k = 1,\ldots,N_b^D,
	\end{aligned}
\end{equation}
then the microscopic dynamics \eqref{eq:binary_matrix} can be rewritten as follows
\begin{equation}\label{eq:binary_finiteapprox}
	X^{n+1} = X^{n} + \Delta t {\bf\Theta}^n\varrho+ \sqrt{2\Delta t {\bf\Lambda}^n\zeta}\odot\Xi^n.
\end{equation}
Hence, the identification problems \eqref{eq:inverse_general} can be addressed considering two different least square problems, one for the drift as
\begin{equation}\label{eq:regression_kern_full}
	\varrho^*
	= \argmin_{\varrho \in \mathcal{A}_P}
	\frac{1}{M_P}
	\sum_{n=1}^{M_P}
	\left\|
	X^{n+1} - X^n
	- \Delta t\,\mathbb{E}\left[{\bf\Theta}^{n}\varrho
	|X^n\right]\right\|^2,
\end{equation}
and one for the diffusion as
\begin{equation}\label{eq:regression_diff_full}
	\zeta^*
	= \argmin_{\zeta \in \mathcal{A}_D}
	\frac{1}{M_D}
	\sum_{n=1}^{M_D}
	\left\|
	\bigl(X^{n+1}-X^n\bigr)\odot\bigl(X^{n+1}-X^n\bigr)
	- 2\Delta t\,\mathbb{E}\left[{\bf\Lambda}^{n}\zeta
	|X^n\right]\right\|^2,
\end{equation}
where the conditional expectation $\mathbb{E}[\cdot|X^n]$ is taken with respect to the stochastic variables associated with the noise term $\Xi^n$ and the random pairing matrix $\mathsf{S}^n$. Note that potentially different numbers of snapshots $M_P,M_D\leq M$ can be employed for each problem. The admissibility constraints $\mathscr{A}_P$ and $\mathscr{A}_D$ are now enforced through the coefficient spaces. For the drift, we use the constrained space 
\begin{equation}\label{eq:AP}
	\mathcal{A}_P= \left\{\rho\in\mathbb{R}^{N_b^P} : \rho_1 = P_0, \,\kappa_P\rho _k \leq \kappa_P \rho_{k + 1}, \, k = 1,\dots,N_b^P-1\right\}, 
\end{equation}
with $\overline \rho$ value of the kernel in $r=0$, $\kappa_P=1$ for repulsive dynamics, $\kappa_P=-1$ for attractive ones, while for $\kappa_P = 0$ we do not impose any monotonicity. For the diffusion, instead, we impose non-negativity via
\begin{equation}\label{eq:AD}
	\mathcal{A}_D := \left\{\zeta \in \mathbb{R}^{N_b^D} : \zeta_k \ge 0, \, \kappa_D\zeta _k \leq \kappa_D \zeta_{k + 1},  \ k = 1,\ldots,N_b^D-1,\, \zeta_i = \overline \zeta_i \mbox{ for some }i  \right\}.
\end{equation}
where the coefficients $\kappa_D$ play the same role of $\kappa_P$ in \eqref{eq:AP}.
This choice of $\mathcal{A}_D$ ensures that the reconstructed diffusion coefficient $\widehat D(r)^2$ in \eqref{eq:Phat} remains non-negative on the reconstruction grid.

\begin{remark}
	In general, the interaction matrices $\mathsf{S}^n$ governing the microscopic dynamics are not directly observable from trajectory data, and they constitutes a source of uncertainty in the reconstruction problem, affecting both the statistical variability of the data and the effective exploration of the dynamics. As a consequence, the stochastic evolution \eqref{eq:binary_matrix} cannot be identified pathwise and must instead be reconstructed in a statistical sense. To this end, in the previous general formulation, we rely on It\^o--Taylor expansions to derive suitable moment-based approximations of the drift and diffusion terms from the observed trajectories, which form the basis of the least squares problem \eqref{eq:inverse_general}. More precisely, for the interaction rule \eqref{eq:binary_matrix}, the following relations hold at each time step for the drift term
	\begin{equation}\label{eq:approx_kernel_vec2}
		\mathbb{E}\!\left[
		\mathbf{P}^n \odot \bigl( \mathsf{S}^{n} X^n - X^n \bigr)
		\;\middle|\; X^n
		\right]
		=
		\frac{1}{\Delta t}\,
		\mathbb{E}\!\left[
		X^{n+1} - X^n
		\;\middle|\; X^n
		\right]
		+ \mathcal{O}(\Delta t),
	\end{equation}
	and for the diffusion term
	\begin{equation}\label{eq:diff_ito_taylor}
		\frac{1}{2}\,
		\mathbb{E}\!\left[
		\bigl( \mathbf{D}^n \odot \mathbf{D}^n \bigr)
		\;\middle|\; X^n
		\right]
		=
		\frac{1}{2\,\Delta t}
		\mathbb{E}\!\left[
		\bigl( X^{n+1} - X^n \bigr)
		\odot
		\bigl( X^{n+1} - X^n \bigr)
		\;\middle|\; X^n
		\right]
		+ \mathcal{O}(\Delta t),
	\end{equation}
	where the conditional expectations are taken with respect to both the latent interaction variables $\mathsf{S}^n$ and the microscopic noise realizations $\Xi^n$, conditioned on the observed state $X^n$. These expansions show that, at leading order in $\Delta t$, the drift and diffusion contributions can be approximated independently through the first and second conditional moments of the observed increments.
	In the following, we first analyze an idealized setting in which the interaction matrices $\mathsf{S}^n$ are assumed to be known.
	This assumption will be relaxed in a subsequent section, where the effects of latent interactions are explicitly taken into account.
\end{remark}

\subsection{Kernel estimations with known interaction matrices}\label{sec:exS}
We first assume to have the dataset $\left\{t_n,X^n,\mathsf{S}^{n}\right\}_{n=0}^M$, meaning that we have access to the state of the agents at certain times $t_n$ and also to their pairwise interaction matrix $\mathsf{S}^{n}$.  
Given this information over $M_P\le M$ time frames, the coefficients of the interaction kernel are obtained by solving the least-squares problem
\begin{equation}\label{eq:regression_kern}
	\varrho^*
	= \argmin_{\varrho \in \mathbb{R}^{N_b^P}}
	\frac{1}{M_P}
	\sum_{n=1}^{M_P}
	\left\|
	X^{n+1} - X^n
	- \Delta t\,{\bf\Theta}^{n}\varrho
	\right\|^2.
\end{equation}
Here, in contrast to \eqref{eq:regression_kern_full}, the interaction matrices $\mathsf{S}^{n}$  are assumed to be known. As a consequence, all interaction dependent quantities, such as the increments $\mathsf{S}^{n}X^n - X^n$  and their associated basis function evaluations, are directly observable from the data. The resulting estimator is therefore affected solely by the statistical fluctuations induced by the stochastic noise, and not by uncertainty in the underlying pairing patterns. This reduced source of variability generally leads to better conditioned least squares problems and more stable parameter estimates. 
Similarly, for the diffusion interaction kernel we consider in general $M_D\le M$ time frames. The diffusion coefficients are determined from
\begin{equation}\label{eq:regression_diff}
	\zeta^*
	= \argmin_{\zeta \in \mathcal{A}_D}
	\frac{1}{M_D}
	\sum_{n=1}^{M_D}
	\left\|
	\bigl(X^{n+1}-X^n\bigr)\odot\bigl(X^{n+1}-X^n\bigr)
	- 2\Delta t\,{\bf\Lambda}^{n}\zeta
	\right\|^2,
\end{equation}
where ${\bf \Lambda}^{n}$ is defined columnwise as in \eqref{eq:matrices_reg}.

Both \eqref{eq:regression_kern} and \eqref{eq:regression_diff} can be cast as quadratic programs with linear constraints and solved with standard constrained least-squares solvers. Here we adopt an interior-point method \cite{NW06}, other suitable choices include projected-gradient methods, active-set algorithms, or nonnegative least-squares (NNLS). Moreover, since the estimators \eqref{eq:regression_kern} and \eqref{eq:regression_diff} are fully decoupled, they can be computed independently and may use different time windows and sample sizes, so in general we allow $M_P \neq M_D$. 

\paragraph{Test 1: numerical validation for known interaction matrices}

We consider a system of $N=10^5$ interacting agents and $M=201$ snapshots of the dynamics, uniformly sampled with time step $\Delta t = 0.01$. 
The interaction kernel governing the drift is given by
\[
P(r) = (1 + r^2)^{-2},
\]
while the diffusion term is defined through the pairwise interaction
\[
D(x,y) = \tilde D(\lvert x - y \rvert)\,(y-x),
\qquad
\tilde D(r) = {0.25}/{(1+r)^2}.
\]
Thus, we are interested in reconstructing the kernel $P$ and the radial part of the diffusion, meaning $\tilde D$. Notice that the functional form of $D(x,y)$ is slightly more general than the one considered in the theoretical discussion above. 
However, this additional structure does not affect the proposed reconstruction procedure.

We choose $N_b^P=10$ basis functions for the kernel reconstruction and $N_b^D=8$ basis functions for the diffusion, and we use $M_P = 20$, and $M_D = 10$ consecutive time frames for the learning process, meaning that we only look up to time $t=0.2$ for the kernel reconstruction and up to time $t = 0.1$ for the diffusion, even with an available time horizon of $[0,2]$. 

\begin{figure}[htbp]
	\centering
	\includegraphics[height=5.4cm]{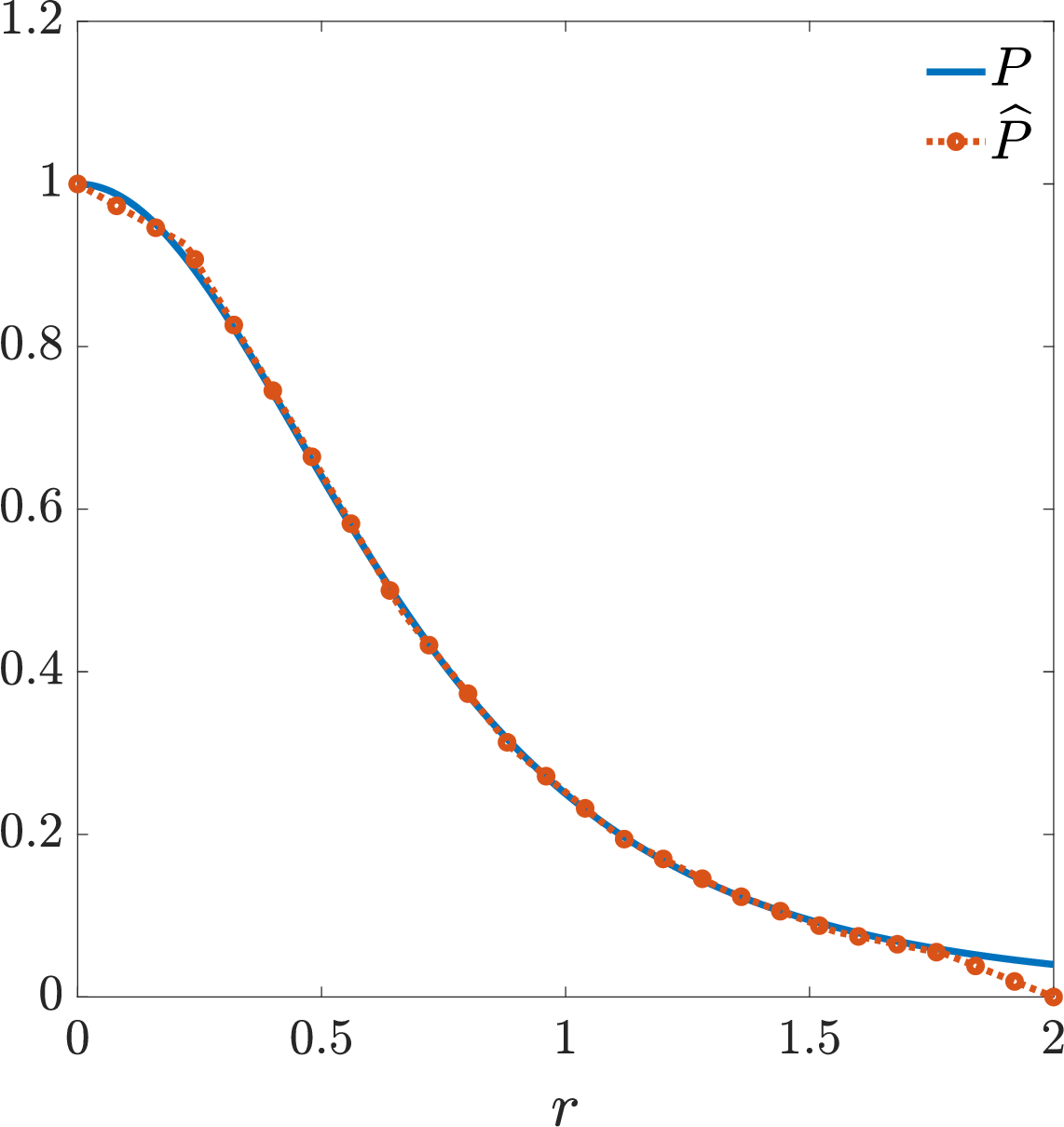}\quad\includegraphics[height=5.4cm]{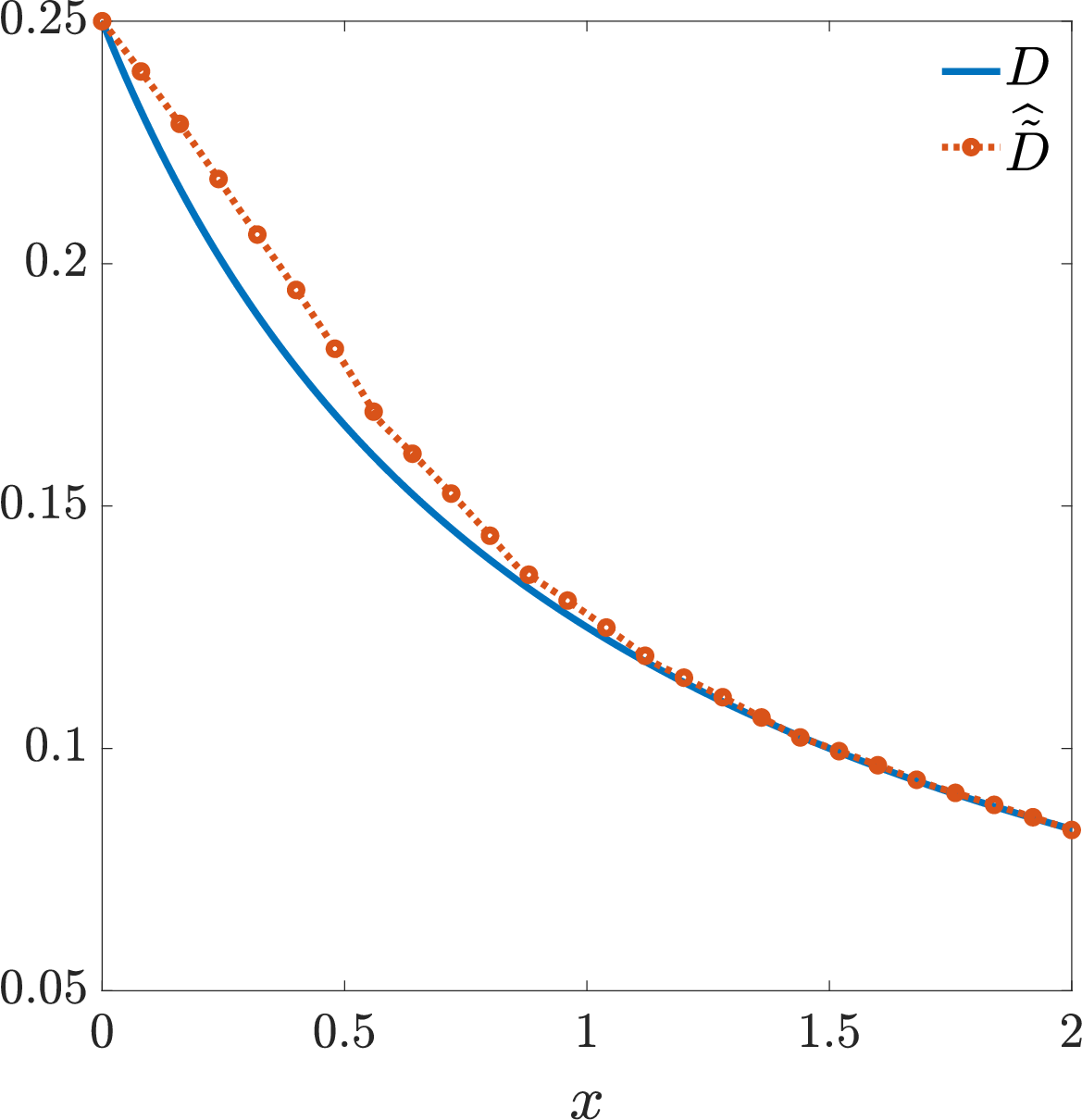}\quad\includegraphics[height=5.4cm]{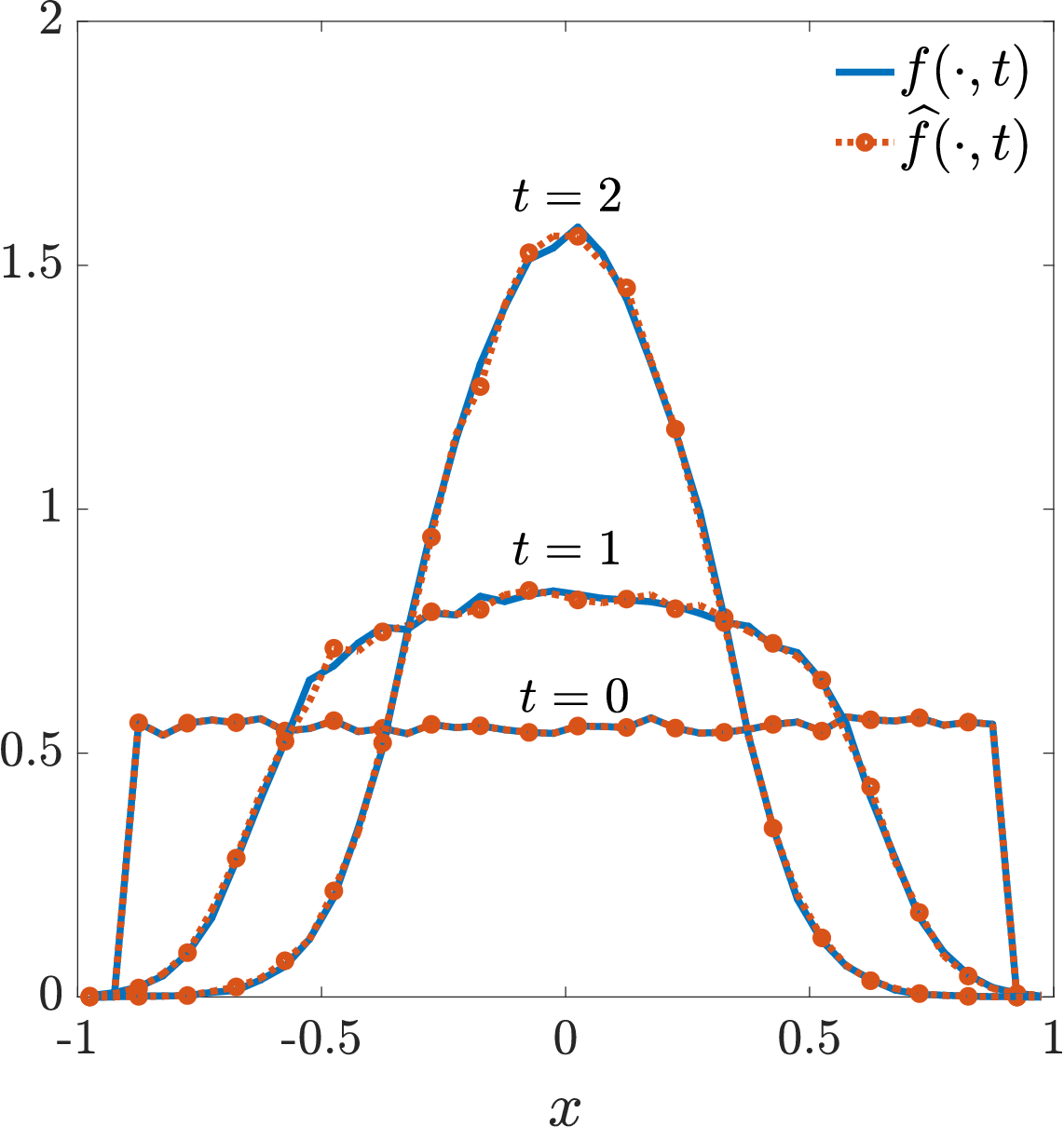}
	\caption{Reconstructed kernel $P(r) = (1 + r^2)^{-2}$ (left) using $N^P_b=10$ basis functions, reconstructed radial diffusion $D(r) = 0.25/(1+r)^2$ using $N^D_p=8$ basis functions (center), comparison between the data density $f$ and the reconstructed $\widehat{f}$ at the final time (right).}
	\label{fig:S_known}
\end{figure}

Figure~\ref{fig:S_known} shows (left) the exact interaction kernel $P$ together with its reconstruction $\widehat{P}$, (center) the exact radial diffusion profile $\tilde D$ and its reconstruction $\widehat{\tilde D}$, and (right) the particle density at various times. 
The latter is compared with the reconstructed density $\widehat f(\cdot,t)$ obtained by simulating the particle dynamics \eqref{eq:binary_matrix} with the learned kernels $\widehat{P}$ and $\widehat{D}$, getting the reconstructed particle trajectories $\widehat X^n = \textrm{vec}(\widehat x_1^n,\dots,\widehat x_N^n)$.
The empirical densities associated with the exact and reconstructed particle systems are defined as
\[
f(x,t_n) := \frac{1}{N}\sum_{i=1}^N \delta\!\bigl(x - x_i^n\bigr),
\qquad
\widehat f(x,t_n) := \frac{1}{N}\sum_{i=1}^N \delta\!\bigl(x - \widehat x_i^n\bigr).
\]

To quantify the reconstruction accuracy, for a generic target function $F$ and its approximation $\widehat {F}$ we consider the relative errors
\begin{equation}\label{eq:errorsk}
	E_{F}^{1} := \frac{\|{F}-\widehat {F}\|_{L^{1}(I)}}{\|{F}\|_{L^{1}(I)}},
	\qquad
	E_F^{\infty} := \frac{\|{F}-\widehat {F}\|_{L^{\infty}(I)}}{\|{F}\|_{L^{\infty}(I)}},
\end{equation}
with $I=[0,2]$ for $F\equiv P$ and 
for $F\equiv \tilde D$. 

To quantify the accuracy of the reconstructed trajectories, we define the time-averaged and final-time reconstruction errors
\begin{equation}\label{eq:w1}
	E_f^{\mathrm{ave}}
	:=
	\frac{1}{N_T}\sum_{n=1}^{N_T} W_1(X^n,\widehat X^n),
	\qquad
	E_f^{\mathrm{fin}}
	:=
	W_1(X^{N_T},\widehat X^{N_T}),
\end{equation}
which measure the discrepancy between the empirical densities $f(\cdot,t_n)$ and $\widehat f(\cdot,t_n)$, and where we use the $1$-Wasserstein distance between the empirical measures associated with the particle configurations $X^n$ and $\widehat X^n$ at time $t_n$, denoted by $W_1(X^n,\widehat X^n)$. In one spatial dimension, this distance admits the explicit representation
\[
W_1(X^n,\widehat X^n)
=
\frac{1}{N}\sum_{i=1}^N
\bigl|x_{\pi(i)}^n - \widehat{x}_{\pi'(i)}^n\bigr|,
\]
where $x_{\pi(i)}^n$ and $\widehat{x}_{\pi'(i)}^n$ denote the sorted samples of the two particle ensembles.

In this simulation, the kernel reconstruction errors are $E_P^{1}=1.49\cdot 10^{-2}$ and $E_P^{\infty}=4.00\cdot 10^{-2}$, while the diffusion errors are $E_D^{1}=3.68\cdot 10^{-2}$ and $E_D^{\infty}=6.65\cdot 10^{-2}$. Moreover, the density reconstruction errors are $E_f^{\mathrm{ave}}=1.09\cdot 10^{-3}$ and $E_f^{\mathrm{fin}}=1.31\cdot 10^{-3}$. We summarize the error in table \ref{tab:errors_PDf}.
\begin{table}[htbp]
	\centering
	\begin{tabular}{l c c}
		& $F\equiv P$ &  $F\equiv D$ \\
		\hline
		\hline
		$E_F^{1}$ & $1.49\cdot 10^{-2}$ & $3.68\cdot 10^{-2}$ \\
		$E_F^{\infty}$ & $4.00\cdot 10^{-2}$ & $6.65\cdot 10^{-2}$ \\
		\hline
	\end{tabular}
	\quad
	\begin{tabular}{l c}
		&  \\
		\hline
		\hline
		$E_f^{\mathrm{ave}}$ & $1.09\cdot 10^{-3}$ \\
		$E_f^{\mathrm{fin}}$ & $1.31\cdot 10^{-3}$ \\
		\hline
	\end{tabular}
	\caption{Test 1: Relative reconstruction errors for the interaction kernel $P$ and diffusion $D$ and for the empirical density $f(\cdot)$.}
	\label{tab:errors_PDf}
\end{table}

\subsection{Random-batch kernel estimation with unknown interaction pairs}\label{sec:batch}
In the general setting, the available data consist only of agent trajectories $\{t_n,X^n\}_{n=0}^M$, while the pairing matrices $\mathsf{S}^{n}$ and noise realizations $\Xi^n$ remain unobserved. 

In this setting, to approximate the conditional expectations in \eqref{eq:regression_kern_full} and \eqref{eq:regression_diff_full} we replace it with a collection of randomly generated batch interaction maps $S^{(n,h)},\ h=1,\ldots,N_p$, where each $S^{(n,h)}$ is a permutation matrix selecting one random interacting pair and $N_p\ll N$ denotes the number of sampled pairs. The conditional expectations are then approximated by empirical averages over these sampled batches, 
and given this information over $M_P\le M$ time frames for the drift kernel, the coefficients are obtained by solving
\begin{equation}\label{eq:regression_kern_batch}
	\varrho^*
	= \argmin_{\varrho \in \mathbb{R}^{N_b^P}}
	\frac{1}{M_P}
	\sum_{n=1}^{M_P}
	\left\|
	X^{n+1} - X^n
	- \frac{\Delta t}{N_p}
	\sum_{h=1}^{N_p}
	{\bf\Theta}^{(n,h)}\,\varrho
	\right\|^2,
\end{equation}
where
\[
{\bf\Theta}^{(n,h)}
:= {\bf\Theta}\left(\|{S}^{(n,h)}X^{n}-X^{n}\|\right)\odot\left({S}^{(n,h)}X^{n}-X^{n}\right)
\in\mathbb{R}^{dN\times N_b^P}.
\]
Here, differently from \eqref{eq:regression_kern}, the interaction matrices are unknown and must be approximated through random batching. This introduces additional variability beyond the statistical fluctuations from stochastic noise, as the estimator now depends on both the noise realization and the uncertainty in pairing patterns.

Similarly, for the diffusion interaction kernel we consider in general $M_D\le M$ time frames. The diffusion coefficients are determined from
\begin{equation}\label{eq:regression_diff_batch}
	\zeta^*
	= \argmin_{\zeta \in \mathcal{A}_D}
	\frac{1}{M_D}
	\sum_{n=1}^{M_D}
	\left\|
	\bigl(X^{n+1}-X^n\bigr)\odot\bigl(X^{n+1}-X^n\bigr)
	- \frac{2\Delta t}{N_p}
	\sum_{h=1}^{N_p}
	{\bf\Lambda}^{(n,h)}\zeta
	\right\|^2,
\end{equation}
where
\[
{\bf\Lambda}^{(n,h)}
:= {\bf\Lambda}\left(\|{S}^{(n,h)}X^{n}-X^{n}\|\right)
\in\mathbb{R}^{dN\times N_b^D}.
\]
These formulations provide unbiased Monte Carlo approximations of the drift and diffusion contributions, enabling kernel reconstruction through batch averaging in analogy with the random batch method \eqref{eq:rbm_disc}.
In particular, the scheme
\begin{equation}\label{eq:regr_Sh}
	X^{n+1} = X^{n}
	+\frac{\,\Delta t}{N_p}
	\sum_{h=1}^{N_p}{\bf\Theta}^{(n,h)}\,\varrho+\left(\sqrt{\frac{2\,\Delta t}{N_p}
		\sum_{h=1}^{N_p}{\bf\Lambda}^{(n,h)}\,\zeta}\right)\Xi^n,
\end{equation}
provides an unbiased estimator of the dynamics \eqref{eq:binary_matrix}.
\paragraph{Robust kernel estimation for unknown interaction pairs}
To enhance the flexibility and robustness of the least-squares estimation, we introduce two refinements. First, we employ an enlarged temporal window $X^{n+\ell}-X^n$ to capture interaction effects over multiple time steps, provided the dynamics remains in the weak-interaction regime over the interval $\ell\Delta t>0$. Second, since each trajectory yields only a noisy surrogate of the unobserved interaction structure, we repeat the random-batch regression procedure $K$ times to improve reconstruction accuracy.
Hence, we consider a sequence of regression problems $k=1,\ldots,K$ respectively for the drift  and diffusion
\begin{align}\label{eq:regression_Pk}
	\varrho_k
	&=
	\arg\min_{\varrho\in \mathcal{A}_P}
	\frac{1}{M_P}
	\sum_{n=1}^{M_P}
	\Bigl\|
	X^{n+\ell}-X^{n}
	-\frac{\ell\,\Delta t}{N_p}
	\sum_{h=1}^{N_p}
	{\bf\Theta}^{(n,h,k)}\,\varrho
	\Bigr\|^{2},
	\\
	\label{eq:regression_Dk}
	\zeta_k
	&=
	\arg\min_{\zeta\in \mathcal{A}_D}
	\frac{1}{M_D}
	\sum_{n=1}^{M_D}
	\Bigl\|
	\bigl(X^{n+\ell}-X^n\bigr)\odot\bigl(X^{n+\ell}-X^n\bigr)
	- \frac{2\ell\Delta t}{N_p}\sum_{h=1}^{N_p}\,{\bf\Lambda}^{(n,h,k)}\zeta
	\Bigr\|^{2},
\end{align}
where the matrices $\Theta^{n,h,k)}$ and $\Lambda^{(n,h,k)}$ are constructed as in \eqref{eq:matrices_reg}, with the interaction structure provided by the batch permutation matrices $\{S^{(n,h,k)}\}_{h,k}$ at time $t_n$.
For each realization, we store the aggregated batch design matrices as
$$\overline{ {\bf\Theta}}^{(n,k)}:=\frac{1}{N_p}\sum_{h=1}^{N_p} {\bf\Theta}^{(n,h,k)},\qquad \overline{ {\bf\Lambda}}^{(n,k)}:=\frac{1}{N_p}\sum_{h=1}^{N_p} {\bf\Lambda}^{(n,h,k)},$$
and we solve the following  weighted regression problems
\begin{align}
	\label{eq:regression_Popt}
	\varrho^*
	&=
	\arg\min_{\varrho\in\mathcal{A}_P}
	\sum_{k=1}^{K}
	\frac{w_k}{M_P}
	\sum_{n=1}^{M_P}
	\Bigl\|
	X^{n+\ell}-X^{n}
	-\ell\,\Delta t\, \overline{{\bf\Theta}}^{(n,k)}\,\varrho
	\Bigr\|^{2},
	\\
	\label{eq:regression_Dopt}
	\zeta^*
	&=
	\arg\min_{\zeta\in \mathcal{A}_D}\sum_{k=1}^K
	\frac{w_k}{M_D}
	\sum_{n=1}^{M_D}
	\Bigl\|
	\bigl(X^{n+\ell}-X^n\bigr)\odot\bigl(X^{n+\ell}-X^n\bigr)
	- 2\ell\Delta t\overline{{\bf\Lambda}}^{(n,k)}\zeta
	\Bigr\|^{2}.
\end{align}

To determine the weights $w_k$, we incorporate information from the underlying physics by assessing the reliability of each candidate estimator
$(\varrho_k,\zeta_k)$ through a comparison between the reconstructed dynamics and the observed data. 
For each $k$, we generate a discrete trajectory $\widehat{\bf X}_{\textrm{traj}}^{(k)}\in\mathbb{R}^{dN\times N_T}$ by evolving the
system with the reconstructed kernels $\widehat P_k$ and $\widehat D_k$, defined in \eqref{eq:Phat}, starting from the same initial condition as the one in the data.  
We then quantify the discrepancy between the reconstructed state at the final
time, $\widehat{\bf X}^{(k)}_{\textrm{traj}}$, and the observed state ${\bf X}_{\textrm{traj}}$ via the averaged error estimate defined as in \eqref{eq:w1}, where for each $k\geq1$ we have 
$E_k := E_f^{\textrm{ave}}({\bf X}_{\textrm{traj}},\widehat{\bf X}^{(k)}_{\textrm{traj}}).$
This approach provides an error measure from which we propose two alternative definition for the weights $w_k$.
\begin{itemize}
	\item[$(i)$] { Averaging rule}
	\begin{equation}\label{eq:weights1}
		\overline{E}_k = \frac{E_k}{\sum_{j=1}^{K} E_j},
		\qquad
		w_k = w_k^\textrm{av}:=\frac{1-\overline{E}_k}{K-\sum_{j=1}^{K}\overline{E}_j}.
	\end{equation}
	leading to the optimization problems \eqref{eq:regression_Popt}, \eqref{eq:regression_Dopt}
	and the coefficients $w_k^\textrm{av}$ with resulting kernels defined as 
	\begin{equation}\label{eq:Phatav}
		\widehat P^\textrm{av}(r):=\sum_{q=1}^{N_b^P} \varrho_k^\textrm{av}\, \phi_q(r),
		\qquad
		(\widehat D^\textrm{av})^{2}(r) := 2 \sum_{q=1}^{N_b^D} \zeta_k^\textrm{av}\, \psi_q(r),
	\end{equation}
	\item[$(ii)$]{  Best-fit rule}
	\begin{equation}\label{eq:weights2}
		w_k = w_k^\textrm{best}:=
		\begin{cases}
			1, & \text{if } E_k = \min_{1\le j\le K} E_j,\\[2mm]
			0, & \text{otherwise},
		\end{cases}
	\end{equation}
	leading to the optimization problems \eqref{eq:regression_Popt}, \eqref{eq:regression_Dopt}
	and the coefficients $w_k^\textrm{best}$ with 
	\begin{equation}\label{eq:Phatbest}
		\widehat P^\textrm{best}(r):=\sum_{q=1}^{N_b^P} \varrho_q^\textrm{best}\, \phi_q(r),
		\qquad
		(\widehat D^\textrm{best})^{2}(r) := 2 \sum_{q=1}^{N_b^D} \zeta_q^\textrm{best}\, \psi_q(r).
	\end{equation}
	
\end{itemize}
We report in Algorithm \ref{alg:1} the proposed methodology.
\begin{algorithm}[h!]
	\caption{Kernel reconstruction based on informed RBM ensamble}\label{alg:1}
	\begin{algorithmic}
		\State \texttt{Fix} $N_b^P,N_b^D$, $K$, $N_p$, $\ell$, $M = M_P = M_D$.
		\For{$k = 1:K$}
		\For{$n=1:M$}
		\State \texttt{Select $N_p$ random pairs}
		\State \texttt{Compute} $\frac{1}{N_p}\sum_{h=1}^{N_p}{\bf\Theta}^{(n,h,k)}$.
		\State \texttt{Compute} $\frac{1}{N_p}\sum_{h=1}^{N_p}\left({\bf\Lambda}(\|\mathsf S^{(n,h,k)}X^n-X^n\|)\right)$. 
		\EndFor
		\State \texttt{Compute $(\varrho_{k},\zeta_k)$ solving the regression} \eqref{eq:regression_Pk}--\eqref{eq:regression_Dk}.
		\State \texttt{Reconstruct the trajectories $\widehat{\bf X}^{(k)}_{\textrm{traj}}$} \eqref{eq:binary} \texttt{for random pairs}.
		\State \texttt{Compute the error $E_k$ with respect to data $\{X^n\}_{n=0}^{N_T}$ as in \eqref{eq:w1}}.
		\EndFor
		\State \texttt{Compute the weigths $w_k^\textrm{av}$ as in \eqref{eq:weights1} and $w_k^\textrm{best}$ as in \eqref{eq:weights2}}. 
		\State \texttt{Solve the regression \eqref{eq:regression_Popt}-\eqref{eq:regression_Dopt} with the different weights}
	\State \Return $\widehat{P}^\textrm{av},\widehat{P}^\textrm{best},\widehat{D}^\textrm{av},\widehat{D}^\textrm{best}$
\end{algorithmic}
\end{algorithm}
\begin{remark}
Note that when both drift and diffusion kernels depend on pairwise interactions, the random batch sampling at each time step influences both $P$ and $D$ simultaneously. Consequently, the same batch realizations must be used in \eqref{eq:regression_Popt} and \eqref{eq:regression_Dopt}, which requires setting $M_P = M_D$ and using identical weights $w_k$ and temporal increments $\ell$ for both problems. This constraint is relaxed when the diffusion depends only on the state of a single agent, as shown later in Section~\ref{sec:num}.
\end{remark}

\subsection{Mean-field kernel reconstruction with unknown interaction pairs}\label{sec:mfadensity}
In this subsection, we introduce a mean-field reconstruction approach that replaces the unobserved pairwise interactions with nonlocal averages computed from an estimated empirical density.

In the mean-field regime the effect of the unobserved microscopic pairings can be represented through nonlocal operators acting on the agent density.
This motivates an alternative reconstruction strategy that does not rely on sampling interaction matrices, but instead on estimating the empirical density from the available trajectories.
Let $\Omega\subset\mathbb{R}^d$ be the computational domain. Given the agent configuration at time $t_n$, i.e. $\{x_i^{n}\}_{i=1}^{N}\subset\Omega$, we approximate the density $f(\cdot,t_n)$ by an empirical histogram on a grid $\{z_m\}_{m=1}^{L_h}\subset\Omega$, where $L_h$ is the number of bins used. Denoting by $f^{n}$ the resulting discrete density, we define it through a standard histogram estimator.
Let $\{C_m\}_{m=1}^{L_h}$ be a partition of $\Omega$ into disjoint cells of volume $\lvert C_m\rvert$ and let $z_m\in C_m$ be the associated grid point. Then, using the notation $\omega_m=\lvert C_m\rvert$,
\[
f^{n}(z_m)
\;:=\;
\frac{1}{N\,\omega_m}
\sum_{i=1}^{N} \mathbf{1}_{C_m}(x_i^{n}),
\qquad m=1,\ldots,L_h,
\]
where $\mathbf{1}_{C_m}$ denotes the indicator function of the cell $C_m$. 
Thus, we use the same finite-dimensional ansatz \eqref{eq:Phat} and, for a temporal window $\ell\ge 1$, we define mean-field design matrices
${\bf\Theta}^{n}_{\textrm{mf}}\in\mathbb{R}^{dN\times N_b^P}$ and ${\bf\Lambda}^{n}_\textrm{mf}\in\mathbb{R}^{dN\times N_b^D}$ as follows.
For each snapshot $n$ and each agent state $x_i^{n}$, the $k$-th column of the drift matrix is given by the quadrature approximation
\begin{equation}\label{eq:coll_P}
\bigl({\bf\Theta}^{n}_\textrm{mf}\bigr)_{i,k}
\;:=\;
\sum_{m=1}^{L_h}
\phi_k\!\left(\lvert z_m-x_i^{n}\rvert\right)
\,(z_m-x_i^{n})\,f^{n}(z_m)\,\omega_m,
\qquad k=1,\ldots,N_b^P,
\end{equation}
where, to approximate the nonlocal mean-field integrals, we apply the quadrature rule induced by this histogram reconstruction.
Equivalently the diffusion matrix is defined by
\begin{equation}\label{eq:coll_D}
\bigl({\bf\Lambda}^{n}_\textrm{mf}\bigr)_{i,k}
\;:=\;
\sum_{m=1}^{L_h}
\psi_k\!\left(\lvert z_m-x_i^{n}\rvert\right)
\,f^{n}(z_m)\,\omega_m,
\qquad k=1,\ldots,N_b^D.
\end{equation}
Then, the drift and diffusion coefficients are obtained by solving the least-squares problems
\begin{align}
\label{eq:regression_mfP}
\varrho^{\mathrm{mf}}
&=
\arg\min_{\varrho\in\mathcal{A}_P}
\frac{1}{M_P}
\sum_{n=1}^{M_P}
\left\|
X^{n+\ell}-X^{n}
-\ell\,\Delta t\,{\bf\Theta}^{n}_\textrm{mf}\,\varrho
\right\|^{2},
\\
\label{eq:regression_mfD}
\zeta^{\textrm{mf}}
&=
\arg\min_{\zeta\in\mathcal{A}_D}
\frac{1}{M_D}
\sum_{n=1}^{M_D}
\left\|
\bigl(X^{n+\ell}-X^n\bigr)\odot\bigl(X^{n+\ell}-X^n\bigr)
- 2\ell\,\Delta t\,{\bf\Lambda}^{n}_\textrm{mf}\,\zeta
\right\|^{2},
\end{align}
where $X^{n}\in\mathbb{R}^{dN}$ is the agent configuration at time $t_n$ in vectorized form. The reconstructed kernels are finally given by
\[
\widehat P^{\textrm{mf}}(r):=\sum_{k=1}^{N_b^P}\varrho^{\mathrm{mf}}_k\,\phi_k(r),
\qquad
\bigl(\widehat D^{\textrm{mf}}\bigr)^{2}(r):=2\sum_{k=1}^{N_b^D}\zeta^{\mathrm{mf}}_k\,\psi_k(r).
\]
We report in Algorithm \ref{alg:2} the proposed methodology.
\begin{algorithm}
\caption{Kernel and diffusion reconstruction using the mean-field approach}\label{alg:2}
\begin{algorithmic}
	\State \texttt{Fix} $N_b^P,N_b^D$, $\ell$, $M=M_P=M_D$.
	\For{$n=1:M$}
	\State \texttt{Reconstruct the empirical density $f^{n}$ from the data $X^{n}$.}
	\State  \texttt{~Assemble} $\mathbf{\Theta}^{(n)}_\textrm{mf}$ by \eqref{eq:coll_P} and $\mathbf{\Lambda}^{(n)}_\textrm{mf}$ by \eqref{eq:coll_D}.
	\EndFor
	\State \texttt{Compute} $\varrho^{\mathrm{mf}}$ by solving \eqref{eq:regression_mfP}.
	\State \texttt{Compute} $\zeta^{\mathrm{mf}}$ by solving \eqref{eq:regression_mfD}.
	\State \Return $\widehat P^{\mathrm{mf}}$, $\widehat D^{\mathrm{mf}}$.
\end{algorithmic}
\end{algorithm}

\begin{remark}
The mean-field approach \eqref{eq:regression_mfP}-\eqref{eq:regression_mfD} approximates nonlocal interactions
through the empirical density, while still relying on
particle increments. Alternatively, identification could
be performed directly from the weak Fokker-Planck equation
\eqref{eq:kinetic_fokkerplanck}, following weak form
identification methods such as
WSINDy~\cite{messenger2022learning,lang2022learning}.
Inserting the kernel expansions \eqref{eq:Phat} gives
\begin{equation}
	\frac{d}{dt}\langle\varphi,f\rangle
	=
	\sum_{k=1}^{N_b^P}\varrho_k
	\mathcal G_k^P[\varphi,f]
	+
	\sum_{k=1}^{N_b^D}\zeta_k
	\mathcal G_k^D[\varphi,f],
\end{equation}
where, for scalar isotropic diffusion,
\begin{align*}
	\mathcal G_k^P[\varphi,f]
	&=\iint_{\mathbb R^{2d}}
	\nabla\varphi(x)\cdot(y-x)\,
	\phi_k(|x-y|)f(x)f(y)\,dx\,dy,\\
	\mathcal G_k^D[\varphi,f]
	&=\iint_{\mathbb R^{2d}}
	\Delta\varphi(x)\,
	\psi_k(|x-y|)f(x)f(y)\,dx\,dy.
\end{align*}
Given density snapshots $\{f^n\}_{n=0}^M$ and
test functions $\{\varphi_j\}_{j=1}^J$, we define
the regression matrices through time quadrature,
for instance using a left-endpoint rule:
\begin{equation}
	\begin{aligned}
		(G_P)_{(n,j),k}
		&=\ell\Delta t\,\mathcal G_k^P[\varphi_j,f^n],\qquad
		(G_D)_{(n,j),k}
		&=\ell\Delta t\,\mathcal G_k^D[\varphi_j,f^n].
	\end{aligned}
\end{equation}
The corresponding density-based identification problem is
\begin{equation}
	\min_{\substack{\varrho\in\mathcal A_P\\
			\zeta\in\mathcal A_D}}
	\sum_{n=0}^{M-\ell}\sum_{j=1}^J
	\left|
	\left\langle\varphi_j,f^{n+\ell}-f^n\right\rangle
	-\sum_{k=1}^{N_b^P}(G_P)_{(n,j),k}\varrho_k
	-\sum_{k=1}^{N_b^D}(G_D)_{(n,j),k}\zeta_k
	\right|^2.
\end{equation}
Compared with trajectory-based regression, this formulation
does not require particle tracking and may benefit from
population averaging. However, stable identification
requires sufficiently informative density observations
and test functions, ensuring suitable coercivity of the
regression matrix $G=[G_P,G_D]$, see
\cite{lang2023identifiability,LMT22}.
Hence, unlike the trajectory-based approach \eqref{eq:regression_mfP}-\eqref{eq:regression_mfD}, where drift and
diffusion are identified separately through the first
and second conditional moments of particle increments,
the weak density-based formulation combines both contributions
within a single regression problem. This joint identification
requires sufficiently informative observations to distinguish
the effects of the interaction and diffusion kernels.
\end{remark}

\begin{remark}\label{rem:dimension}
Since the kernels are radial, the basis functions are
defined on the one-dimensional domain of the distance
variable $r$. Thus, for a fixed radial discretization,
the number of unknowns $N_b^P+N_b^D$ is independent
of the state dimension $d$, avoiding the exponential
growth associated with general multivariate bases. When the diffusion acts component-wise, as in Test~4, the number of unknowns is $N_b^P+d\,N_b^D$ and grows linearly with $d$. The dimension enters only through the assembly of the regression problems, in which the design matrices of the $M$ snapshots are stacked into a single least-squares problem with $M\,dN$ rows and $N_b^P$ unknowns, and analogously for the diffusion.

In Algorithm~\ref{alg:1}, each entry of ${\bf\Theta}^{(n,h)}$ in \eqref{eq:regression_kern_batch} requires the distance between an agent and one of its sampled partners, at cost $\mathcal{O}(d)$, and each row has at most two nonzero entries, since piecewise linear hat functions have at most two nonzero values at any distance $r$. Hence the assembly requires $\mathcal{O}(M N_p\,dN)$ operations and the least-squares problem is solved at cost $\mathcal{O}\bigl(M\,dN\,(N_b^P)^2\bigr)$; both costs are linear in $d$, and no spatial grid is involved. In Algorithm~\ref{alg:2}, each row of the design matrices in \eqref{eq:coll_P}--\eqref{eq:coll_D} is a sum over the cells of the histogram grid, and each term requires the distance between an agent and a cell center, at cost $\mathcal{O}(d)$. A full grid on $\Omega$ with cells of side $\Delta_C$ has $L_h=\mathcal{O}(\Delta_C^{-d})$ cells, so that the assembly requires $\mathcal{O}(M\,dN\,L_h)$ operations, which grows exponentially with $d$. This is the only step of the method affected by the curse of dimensionality, and it can be avoided. Indeed, with $\omega_m=\lvert C_m\rvert$, \eqref{eq:coll_P} reduces to
\[
\bigl({\bf\Theta}^{n}_{\textrm{mf}}\bigr)_{i,k}=\frac1N\sum_{j=1}^N\phi_k\bigl(\lvert z_{m(j)}-x_i^n\rvert\bigr)\,(z_{m(j)}-x_i^n),
\]
where $z_{m(j)}$ is the center of the cell containing $x_j^n$; that is, each agent is moved to the center of its cell, and only the (at most) $\min(L_h,N)$ occupied cells contribute. Restricting the sum to these cells reduces the assembly to $\mathcal{O}\bigl(M\,dN\min(L_h,N)\bigr)$ operations, at most $\mathcal{O}(M\,dN^2)$, while the least-squares problem is solved at cost $\mathcal{O}\bigl(M\,dN\,(N_b^P)^2\bigr)$, as for Algorithm~\ref{alg:1}.

Finally, for non-radial kernels $P(x,y)$, $D(x,y)$, the basis must instead be defined on $\mathbb{R}^{2d}$, and tensor-product bases such as the ones used here grow exponentially with $d$; in that case, neural-network representations are a possible alternative (see Section~\ref{sec:conclusions}).
\end{remark}

\section{A priori error estimates for the reconstructed trajectories}\label{sec:stima}
Let us now consider the reconstructed functions $\widehat P$ and $\widehat D$ obtained through either Algorithms \ref{alg:1} or \ref{alg:2}. In this section, we study the discrepancy between the  trajectories generated by the original dynamics and those obtained using the reconstructed quantities. 
The predicted error shown in Theorem \ref{thm:1} further validates our method, since we proved boundedness of the reconstructed trajectories on average, {and, under the same noise realization, it is consistent with exact reconstruction.}
To this end, we consider the original dynamics
\begin{equation}\label{eq:og_dynamics}
X^{n+1}
=
X^n
+ \Delta t\, {\mathbf P}(R^n) \odot \bigl( \mathsf{S}^n  X^n -  X^n \bigr)
+ \sqrt{\Delta t}\,{\mathbf D}(R^n) \odot \Xi^n,
\end{equation}
where $ R^n = \|{\mathsf{S}}^n X^n - X^n \|_2$, and the fully reconstructed dynamics
\begin{equation}\label{eq:reconstructed_dynamics}
\widetilde X^{n+1}
=
\widetilde X^n
+ \Delta t\, \widehat{\mathbf P}\bigl(\widetilde R^n\bigr)\odot \bigl( \widehat{\mathsf{S}}^n \widetilde X^n - \widetilde X^n \bigr)
+ \sqrt{\Delta t}\,\widehat{\mathbf D}\bigl(\widetilde R^n\bigr) \odot \widetilde \Xi^n,
\end{equation}
where $\widetilde R^n = \|\widehat{\mathsf{S}}^n \widetilde X^n - \widetilde X^n \|_2$. Here $\widehat{\mathsf S}^n$ denotes the reconstructed interaction matrix, 
reflecting the fact that we assume no a priori knowledge of the true interaction operator $\mathsf S^n$. 
Here, $(\Xi^n)_{n\ge0}$ and $(\widetilde\Xi^n)_{n\ge0}$ denote two sequences of independent standard Gaussian random vectors representing the normalized Brownian increments driving the original and reconstructed systems, respectively. 

In order to estimate the total reconstruction error, we additionally  introduce the intermediate trajectory
\begin{equation}\label{eq:xbar}
\bar X^{n+1}
=
\bar X^n
+ \Delta t\, \widehat{\mathbf P}\bigl(\bar R^n\bigr)\odot \bigl( {\mathsf{S}}^n  \bar X^n -  \bar X^n \bigr)
+ \sqrt{\Delta t}\,\widehat{\mathbf D}\bigl(\bar R^n\bigr) \odot \Xi^n,
\end{equation}
with $\bar R^n = \| \mathsf{S}^n \bar X^n - \bar X^n\|_2$, where we consider the dynamics obtained by using the exact interaction 
matrix $\mathsf S^n$, the same Brownian increments $\Xi^n$ as in \eqref{eq:og_dynamics}, while replacing the kernel functions with their reconstructed counterparts.

Hence, we can compute the total reconstruction error into two separate contributions: one 
due to the approximation of the kernels, and the
other due to the reconstruction of the interaction matrix $\mathsf S^n$. Therefore, we define
\begin{equation}\label{eq:err_e}
e^n := X^n-\widetilde X^n, \quad \bar e^n := X^n - \bar X^n, \quad \widetilde e^n = \bar X^n - \widetilde X^n,
\end{equation}
so that by triangular inequality we have
\begin{equation}\label{eq:triang}
\mathbb{E}\left[\|e^{n+1}\|_2^2\right]\le2\mathbb{E}\left[\| \bar e^{n+1} \|^2_2\right] + 2\mathbb{E}\left[\| \widetilde e^{n+1} \|^2_2\right].
\end{equation}
In what follows we will provide an estimate of these quantities, and we recall the following elementary bounds which are useful for later analysis.
We have that the following theorem holds true.
\begin{theorem}\label{thm:1}
Suppose that $\Delta t \leq 1$ and that the trajectories of \eqref{eq:og_dynamics}--\eqref{eq:reconstructed_dynamics} are generated using the same Brownian increments, namely $\Xi^n\equiv \widetilde{\Xi}^n$. Assume that $X^n,\widetilde X^n, \bar X^n \in [-L,L]^{N_d}$ for all $n=0,1,\ldots,M$ with $N_d:=dN$. In addition, we assume that $\mathbf{P},\,\mathbf{D},\,\widehat{\mathbf{P}}$, and $\widehat{\mathbf{D}}$ are Lipschitz continuous with Lipschitz constants $L_P,\,L_D,\,L_{\widehat P}$, and $L_{\widehat D}$, respectively.
Then, defining $R:=2L\sqrt{d}$ and
\begin{align}\label{eq:bounds}
	&\delta_P := \sup_{r\in[0,R]} \bigl|P(r)-\widehat P(r)\bigr|,\quad
	\delta_D := \sup_{r\in[0,R]} \bigl|D(r)-\widehat D(r)\bigr|,\cr
	&P_{\max} := \|\mathbf P\|_\infty,\,\,\widehat{P}_{\max} := \|\widehat{\mathbf P} \|_\infty,\,\, D_{\max} := \|\mathbf D\|_\infty,\,\,\widehat{D}_{\max} := \| \widehat{\mathbf D} \|_\infty
\end{align}
and $\eta_{S}:=\max_{0\leq n \leq M-1}\|\mathsf{S}^n-\hat{\mathsf{S}}^n\|_2\leq 2$,
the following error bound holds
\[\max_{0\leq n\leq M}\mathbb{E}\left[\| e^{n}\|^2_2\right] \leq (\delta_P^2+\delta_D^2+\eta_S^2)\widehat{C}_2\left(\exp\left\{\widehat{C}_1T \right\}-1\right),\]
with $\widehat{C}_1,\widehat {C}_2$ positive depending on all the Lipschitz constants, $P_{\max}$, $D_{\max}$, $L$, and $N_d$.
\end{theorem}
The proof is reported in Appendix \ref{app:conti} and relies on standard Gr\"onwall estimates.
\begin{remark}
The error estimate in Theorem~\ref{thm:1} shows stability of the reconstructed trajectories in mean square. In particular, the assumption $\Xi^n\equiv\widetilde\Xi^n$ is essential to compare the two trajectories pathwise and recover zero error in the ideal case
$\widehat{\mathbf P}=\mathbf P, \widehat{\mathbf D}=\mathbf D$, and $\widehat{\mathsf S}^n=\mathsf S^n.$ If the two systems are instead driven by independent realizations of the Brownian increments, i.e.\ $\Xi^n\neq\widetilde{\Xi}^n$, an additional error contribution appears, namely
\[
\max_{0\le n\le M}\mathbb E\left[\|e^n\|_2^2\right]
\le
\bigl(\delta_P^2+\delta_D^2+\eta_S^2+\widehat{\sigma}^2\bigr)\widehat{C}_2\bigl(e^{\widehat{C}_1T}-1\bigr),
\]
where the additional term $\widehat{\sigma}^2$ accounts for the mismatch between the driving noises. Consequently, the error remains nonzero even under exact reconstruction of the kernels and interaction matrices.

A complementary direction to the trajectory-based stability
analysis of Theorem~\ref{thm:1} is to investigate approximation
errors at the level of probability distributions.
In this setting, relative-entropy methods, as developed
in~\cite{huang2025mean} for random-batch particle systems,
could provide a framework for studying the distributional
consistency of the reconstruction approaches in
Section~\ref{sec:discovery}, particularly the approximation
of nonlocal interaction operators through empirical densities.
Unlike mean square trajectory estimates, a distributional approach avoids pathwise coupling and direct comparison of interaction matrices. However, extending the relative entropy
analysis of~\cite{huang2025mean} to possibly degenerate,
state-dependent or nonlocal diffusion requires additional
regularity assumptions.
\end{remark}

\begin{remark}\label{stima_inversa}
We finally remark that an estimate can be retrieve also for the inverse reconstruction error. Indeed,
consider the unconstrained drift regression
\eqref{eq:regression_kern_batch}, and 
\[
P_h(r)=\sum_{k=1}^{N_b^P}\bar\varrho_k\phi_k(r),
\qquad
\widehat P(r)=\sum_{k=1}^{N_b^P}\varrho_k^*\phi_k(r),
\]
where $P_h$ is the piecewise linear interpolant of $P$,
$\varrho^*$ denotes the estimated coefficients, and
$\varepsilon_h=\|P-P_h\|_{L^\infty(0,R)}$.
Moreover, assuming  the following coercivity condition
\[
\lambda_{\min}\left(
\frac{1}{M_PdN}
\sum_{n=1}^{M_P}
(\hat{\bf\Theta}^{n})^\top
\hat{\bf\Theta}^{n}
\right)\ge c>0
\]
almost surely, with a deterministic constant $c$. Then, the following estimate holds:
\begin{equation}\label{eq:mainbound}
	\mathbb{E}\left[\|\varrho^*-\bar\varrho\|_2\right]
	\le \frac1c\left(
	4L^2\varepsilon_h
	+\left(
	\frac{4L^2P_{\max}}{\sqrt{N_p}}
	+\frac{8L^2P_{\max}}{\sqrt{M_P}}
	\right)\chi_{\rm RB}
	+\frac{2LD_{\max}}{\sqrt{dNM_P\Delta t}}
	\right),
\end{equation}
where $\chi_{\rm RB}\equiv0$ for known interaction pairs
and $\chi_{\rm RB}\equiv1$ for random-batch reconstruction.
Since the piecewise linear basis functions are nonnegative
and sum to one at every point of $[0,R]$, taking
expectations yields
\[
\mathbb E\left[\|P-\widehat P\|_{L^\infty(0,R)}\right]
\le \varepsilon_h+
\mathbb E\left[\|\bar\varrho-\varrho^*\|_2\right].
\]
Thus, together with \eqref{eq:mainbound}, this provides
a bound on the expected uniform reconstruction error,
accounting for kernel approximation, uncertainty in
interaction pairings, and stochastic fluctuations.
The detailed proof is given in Appendix~\ref{app:conti2}.
\end{remark}

\section{Numerical experiments}\label{sec:num}
In this section we present numerical experiments assessing the performance of the reconstruction strategies proposed in Algorithms~\ref{alg:1} and~\ref{alg:2}. In the first approach, the pairing matrices are assumed to be unobserved; therefore, at each time step we approximate the missing interaction information by sampling $N_p \ll N$ random interacting pairs and use these samples to reconstruct the interaction kernel. In the second approach, random sampling is replaced by a mean-field reconstruction based on density approximations over the computational cells, leading to a local quadrature approximation of the design matrices.

For the one-dimensional tests, the reconstruction quality is evaluated using the error indicators \eqref{eq:errorsk} and \eqref{eq:w1}. Additional indicators for the two-dimensional examples will be introduced in Section~\ref{sec:2dim}. These examples extend those presented in Section~\ref{sec:exS}, since here the particle interactions are assumed unknown. In contrast, knowledge of the interaction matrices $S$ implies full access to the dataset at all time instances, e.g. $\ell = 1$ in Algorithms~\ref{alg:1} and~\ref{alg:2}. In the following we also consider the case $\ell > 1$. The numerical results show the same order of accuracy as those reported in \cite{LMT21}. We emphasize that our method learns each kernel using only a single trajectory.

\subsection{One dimensional tests}
In this Section, we will focus on four tests with synthetic data. The algorithms used for the solution are the ones described in Algorithm \ref{alg:1} and \ref{alg:2}. We choose three different radial interaction kernels, while for the diffusion we made two different choices: in the first three simulations the diffusion function is only depending on the status of the agent and is given by $D(x) = (1-x^2)^2/2$, the fourth simulation is instead in the more general setting in which the diffusion is nonlocal. The dataset consists of $N=10^5$ particles and $201$ time frames, and we set $\Delta t = 0.05$ among two consecutive time frames in the data. We focus on three different settings $\mathscr{S}$ to recover the nonlocal interaction kernel. In Setting $\ScrS=1$ we select $M_P=100$ and $\ell=1$, in $\ScrS=2$ $M_P=50$ and $\ell=2$, while in $\ScrS=3$ we have $M_P = 25$ and $\ell = 4$. When the diffusion is not radial, we set $M_D = 10$ and $\ell=1$, otherwise we maintain the same Setting as for the kernel learning.
Note that the training is not done over the whole time horizon $[0,T]$, but only until time $t=T/2$ for the particles. This is also to test the method, specifically, the discovered problem outside the training regime.

\begin{figure}[htbp]
\centering
\includegraphics[width=0.45\linewidth]{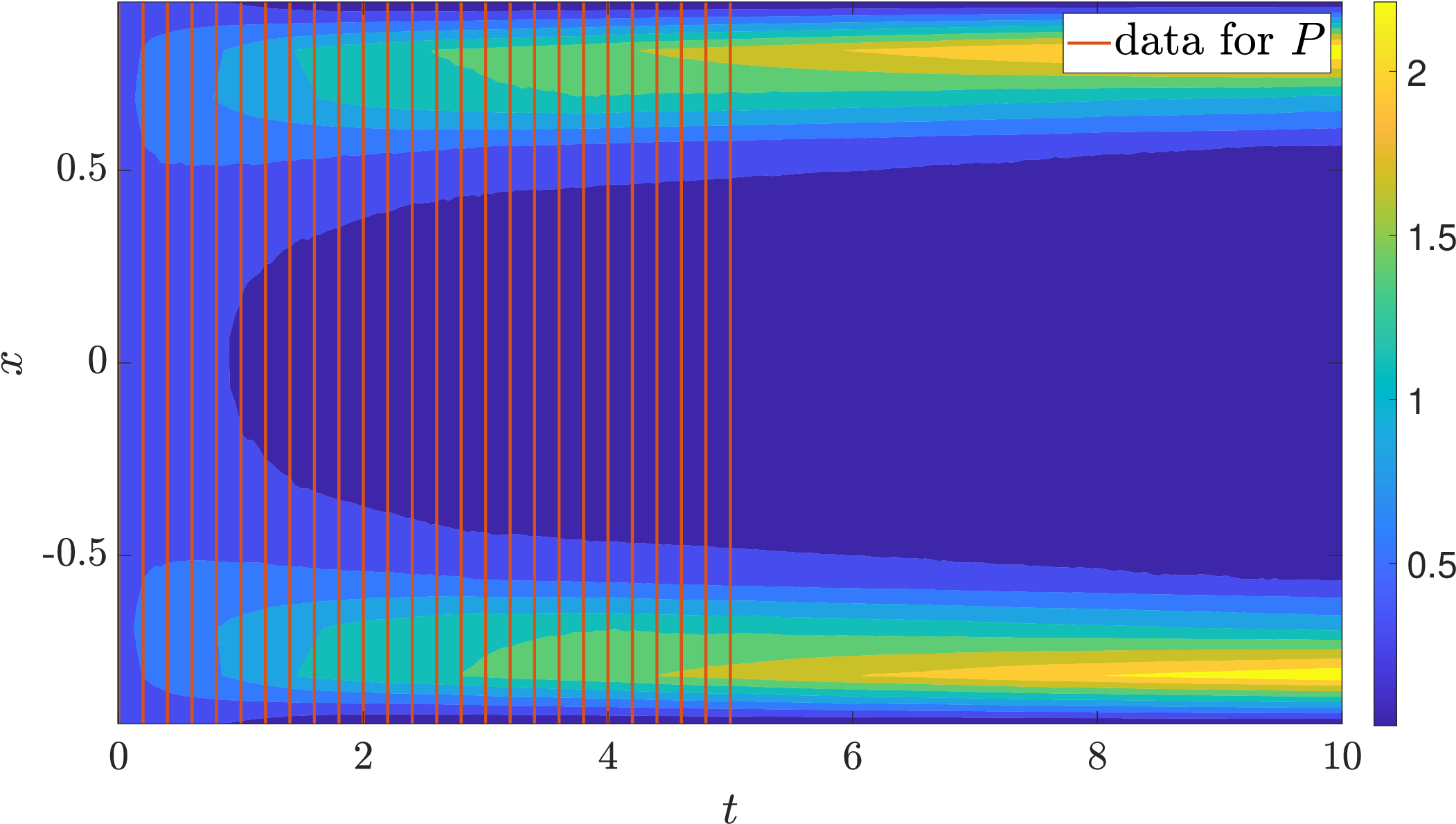}\quad\includegraphics[width=0.45\linewidth]{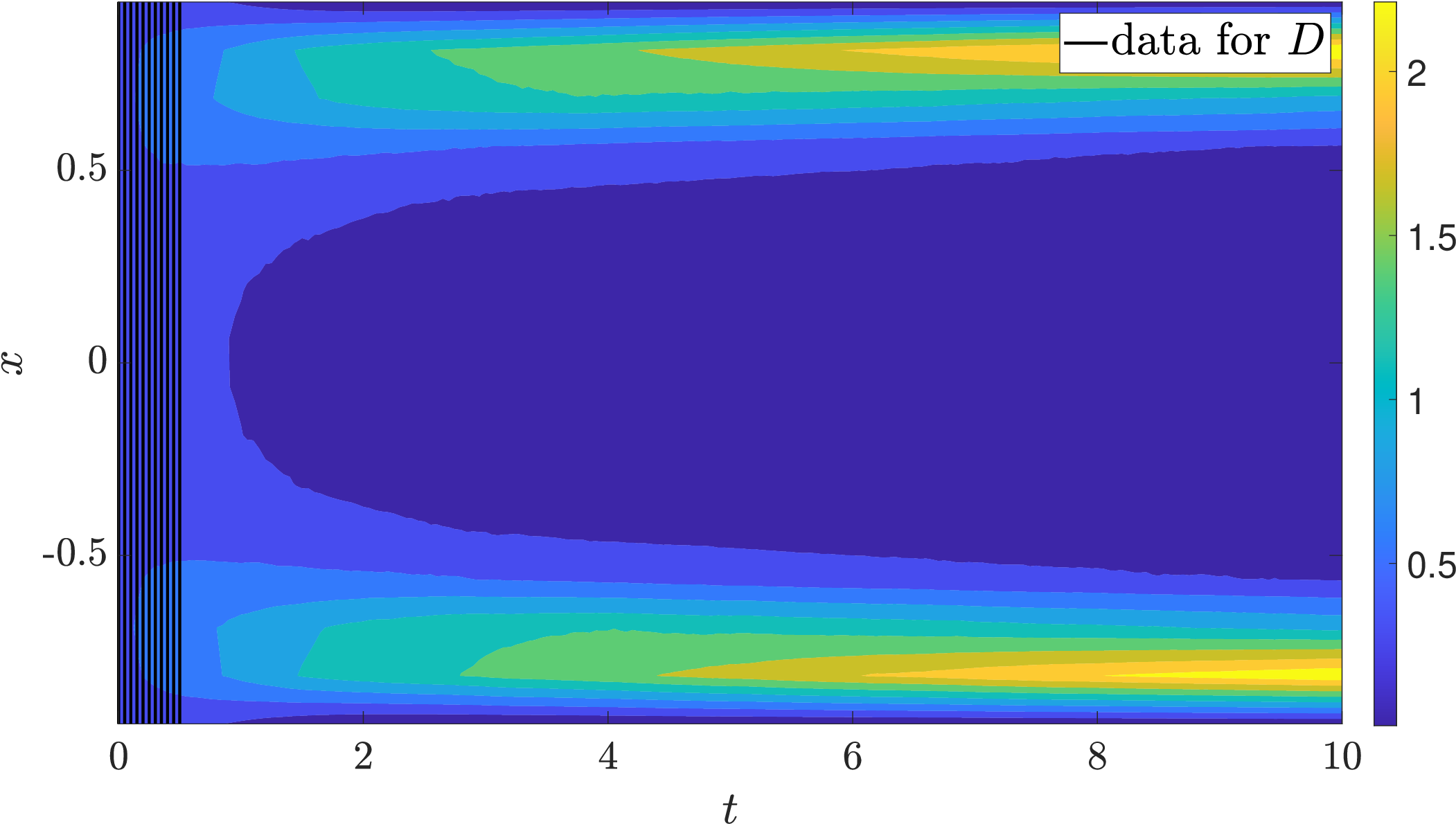}\
\caption{Test 1: Data used for the reconstruction of the kernel when $\Delta = 4$ and $M_P=25$ (left) and data used for the reconstruction of the diffusion with $\ell = 1$ and $M_D=10$.}
\label{fig:data_kernel1}
\end{figure}

\paragraph{Test 1: Bounded Confidence model}
The first simulation concerns the bounded confidence model, a classical framework for describing opinion dynamics in multi-agent systems. In this setting the variable $x\in[-1,1]$ can be interpreted as the opinion on a matter, and individuals adjust their own convictions through interactions with others only if their views are sufficiently close, i.e if the distance between the opinions is below a given threshold $\tau$. Formally, this can be modeled with a radial kernel $P(x,y) = P(\lvert x - y\rvert) = P(r) = \chi(r<\tau)$, where we recall the notation $r = \vert x-y\rvert$ to indicate the distance between the opinions of the interacting agents and use $\chi$ for the indicator function. Depending on the values of the parameter $\tau$, this model captures the emergence of opinion clustering and polarization in societies: small values of $\tau$ lead to multiple distinct opinion groups, while larger values, instead, promote global consensus. The presence of the diffusion in the interaction smooths out the resulting density of the agents, that otherwise would reach a steady state which is a sum of Dirac deltas. In our simulation, the initial condition is a uniform distribution and we set $\tau=0.5$, so the kernel to learn is $P(r) = \chi(r<0.5)$, leading to a polarized configuration at the final observed time. For what concerns the optimization problem, we set $\kappa_P =-1$ and $P_0=1$ in \eqref{eq:AP}, whereas in \eqref{eq:AD} we set $\kappa_D=0, \overline{\zeta}_0=0 = \overline{\zeta}_{N_b^D.}$ In other words, we require that the kernel is non-increasing and that $P(0)=1$, while the diffusion is nonegative with $D(\pm 1)=0$.
Figure \ref{fig:data_kernel1} shows the data used for the kernel reconstruction in the case of $\ell = 4$ and $M_P = 25$ (left) and with $\ell = 1$ and $M_D=10$ for the diffusion (right). We clearly observe that a larger $\ell$ provides data less frequently.

\begin{figure}[htbp]
\centering
\includegraphics[height=5.4cm]{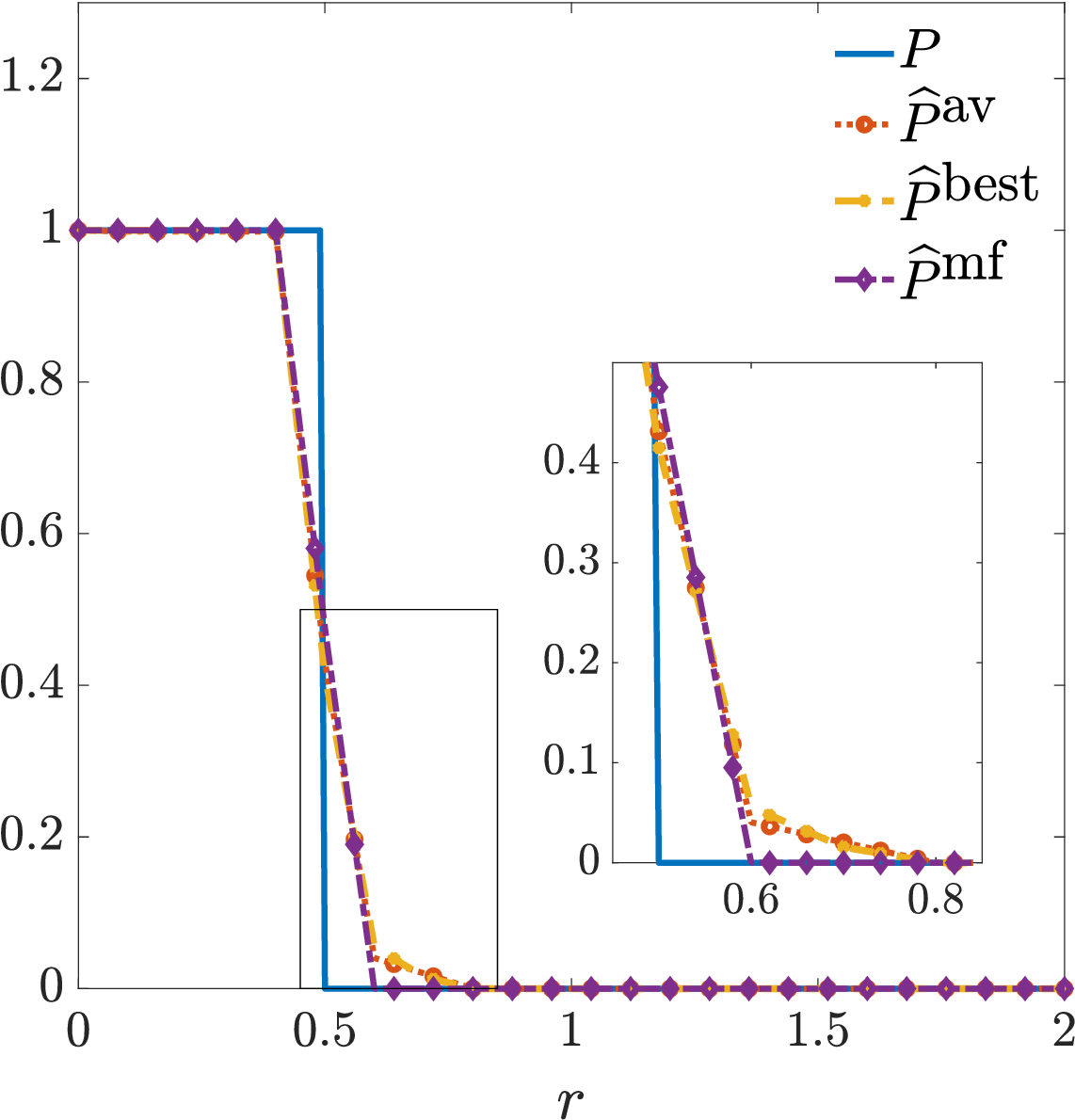}\quad\includegraphics[height=5.45cm]{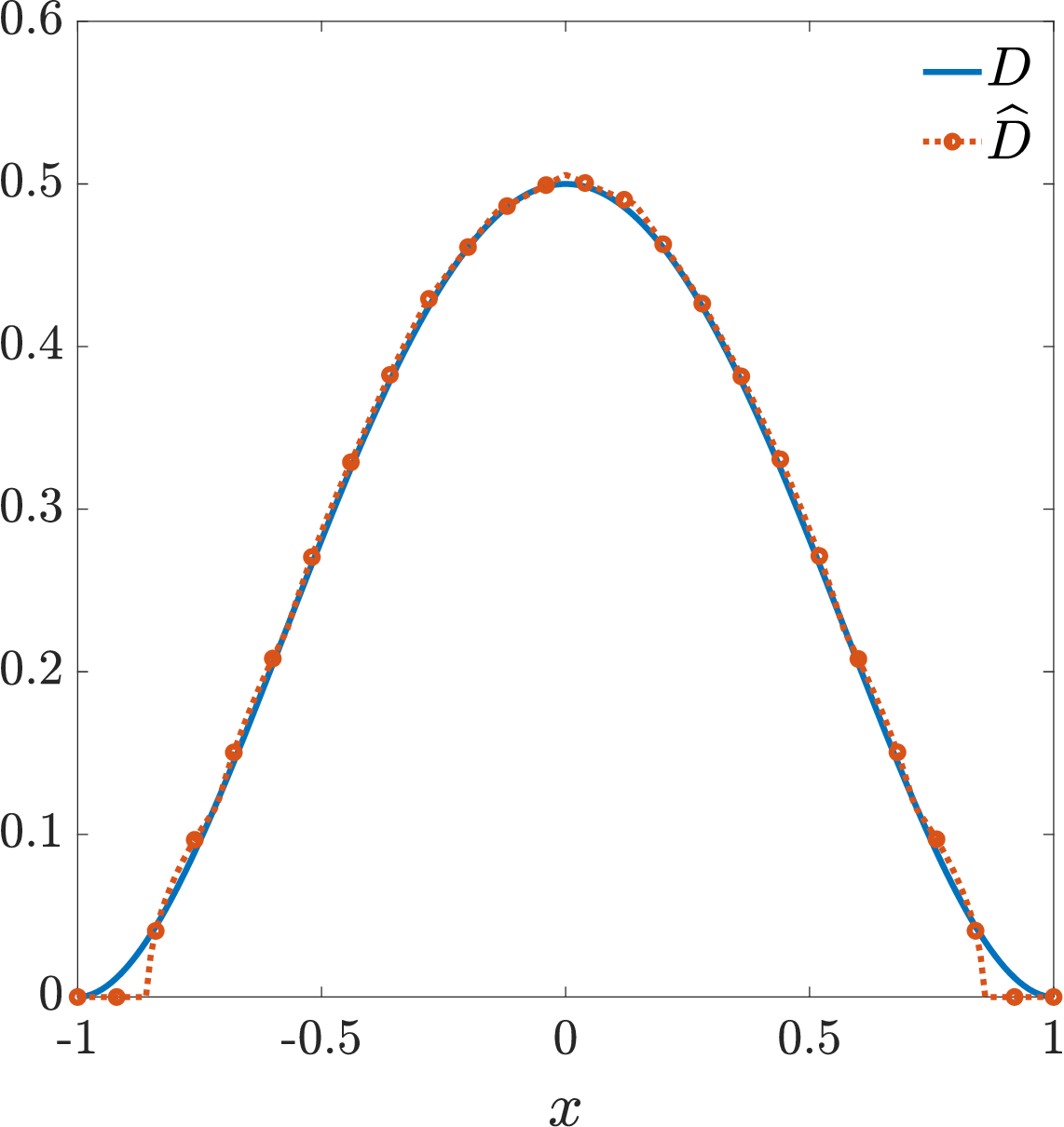}\quad\includegraphics[height=5.45cm]{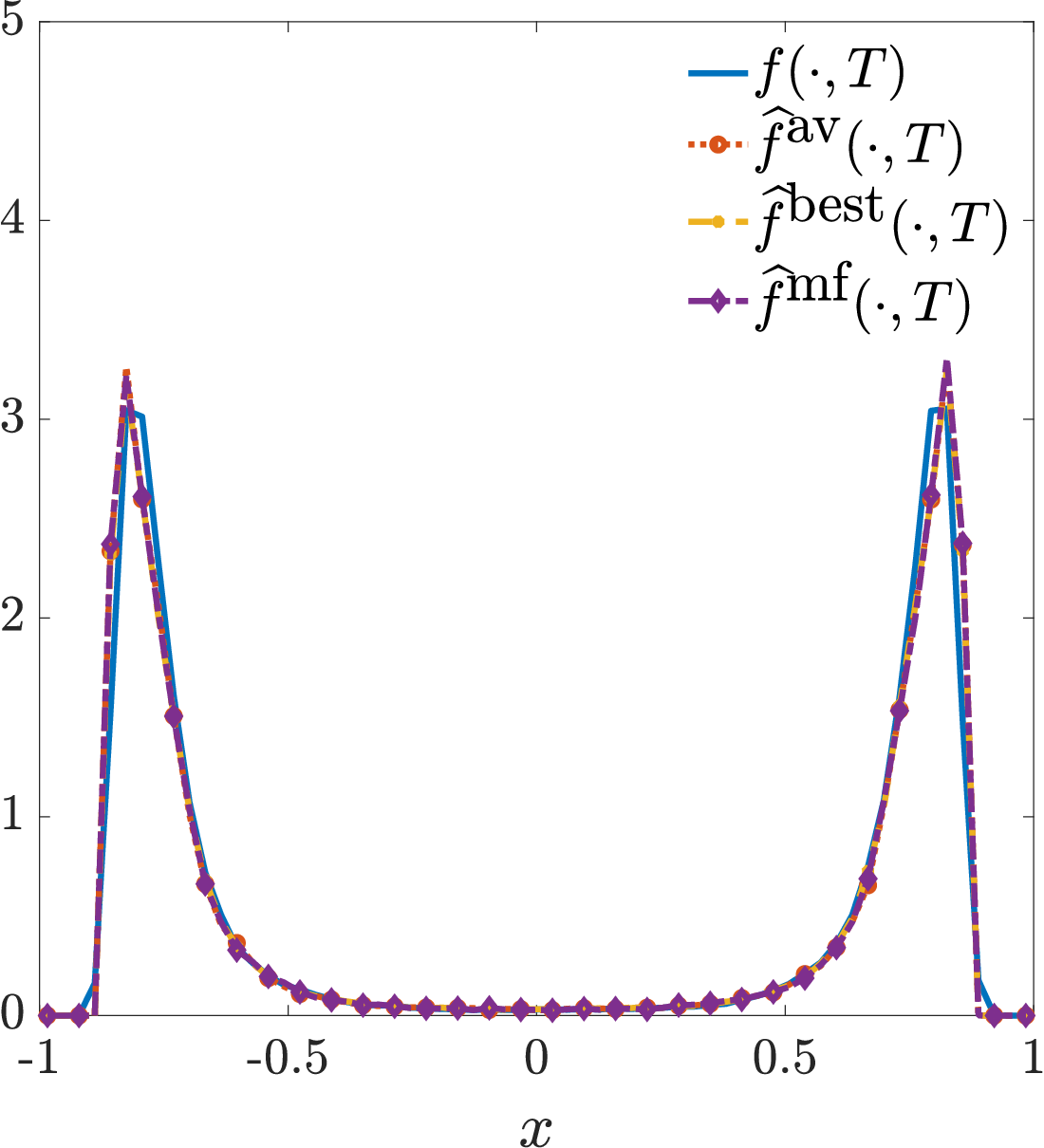}
\caption{Test 1. Reconstruction of the kernel $P=\chi(r<0.5)$ using $21$ basis functions, $\ell = 2$ and $M_P=50$ using approach \eqref{eq:weights1} (averaging) and \eqref{eq:weights2} (best) in Alg. \ref{alg:1} and also the mean field approach described in Alg. \ref{alg:2}. (left), reconstruction of the diffusion function $D(x)$ using $15$ basis functions $\ell=1$ and $M_D = 10$ (center), comparison between the data and the reconstructed density at the final time (right).}
\label{fig:av_bc_new}
\end{figure}

In the left panel of Figure \ref{fig:av_bc_new}, we show a comparison between the exact kernel, the reconstruction made with the weights as in \eqref{eq:weights1} and the one using the weights in \eqref{eq:weights2} as well as the mean field approach. The comparison between the exact diffusion function and the reconstructed one is shown in the middle panel of Figure \ref{fig:av_bc_new}, whereas a comparison between the profile of the density of the data at final time and the reconstructed densities obtained using the two different approximations of the kernel can be found in the right panel. 
The results are qualitatively very similar for all the methods. We can see that both methods captures the behavior of the kernel with a little difference close to the non smooth part in $r=0.5$ (see the zoom in the left panel of Figure \ref{fig:av_bc_new}). 

Quantitive results are shown in Table \ref{tab:tabella_rb_bc}. We can see that all the methods, including the variations in Alg. \ref{alg:1} using either \eqref{eq:weights1} or \eqref{eq:weights2} and Alg. \ref{alg:2} provide exactly the same order of the error.

\begin{table}[htbp]
\small
\centering
\begin{tabular}{|c|c|c|c|c|}
	\hline
	\multicolumn{5}{|c|}{Averaging method (Alg. \ref{alg:1})}\\
	\hline
	$\mathscr{S}$ & $E^1_P$ & $E^\infty_P$& $E^\textrm{ave}_f$ & $E^\textrm{fin}_f$ \\
	\hline
	$1$ & $1.61\cdot 10^{-1}$ & $5.35\cdot 10^{-1}$ & $7.36\cdot 10^{-3}$ & $8.89\cdot 10^{-3}$ \\
	$2$ & $1.12\cdot 10^{-1}$ & $5.12\cdot 10^{-1}$ & $6.68\cdot 10^{-3}$ & $9.20\cdot 10^{-3}$  \\
	$3$ & $1.61\cdot 10^{-1}$ & $5.13\cdot 10^{-1}$ & $6.92\cdot 10^{-3}$ & $1.01\cdot 10^{-2}$ \\
	\hline
	\multicolumn{5}{|c|}{Best result method (Alg. \ref{alg:1})}\\
	\hline
	$\mathscr{S}$ & $E^1_P$ & $E^\infty_P$& $E^\textrm{ave}_f$ & $E^\textrm{fin}_f$ \\
	\hline
	$1$ & $1.64\cdot 10^{-1}$ & $5.12\cdot 10^{-1}$ & $9.60\cdot 10^{-3}$ & $1.45\cdot 10^{-2}$ \\
	$2$ & $1.13\cdot 10^{-1}$ & $5.27\cdot 10^{-1}$ & $7.78\cdot 10^{-3}$ & $1.11\cdot 10^{-2}$  \\
	$3$ & $1.62\cdot 10^{-1}$ & $4.95\cdot 10^{-1}$ & $7.07\cdot 10^{-3}$ & $1.01\cdot 10^{-2}$ \\
	\hline
	\multicolumn{5}{|c|}{Mean field approach (Alg. \ref{alg:2})}\\
	\hline
	$\mathscr{S}$ & $E^1_P$ & $E^\infty_P$& $E^\textrm{ave}_f$ & $E^\textrm{fin}_f$ \\
	\hline
	$1$ & $1.01\cdot 10^{-1}$ & $5.28\cdot 10^{-1}$ & $6.39\cdot 10^{-3}$ & $1.02\cdot 10^{-2}$ \\
	$2$ & $1.14\cdot 10^{-1}$ & $5.31\cdot 10^{-1}$ & $6.69\cdot 10^{-3}$ & $1.00\cdot 10^{-2}$  \\
	$3$ & $1.54\cdot 10^{-1}$ & $5.19\cdot 10^{-1}$ & $7.07\cdot 10^{-3}$ & $9.45\cdot 10^{-3}$ \\
	\hline
\end{tabular}
\caption{Test 1. Errors for the identification of kernel $P=\chi(r<0.5)$ using $N_b^P=21$ basis functions. }
\label{tab:tabella_rb_bc}
\end{table}

\paragraph{Test 2: Attraction-repulsion model}

The second simulation consists in reconstructing the kernel $P(r) = ((0.1+r)^2 - 0.05(0.1+r)^{-2})/5$ and the diffusion function $D(x) = (1-x^2)^2/2$. Such kernel models attraction-repulsion interactions, since it combines short-range repulsion and long-range attraction between agents. This mechanism captures the balance between the tendency of individuals to maintain personal space and their inclination to align or aggregate, as in the phenomenon of flocking, but can also be interpreted in the framework of opinion dynamics. In fact, it can be seen as a form of coexistence of consensus and differentiating behaviors: agents close in opinion repel to preserve individuality, while those moderately distant are drawn together, reflecting realistic mechanisms of social interaction. The patterns resulting in the density of the agents depend on the balance between the attraction and the repulsion terms. In the optimization problem, we set $\kappa_P = 1$ and $P_0=-1$ in \eqref{eq:AP}, whereas in \eqref{eq:AD} we set $\kappa_D=0, \overline{\zeta}_0=0 = \overline{\zeta}_{N_b^D.}$ In other words, we require that the kernel is non-decreasing and $P(0)=-1$ while the diffusion is nonegative with $D(\pm 1)=0$.

\begin{figure}[htbp]
\centering
\includegraphics[width=0.45\linewidth]{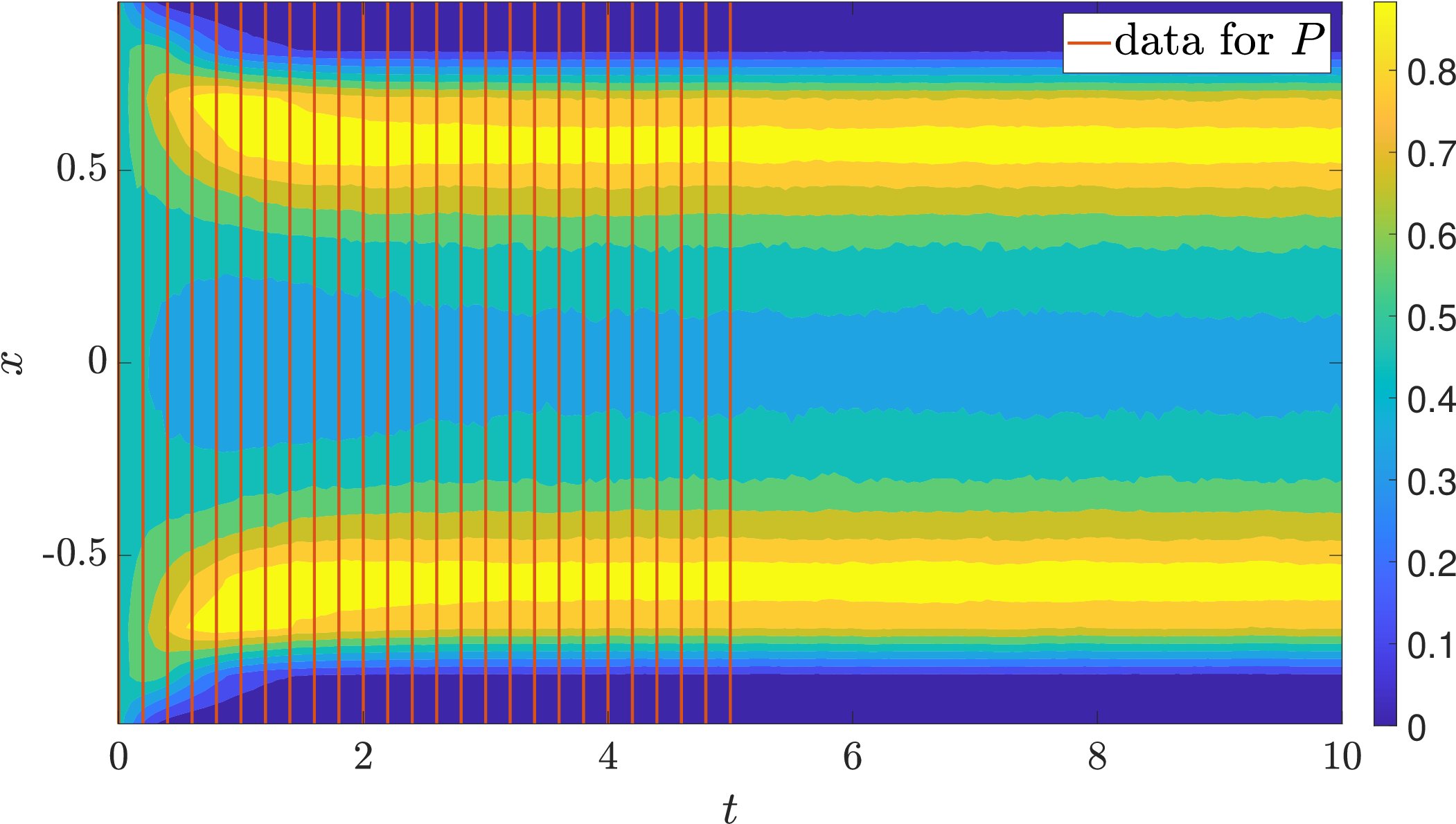}\quad\includegraphics[width=0.45\linewidth]{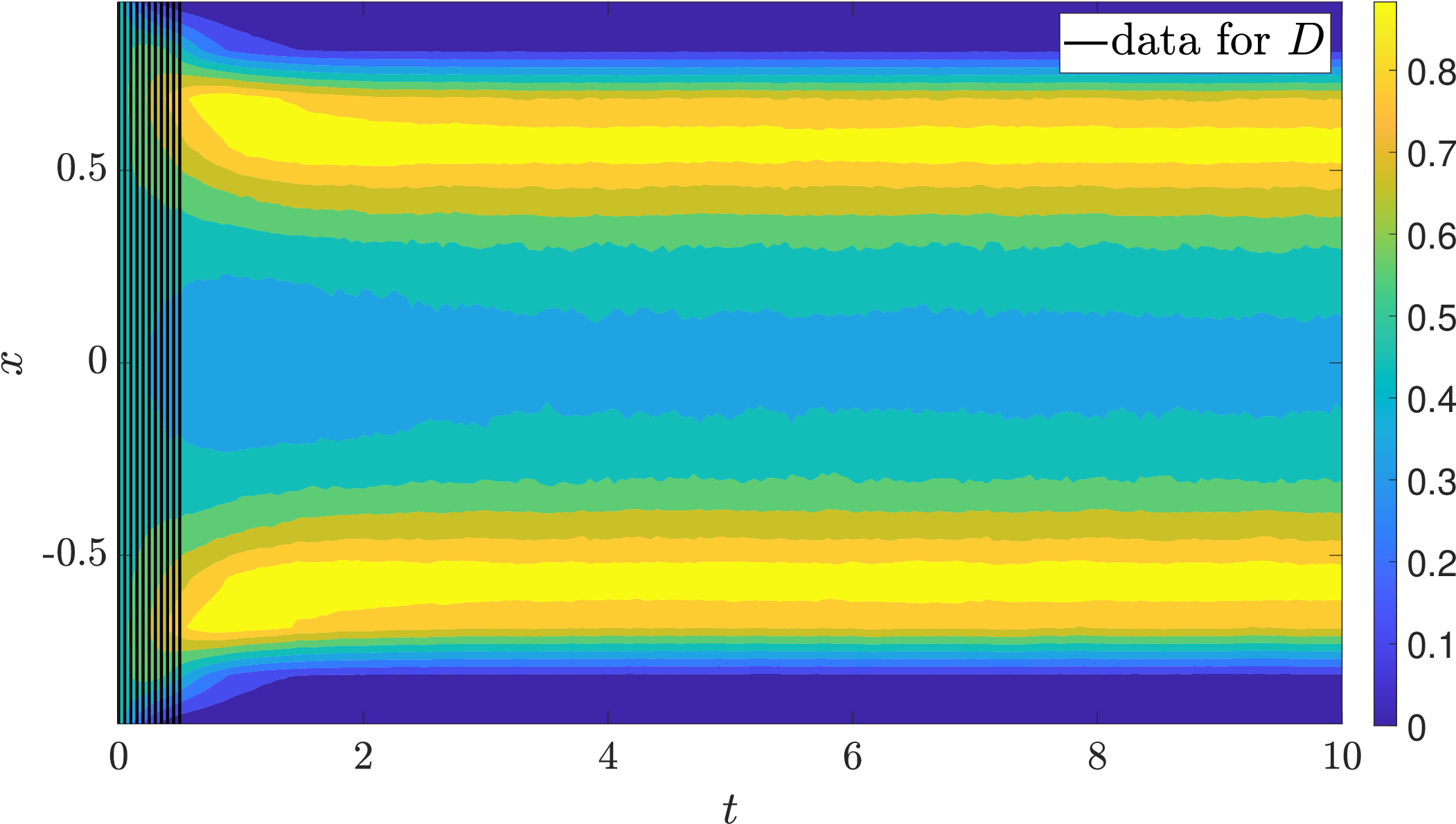}\
\caption{Test 2. Data used for the reconstruction of the kernel when $\ell = 4$ and $M_P=25$ (left) and data used for the reconstruction of the diffusion with $\ell = 1$ and $M_D=10$.}
\label{fig:data_kernel3}
\end{figure}

\begin{figure}[htbp]
\centering
\includegraphics[height=5.4cm]{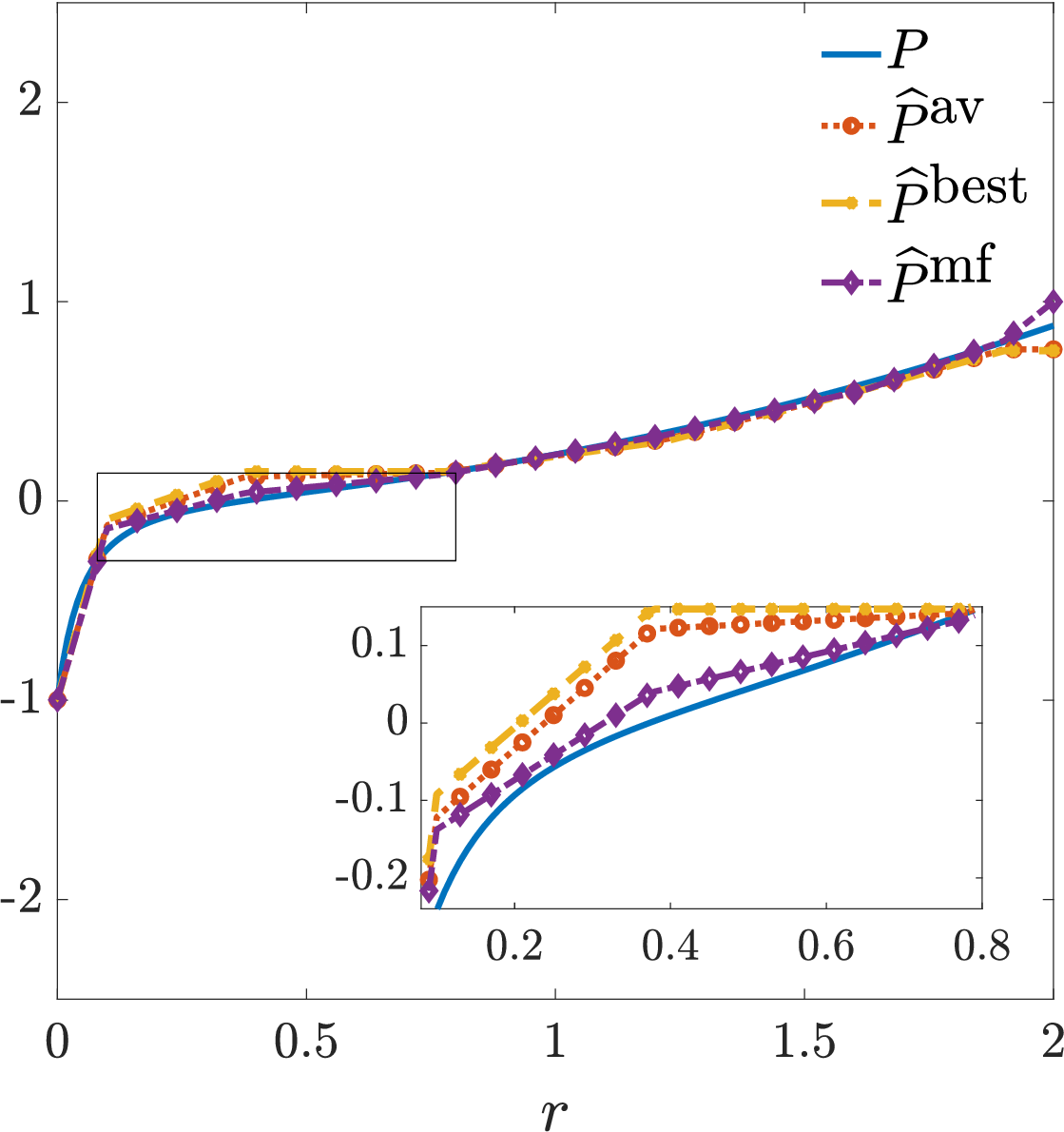}\quad\includegraphics[height=5.45cm]{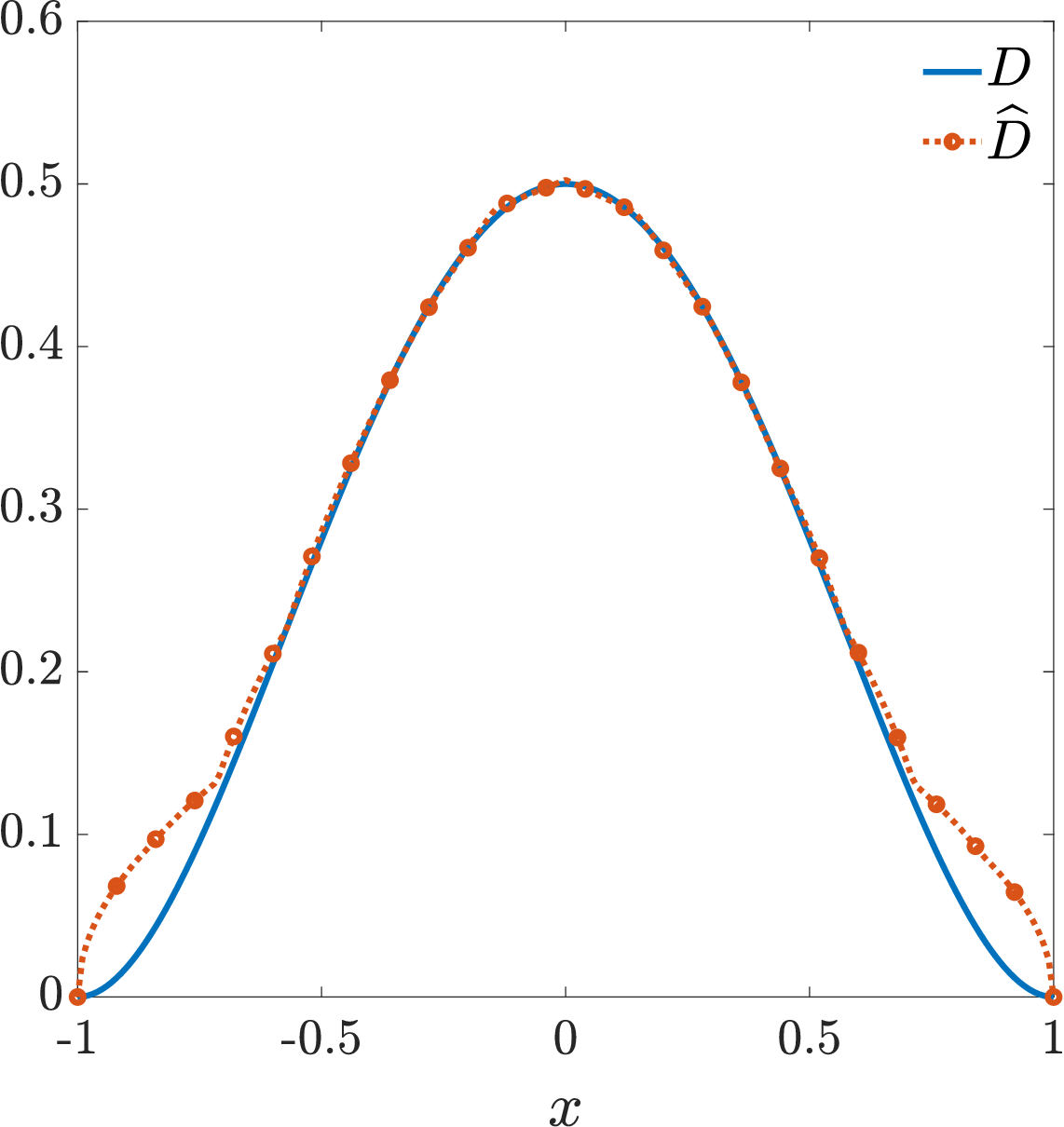}\quad\includegraphics[height=5.4cm]{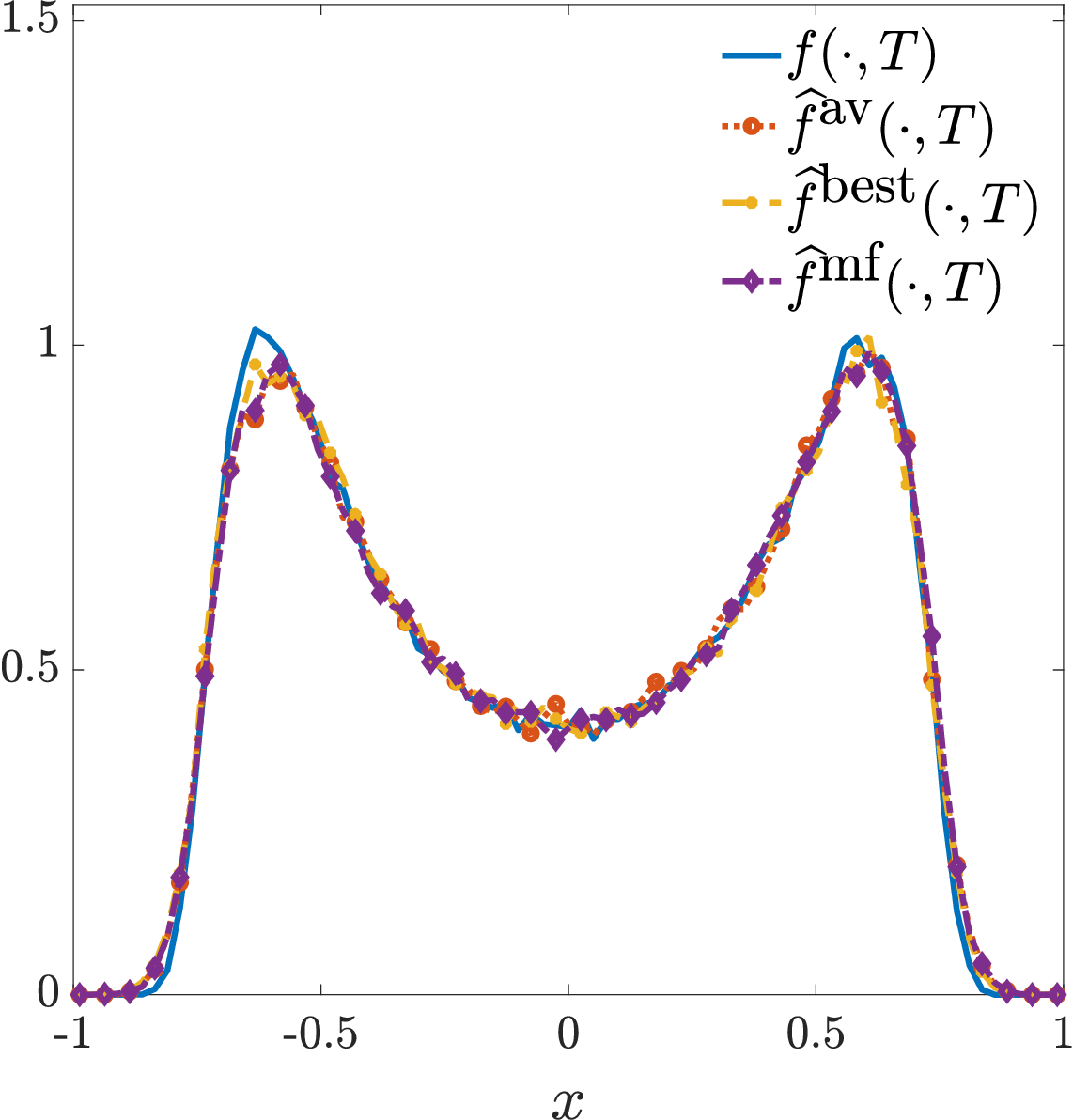}
\caption{Test 2. Reconstruction of kernel $P(r) = ((0.1+r)^2 - 0.05(0.1+r)^{-2})/5$ using $8$ basis functions, $\ell=2$ and $M_P=50$ using approach \eqref{eq:weights1} (averaging) and \eqref{eq:weights2} (best) (left), with $N_p = 1000$ and $K=10$. Reconstruction of the diffusion function $D(x)$ using $15$ basis functions $\ell=1$ and $M = 10$ (center), comparison between the data and the reconstructed density at the final time (right).}
\label{fig:av_ar_new}
\end{figure}

Figure \ref{fig:data_kernel3} shows in orange the data distribution for the case $\ell = 4$ and $M_P = 25$ (left), and in black the ones used to reconstruct the diffusion using $\ell = 1$ and $M_D = 10$ (right). Table \ref{tab:tabella_rb_ar} shows the qualitative results using both weights defined in \eqref{eq:weights1} and in \eqref{eq:weights2}, and Figure \ref{fig:av_ar_new} presents the comparison between the approximated kernels and the exact one (left), the comparison between the reconstructed diffusion function and the exact one (center), and the comparison between the profile of the data at the final time and the reconstructed density using the two approximations of the interaction kernel (right).
We can see that the methods agree on the reconstruction of the kernel. The diffusion has a slight difference at the border which is somehow expected due to the fact that we do not have enough information there coming from the data.

\begin{figure}[htbp]
\centering
\includegraphics[width=0.45\linewidth]{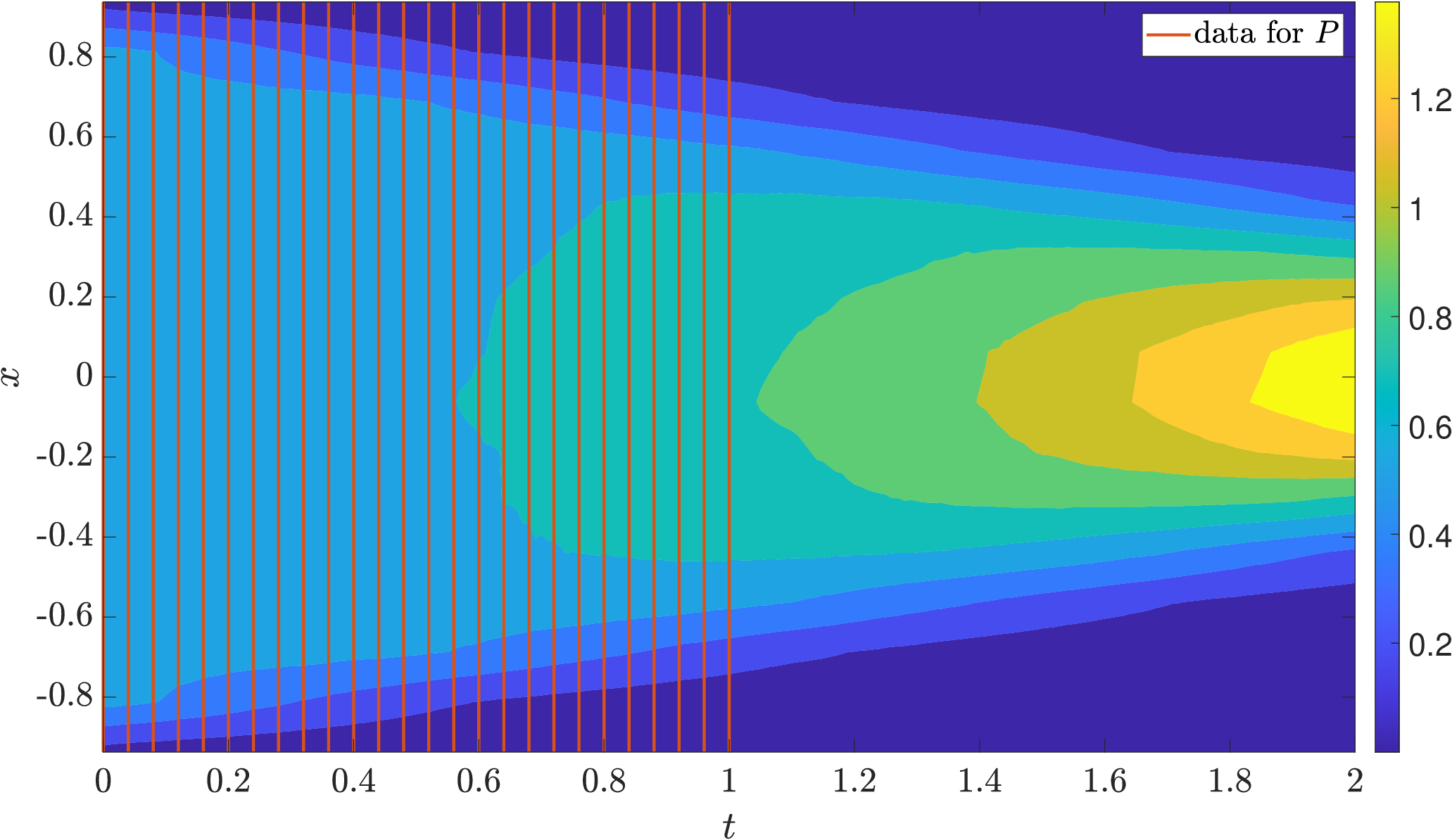}\quad\includegraphics[width=0.45\linewidth]{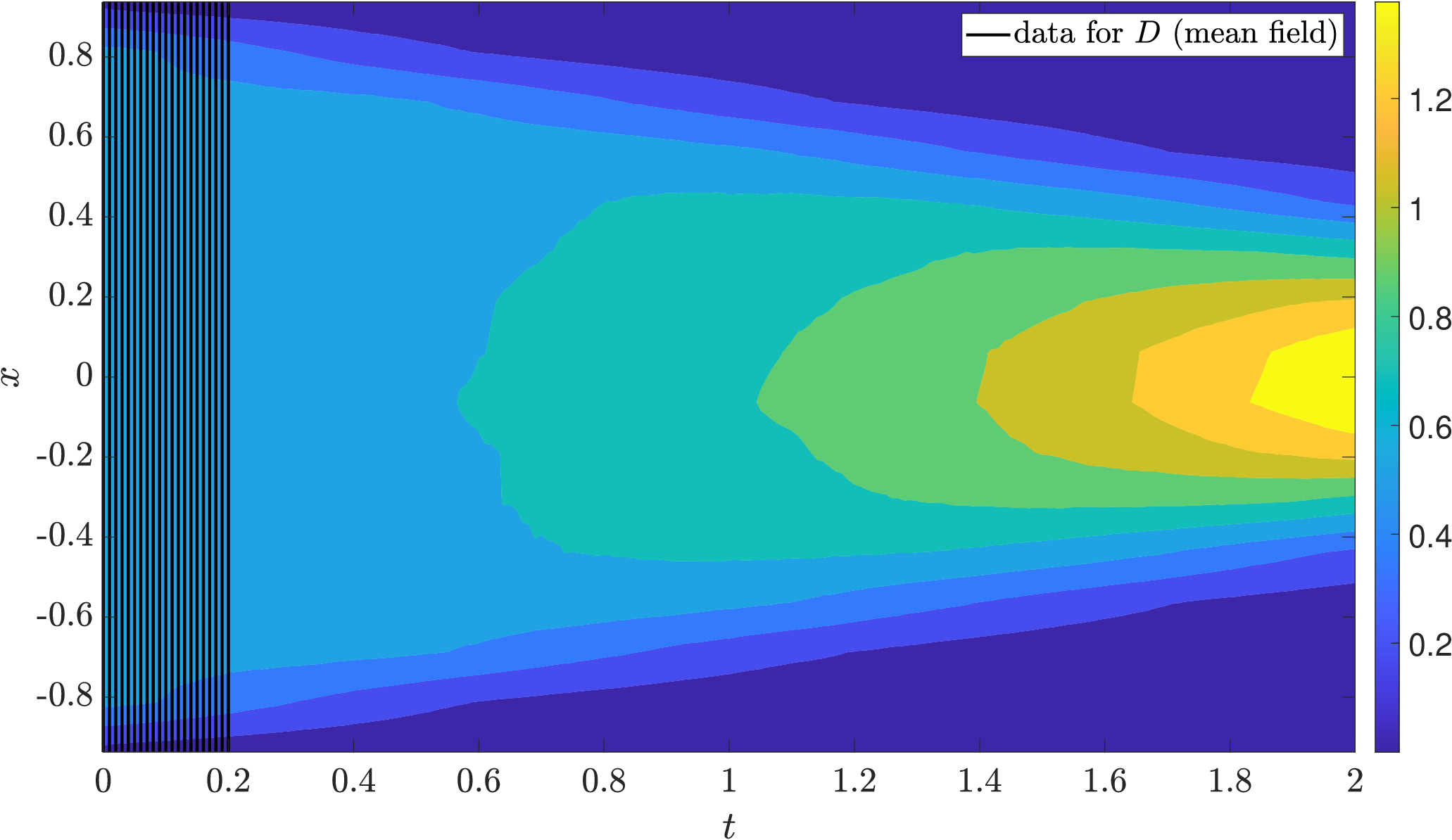}\
\caption{Test 3. Data used for the reconstruction of the kernel when $\ell = 4$ and $M_P=25$ (left) and data used for the reconstruction of the diffusion in the mean field approach with $\ell = 1$ and $M_D=20$ (right).}
\label{fig:data_kernel4}
\end{figure}

\begin{figure}[htbp]
\centering
\includegraphics[height=5.4cm]{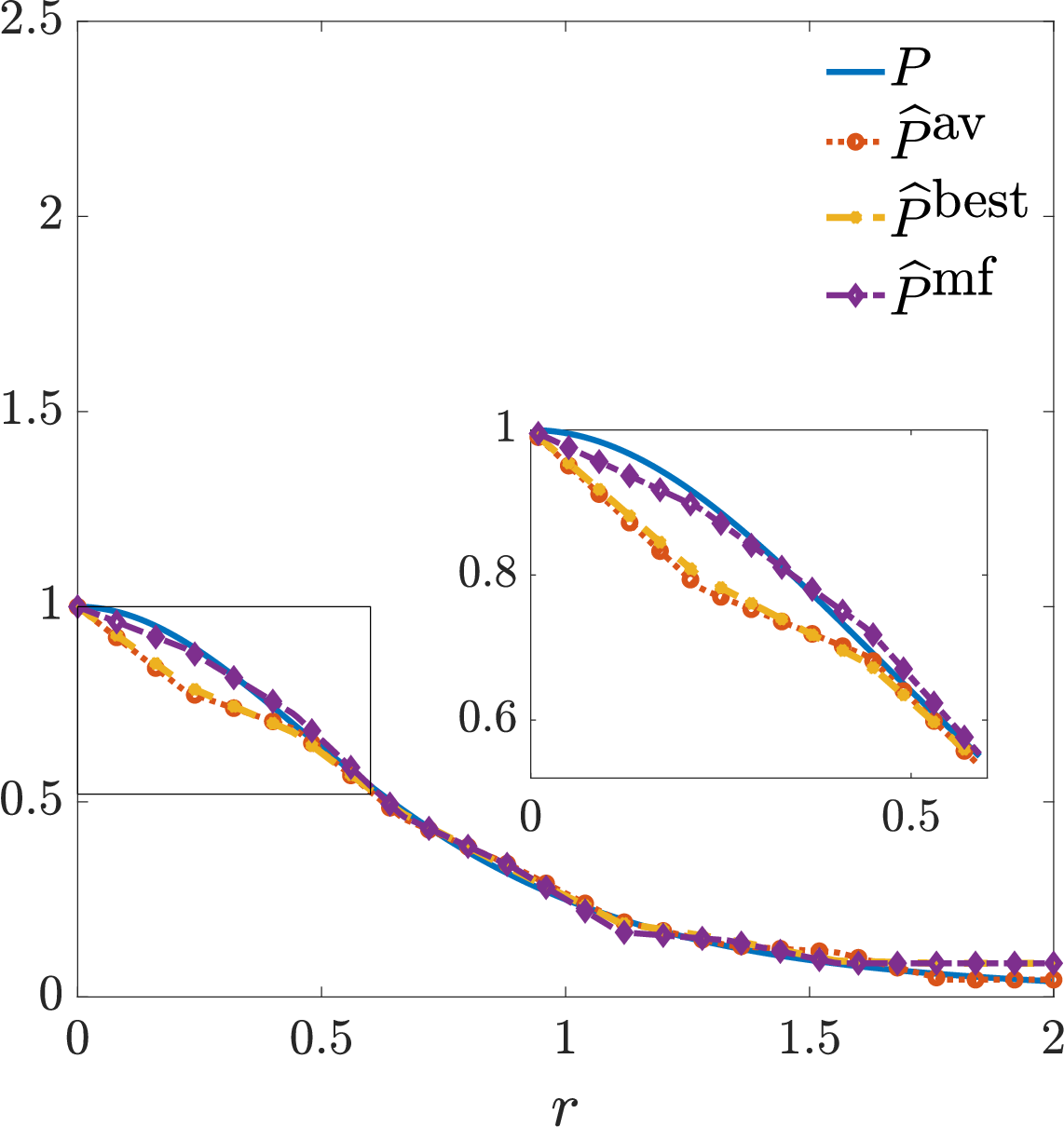}\quad\includegraphics[height=5.4cm]{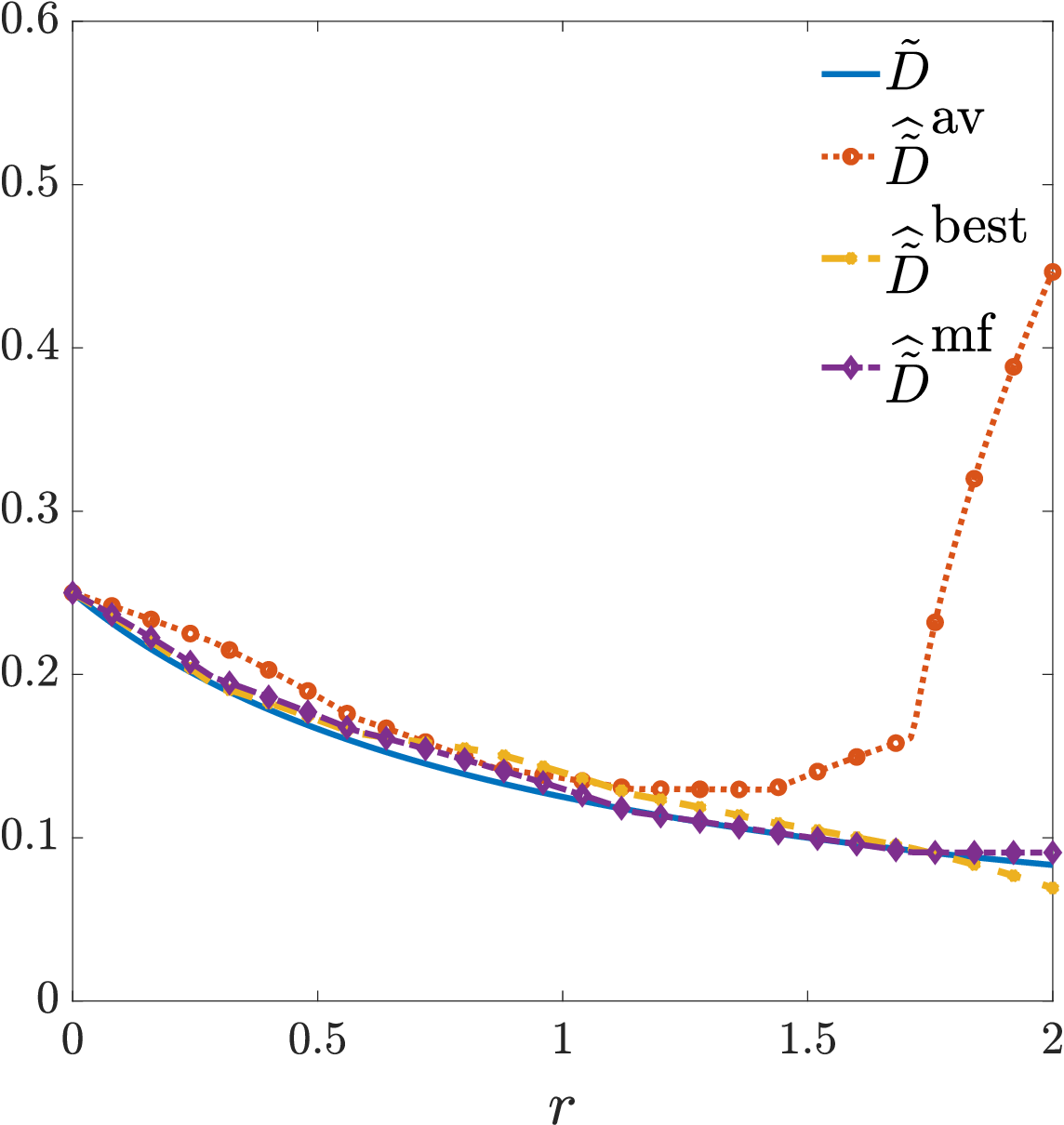}\quad\includegraphics[height=5.4cm]{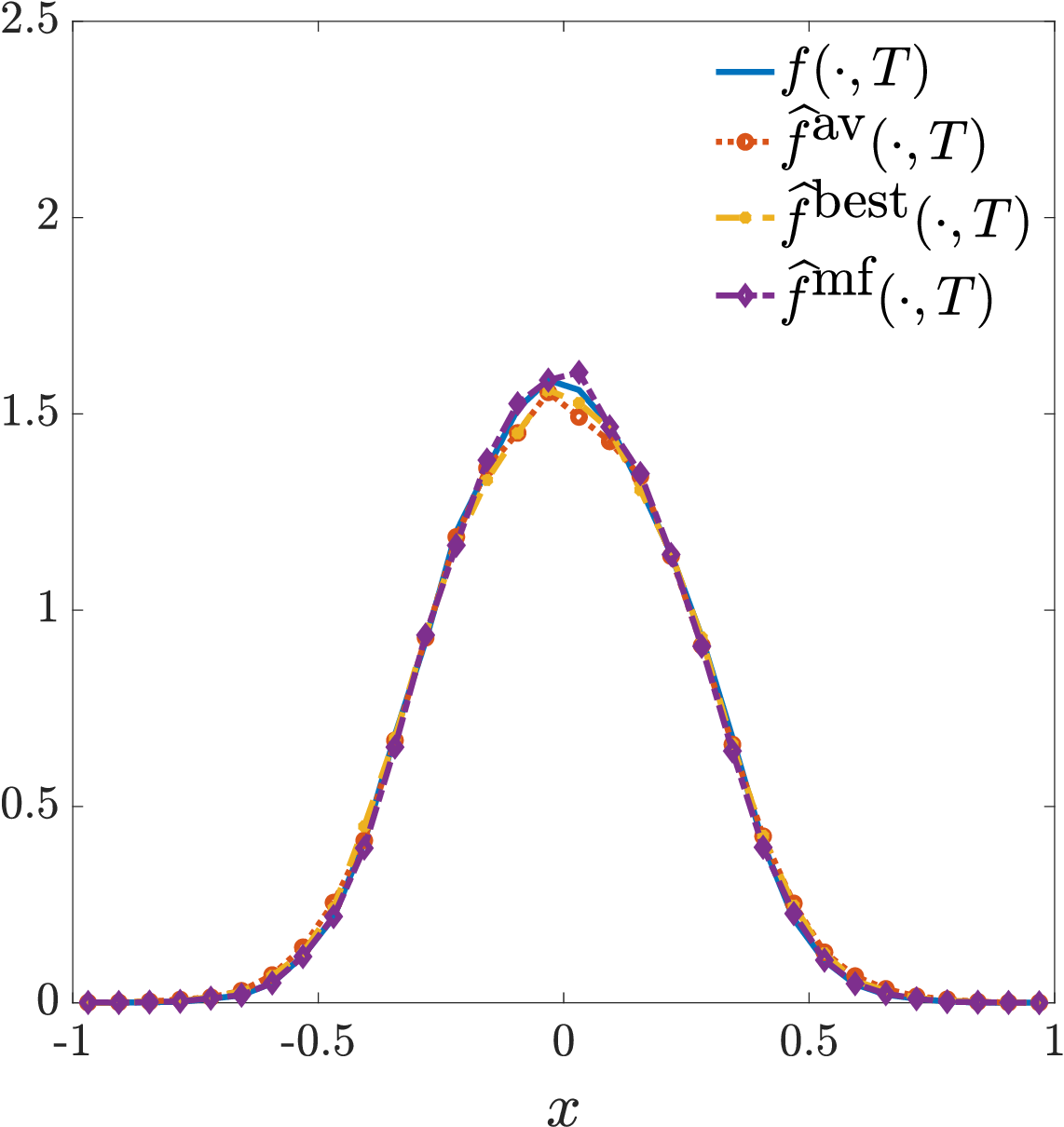}
\caption{Test 3. Reconstruction of the kernel $P(r)=(1+r^2)^{-2}$ using $10$ basis functions, $\mathscr{S}=3$, with approaches \eqref{eq:weights1} (averaging) and \eqref{eq:weights2} (best), $N_p=1000$, and $K=10$, compared with the mean-field method (left). Reconstruction of the diffusion $D(x)$ using $8$ basis functions with $\mathscr{S}=2$ for the random batch approach and $\ell=1$, $M_D=10$ for the mean-field method (center). Comparison between the data and the reconstructed density at the final time (right).} \label{fig:diffnonloc2}
\end{figure}

\begin{figure}[htbp]
\centering
\includegraphics[width=0.3\linewidth]{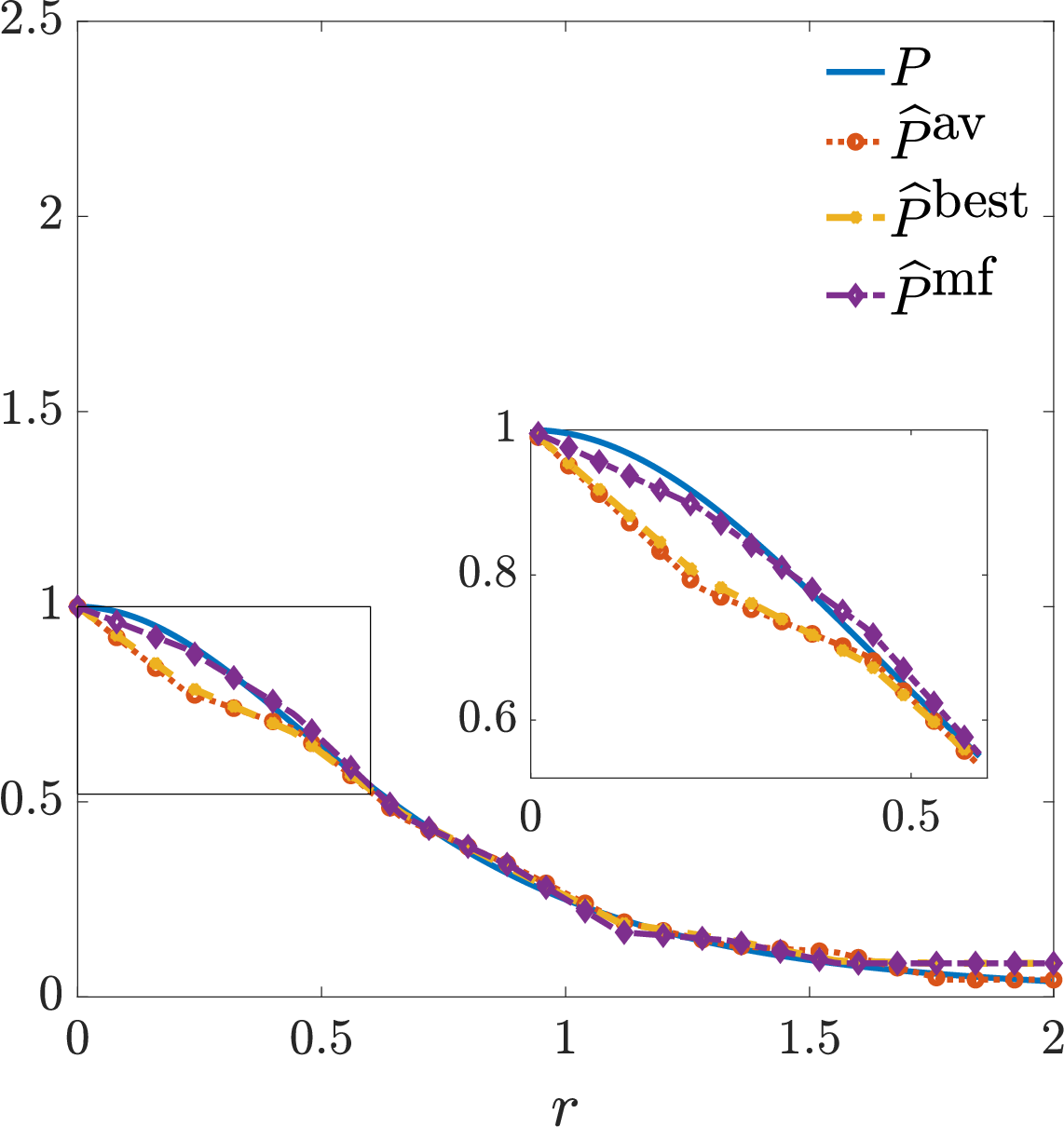}\quad\includegraphics[width=0.305\linewidth]{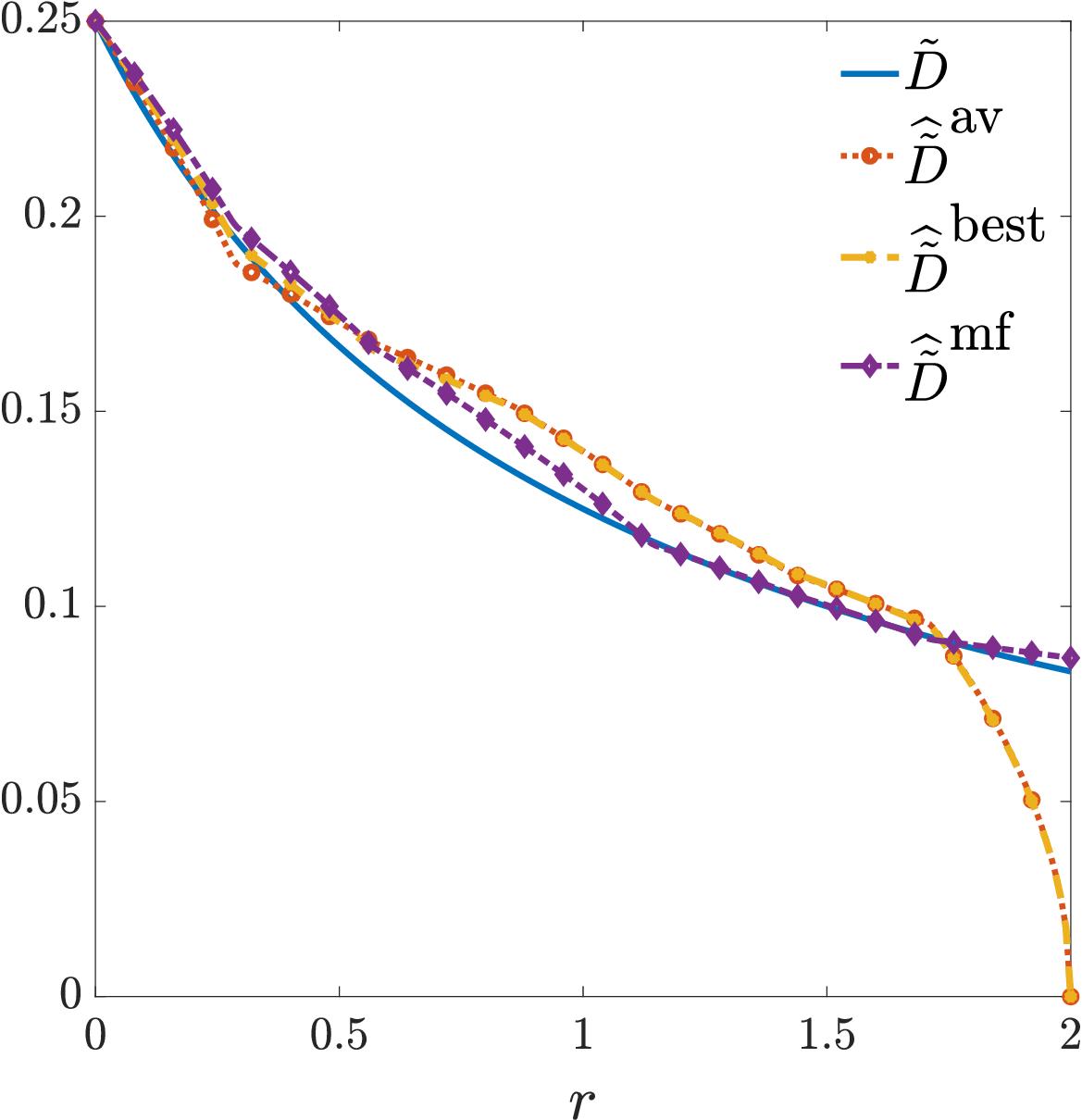}\quad\includegraphics[width=0.3\linewidth]{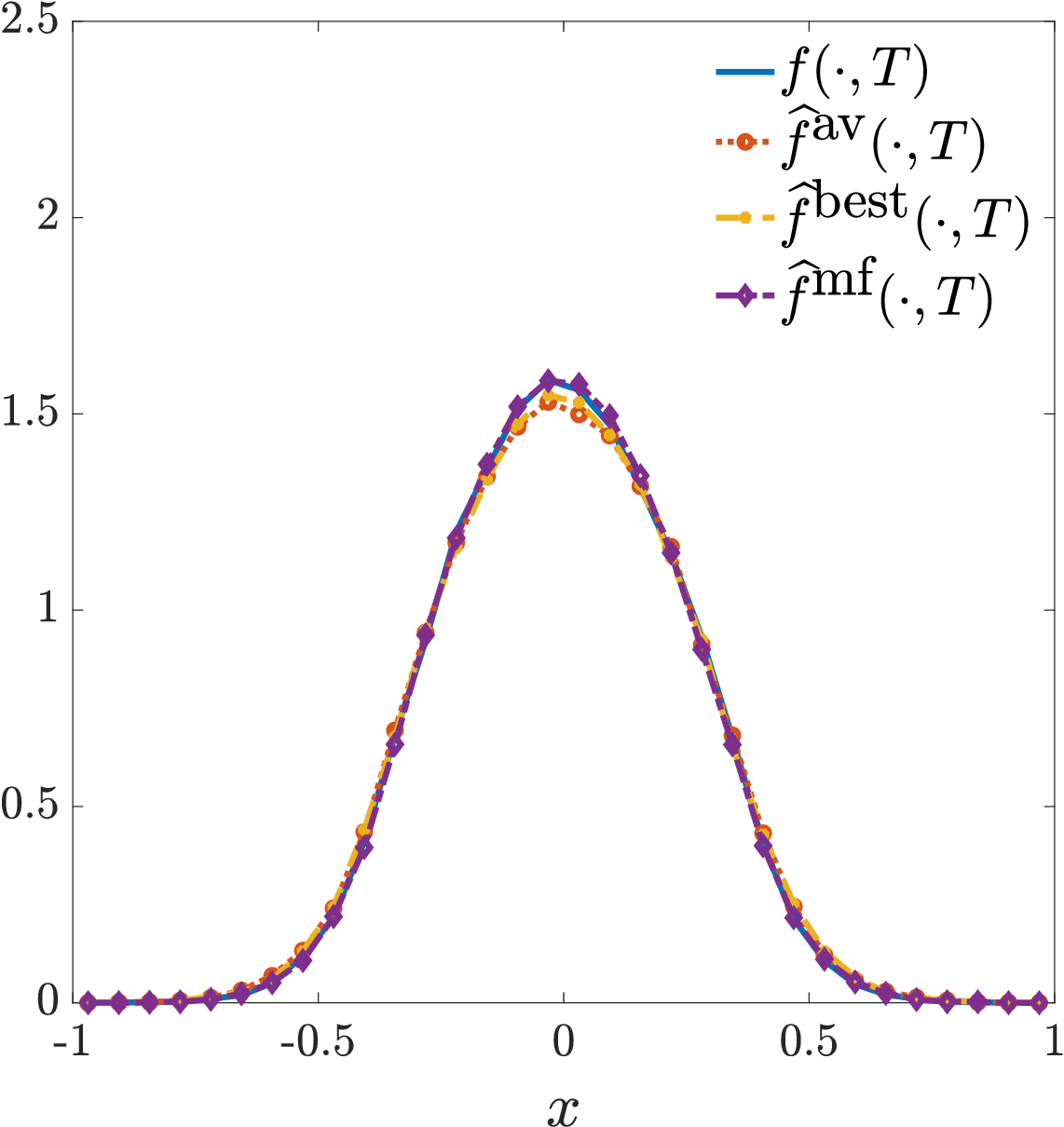}
\caption{Test 3. Reconstruction of kernel $P(r) = ((1+r^2)^{-2}$ using $10$ basis functions, $\ell=4$ and $M_P=25$ using approach \eqref{eq:weights1} (averaging) and \eqref{eq:weights2} (best), with $N_p = 1000$ and $K=10$, superposed with the mean field (left). Reconstruction of the diffusion function $D(x)$ using $8$ basis functions $\ell=2$ and $M_D=50$ for the random batch approach and $\ell=1$ and $M_D = 10$ for the mean field (center). Comparison between the data and the reconstructed density at the final time (right) using the reconstructed kernels and densities.}
\label{fig:diffnonloc2_monotonia}
\end{figure}

\begin{table}[htbp]
\small
\centering
\begin{tabular}{|c|c|c|c|c|}
	\hline
	\multicolumn{5}{|c|}{Averaging method (Alg. \ref{alg:1})}\\
	\hline
	$\mathscr{S}$ & $E^1_P$ & $E^\infty_P$& $E^\textrm{ave}_f$ & $E^\textrm{fin}_f$\\
	\hline
	$1$ & $1.41\cdot 10^{-1}$ & $1.65\cdot 10^{-1}$ & $4,98\cdot 10^{-3}$ & $4.77\cdot 10^{-3}$ \\
	$2$ & $1.30\cdot 10^{-1}$ & $1.45\cdot 10^{-1}$ & $5.30\cdot 10^{-3}$ & $4.83\cdot 10^{-3}$  \\
	$3$ & $2.09\cdot 10^{-1}$ & $2.48\cdot 10^{-1}$ & $5.14\cdot 10^{-3}$ & $5.12\cdot 10^{-3}$ \\
	\hline
	\multicolumn{5}{|c|}{Best result method (Alg. \ref{alg:1})}\\
	\hline
	$\mathscr{S}$ & $E^1_P$ & $E^\infty_P$& $E^\textrm{ave}_f$ & $E^\textrm{fin}_f$\\
	\hline
	$1$ & $1.20\cdot 10^{-1}$ & $1.42\cdot 10^{-1}$ & $5.53\cdot 10^{-3}$ & $8.02\cdot 10^{-3}$ \\
	$2$ & $1.58\cdot 10^{-1}$ & $1.50\cdot 10^{-1}$ & $4.85\cdot 10^{-3}$ & $4.78\cdot 10^{-3}$  \\
	$3$ & $2.09\cdot 10^{-1}$ & $2.56\cdot 10^{-1}$ & $5.44\cdot 10^{-3}$ & $5.67\cdot 10^{-3}$ \\
	\hline
	\multicolumn{5}{|c|}{Mean Field Approach (Alg. \ref{alg:2})}\\
	\hline
	$\mathscr{S}$ & $E^1_P$ & $E^\infty_P$& $E^\textrm{ave}_f$ & $E^\textrm{fin}_f$ \\
	\hline
	$1$ & $6.45\cdot 10^{-2}$ & $1.51\cdot 10^{-1}$ & $6.56\cdot 10^{-3}$ & $9.67\cdot 10^{-3}$ \\
	$2$ & $6.39\cdot 10^{-2}$ & $1.50\cdot 10^{-1}$ & $3.87\cdot 10^{-3}$ & $3.71\cdot 10^{-3}$  \\
	$3$ & $1.16\cdot 10^{-1}$ & $1.44\cdot 10^{-1}$ & $4.28\cdot 10^{-3}$ & $3.61\cdot 10^{-3}$ \\
	\hline
\end{tabular}
\caption{Test 2. Errors for the identification of kernel $P(r)=((0.1+r)^2 - 0.05(0.1+r)^{-2})/5$ with $N_b^P=8$ basis functions. }
\label{tab:tabella_rb_ar}
\end{table}
\paragraph{Test 3: non-local diffusion}
In this last one-dimensional simulation we consider a model  with a Cucker--Smale-type kernel, meaning a radial kernel of the form $P(r) = (1+r^2)^{-2}$. This interaction function models the decay of influence with increasing distance $r$ between agents, capturing the realistic weakening of influence as agents become more distant in the opinion or spatial domain. 
The diffusion function now depends on the relative position of the two agents interacting, specifically $D(x,y) = \tilde D(x,y)(y-x)$ with $\tilde D(x,y) = \tilde D(\lvert x - y \rvert ) = \tilde D (r) = 0.25/(1+r)^{2}$.
The objective of this simulation is to discover $P$ and $\tilde D$. We notice that in this case, both for the random batch approach and for the mean field one, the subproblem for learning the diffusion function gets modified. 
The dataset is composed of $10^5$ particles and $201$ time frames with $\Delta t = 0.01$. Figure \ref{fig:data_kernel4} shows on the left an example of the data used for the case $\ell=4$ and $M_P = 25$, which is true for both the kernel and the diffusion reconstruction in the random batch case, only for the kernel in the mean field approach. On the right, we have the data used for problem \eqref{eq:regression_mfD} in the setting $\ell = 1$, $M_D = 20$. In the optimization problem, we set $\kappa_P =-1$ and $P_0=1$ in \eqref{eq:AP}. Further, in this test, instead of setting the value of the diffusion equal to zero at the boundary of the domain, we tried two different approaches: in the first, we only impose the value of the diffusion in $\lvert x-y \rvert = r = 0$, obtaining the results presented in Figure \ref{fig:diffnonloc2} and in Table \ref{tab:tabella_nonloc}. In \eqref{eq:AD} we set $\kappa_P=0$ and $\overline{\zeta}_0=0.25$.  As it is clear from the central picture in Figure \ref{fig:diffnonloc2}, in this problem the averaging approach for Algorithm \ref{alg:1} isn't able to reconstruct the value of the diffusion for large distances between particles. This is due to the structure of the data implied, since the large majority of the particles are close together and the probability of finding a random couple that has a distance $r \approx 2$ is low. This lack of information on far couples is what causes such strong inaccuracy for said approach for larger value of $r\in \left[ 0, 2 \right]$.
Figure \ref{fig:diffnonloc2_monotonia} and Table \ref{tab:tabella_nonloc_monotonia} show the results when monotonicity is imposed for the diffusion learning problem (in \eqref{eq:AD} we set $\kappa_P=-1$ and $\overline{\zeta}_0=0.25$). Again, as in the kernel reconstruction, it is hard for the random batch method to reconstruct the values of the diffusion when the distance between interacting particles $r$ is close to $2$, for the same reason stated before. It is worth noting that the mean-field approach overcomes this lack of information by reconstructing the density $f$ on the entire domain $[-1,1]$.
\begin{table}[htbp]
\small
\centering
\begin{tabular}{|c|c|c|c|c|c|c|}
\hline
\multicolumn{7}{|c|}{Averaging method (Alg. \ref{alg:1})}\\
\hline
$\mathscr{S}$ & $E^1_P$ & $E^\infty_P$& $E^\textrm{ave}_f$ & $E^\textrm{fin}_f$ & $E^1_D$ & $E^\infty_D$\\
\hline
$1$ & $7.94\cdot 10^{-2}$ & $1.21\cdot 10^{-1}$ & $3.34\cdot 10^{-3}$ & $6.38\cdot 10^{-3}$ & $2.97\cdot 10^{-1}$ & $7.42\cdot 10^{-1}$ \\
$2$ & $7.53\cdot 10^{-2}$ & $1.06\cdot 10^{-1}$ & $3.35\cdot 10^{-3}$ & $7.37\cdot 10^{-3}$ & $3.04\cdot 10^{-1}$ & $7.42\cdot 10^{-1}$ \\
$3$ & $6.78\cdot 10^{-2}$ & $1.01\cdot 10^{-1}$ & $2.82\cdot 10^{-3}$ & $6.58\cdot 10^{-3}$ & $3.88\cdot 10^{-1}$ & $1.45$\\
\hline
\multicolumn{7}{|c|}{Best result method (Alg. \ref{alg:1})}\\
\hline
$\mathscr{S}$ & $E^1_P$ & $E^\infty_P$& $E^\textrm{ave}_f$ & $E^\textrm{fin}_f$ & $E^1_D$ & $E^\infty_D$\\
\hline
$1$ & $7.30\cdot 10^{-2}$ & $1.26\cdot 10^{-1}$ & $2.05\cdot 10^{-3}$ & $5.63\cdot 10^{-3}$& $5.33\cdot 10^{-2}$ & $1.32\cdot 10^{-1}$ \\
$2$ & $6.39\cdot 10^{-2}$ & $6.16\cdot 10^{-2}$ & $1.95\cdot 10^{-3}$ & $4.73\cdot 10^{-3}$ & $5.12\cdot 10^{-2}$ & $9.25\cdot 10^{-2}$ \\
$3$ & $7.08\cdot 10^{-2}$ & $1.31\cdot 10^{-1}$ & $2.58\cdot 10^{-3}$ & $6.04\cdot 10^{-3}$ & $5.46\cdot 10^{-2}$ & $6.99\cdot 10^{-2}$\\
\hline
\multicolumn{7}{|c|}{Mean Field Approach (Alg. \ref{alg:2})}\\
\hline
$\mathscr{S}$ & $E^1_P$ & $E^\infty_P$& $E^\textrm{ave}_f$ & $E^\textrm{fin}_f$ & $E^1_D$ & $E^\infty_D$\\
\hline
$1$ & $4.76\cdot 10^{-2}$ & $4.71\cdot 10^{-2}$ & $1.32\cdot 10^{-3}$ & $1.79\cdot 10^{-3}$ & $3.11\cdot 10^{-2}$ & $3.66\cdot 10^{-2}$\\
$2$ & $4.93\cdot 10^{-2}$ & $4.85\cdot 10^{-2}$ & $1.15\cdot 10^{-3}$ & $1.42\cdot 10^{-3}$ & $3.11\cdot 10^{-2}$ & $3.66\cdot 10^{-2}$ \\
$3$ & $4.42\cdot 10^{-2}$ & $4.65\cdot 10^{-2}$ & $1.53\cdot 10^{-3}$ & $2.15\cdot 10^{-3}$ & $3.11\cdot 10^{-2}$ & $3.66\cdot 10^{-2}$\\
\hline
\end{tabular}
\caption{Test 3. Errors for the identification of kernel $P(r)=(1+r^2)^{-2}$ using $10$ basis functions and diffusion $\tilde D(r) = 0.25 (1+r)^{-2}$ using $8$ basis functions. }
\label{tab:tabella_nonloc}
\end{table}

\begin{table}[htbp]
\small
\centering
\begin{tabular}{|c|c|c|c|c|c|c|}
\hline
\multicolumn{7}{|c|}{Averaging method (Alg. \ref{alg:1})}\\
\hline
$\mathscr{S}$ & $E^1_P$ & $E^\infty_P$& $E^\textrm{ave}_f$ & $E^\textrm{fin}_f$ & $E^1_D$ & $E^\infty_D$\\
\hline
$1$ & $7.20\cdot 10^{-2}$ & $1.21\cdot 10^{-1}$ & $1.78\cdot 10^{-3}$ & $3.58\cdot 10^{-3}$ & $7.68\cdot 10^{-2}$ & $3.33\cdot 10^{-1}$ \\
$2$ & $8.77\cdot 10^{-2}$ & $1.49\cdot 10^{-1}$ & $2.18\cdot 10^{-3}$ & $5.86\cdot 10^{-3}$ & $7.66\cdot 10^{-2}$ & $3.33\cdot 10^{-1}$ \\
$3$ & $6.70\cdot 10^{-2}$ & $1.26\cdot 10^{-1}$ & $2.24\cdot 10^{-3}$ & $5.29\cdot 10^{-3}$ & $7.59\cdot 10^{-2}$ & $3.33\cdot 10^{-1}$\\
\hline
\multicolumn{7}{|c|}{Best result method (Alg. \ref{alg:1})}\\
\hline
$\mathscr{S}$ & $E^1_P$ & $E^\infty_P$& $E^\textrm{ave}_f$ & $E^\textrm{fin}_f$ & $E^1_D$ & $E^\infty_D$\\
\hline
$1$ & $6.36\cdot 10^{-2}$ & $8.54\cdot 10^{-2}$ & $1.68\cdot 10^{-3}$ & $5.01\cdot 10^{-3}$& $7.68\cdot 10^{-2}$ & $3.33\cdot 10^{-1}$ \\
$2$ & $6.21\cdot 10^{-2}$ & $8.83\cdot 10^{-2}$ & $1.83\cdot 10^{-3}$ & $4.69\cdot 10^{-3}$ & $7.53\cdot 10^{-2}$ & $3.33\cdot 10^{-1}$ \\
$3$ & $7.34\cdot 10^{-2}$ & $1.10\cdot 10^{-1}$ & $2.06\cdot 10^{-3}$ & $5.08\cdot 10^{-3}$ & $7.57\cdot 10^{-2}$ & $3.33\cdot 10^{-1}$\\
\hline
\multicolumn{7}{|c|}{Mean Field Approach (Alg. \ref{alg:2})}\\
\hline
$\mathscr{S}$ & $E^1_P$ & $E^\infty_P$& $E^\textrm{ave}_f$ & $E^\textrm{fin}_f$ & $E^1_D$ & $E^\infty_D$\\
\hline
$1$ & $4.76\cdot 10^{-2}$ & $4.71\cdot 10^{-2}$ & $1.59\cdot 10^{-3}$ & $1.53\cdot 10^{-3}$ & $2.85\cdot 10^{-2}$ & $3.71\cdot 10^{-2}$\\
$2$ & $4.93\cdot 10^{-2}$ & $4.85\cdot 10^{-2}$ & $1.07\cdot 10^{-3}$ & $1.57\cdot 10^{-3}$ & $2.85\cdot 10^{-2}$ & $3.71\cdot 10^{-2}$ \\
$3$ & $4.42\cdot 10^{-2}$ & $4.65\cdot 10^{-2}$ & $1.33\cdot 10^{-3}$ & $1.84\cdot 10^{-3}$ & $2.85\cdot 10^{-2}$ & $3.71\cdot 10^{-2}$\\
\hline
\end{tabular}
\caption{Test 3. Errors for the identification of kernel $P(r)=(1+r^2)^{-2}$ using $10$ basis functions and diffusion $\tilde D(r) = 0.25(1+r)^{-2}$ using $8$ basis functions. }
\label{tab:tabella_nonloc_monotonia}
\end{table}

\newpage
\subsection{Two-dimensional examples}\label{sec:2dim}
In this section, we provide two different two-dimensional examples. In the first one, the interaction kernel is of the attraction-repulsion type, and the diffusion is anisotropic. In the second simulation, instead, the kernel is attractive and the diffusion is nonlocal and radial.
In the two dimensional case we report, instead of the errors defined in \eqref{eq:w1}, we reconstruct the density through histograms, recovering both the density of the data $f^n(x)$ and the reconstructed one using the learned kernels, $\widehat{f}^n(x)$, then we compute
\begin{equation}\label{eq:errors_f}
E^1_f = \frac{\sum_{n=1}^{N_T}\|f^{n} -\widehat{f}^{n}\|_{L^1([-1,1]^2)}}{\sum_{n=1}^{N_T}\|f^{n}\|_{L^1([-1,1]^2)}},\quad
E^1_{f_T} = \frac{\| f^{N_T} -\widehat{f}^{N_T}\|_{L^1([-1,1]^2)}}{\|{f}^{N_T}\|_{L^1([-1,1]^2)}}.
\end{equation}
This numerical error analysis will hold true for both presented algorithms.
\paragraph{Test 4: attraction-repulsion model with anisotropic diffusion} In this first two-dimensional simulation, we consider a model with the attraction-repulsion radial kernel given by $P(r) = (-(0.1+r)^{-1.15} + r^2)/2$, with $r$ being the euclidean distance between the two bi-dimensional opinions. The density is anisotropic and it is equal to $D_1(x_1) = (1-x_1^2)/4$ along the first opinion $x_1$, while it is equal to and $D_2(x_2) = \sqrt{1-x_2^2}/5$ along the second one $x_2$. The dataset is composed of $N=10^5$ agents, with $201$ available time frames for each agent, having a distance of $\Delta t = 0.01$ between them. The initial condition is a uniform distribution in $\left[-0.85,0.85\right]^2$. 
In the optimization problem, we set $\kappa_P =1$ and $P_0=-7$ in \eqref{eq:AP}, whereas in \eqref{eq:AD} we set $\kappa_D=0, \,\overline{\zeta}_0=0 = \overline{\zeta}_{N_b^D.}$ for both diffusion functions.

\begin{figure}[htbp]
\centering
\includegraphics[width=0.3\linewidth]{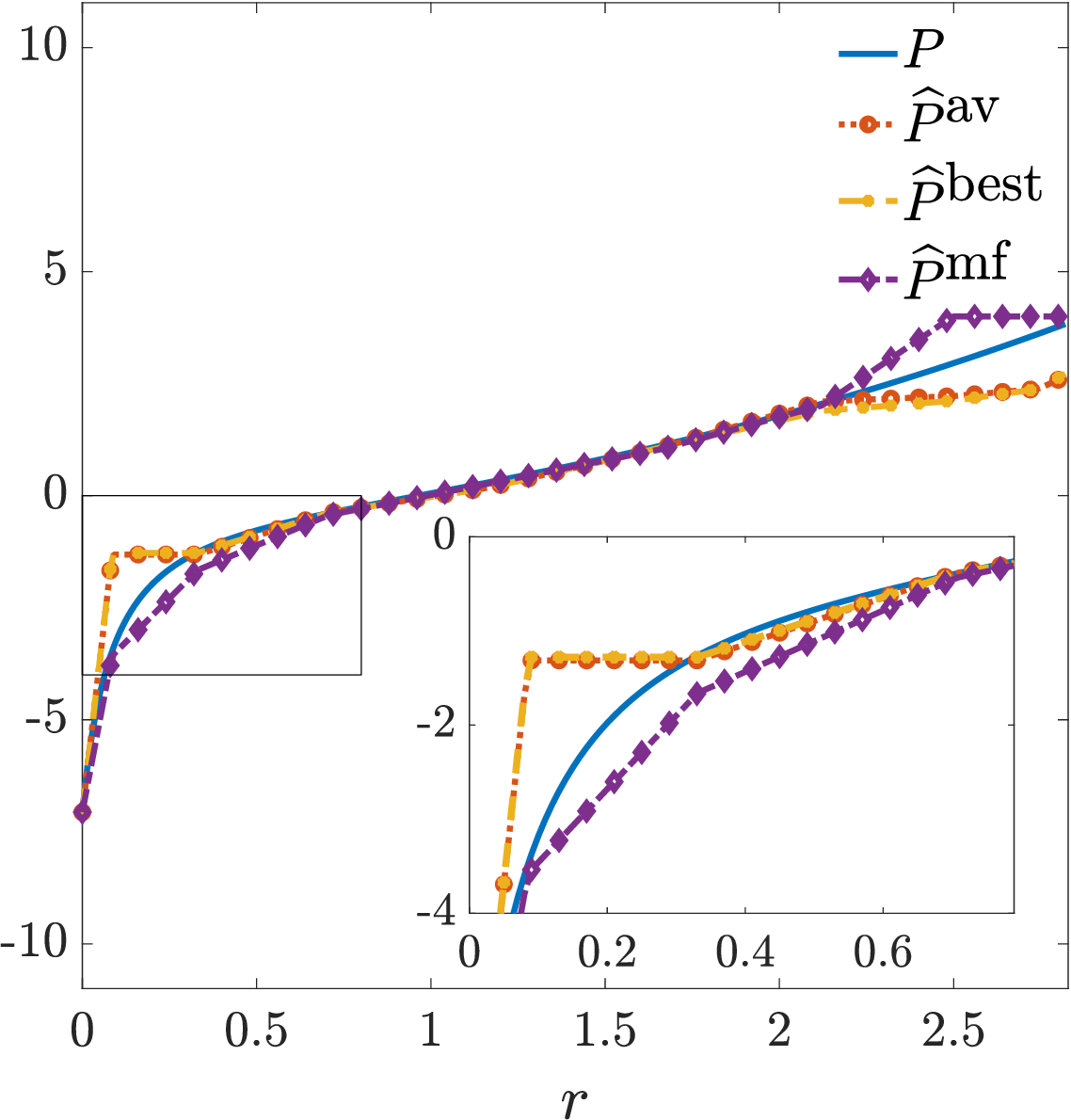}\quad\includegraphics[width=0.305\linewidth]{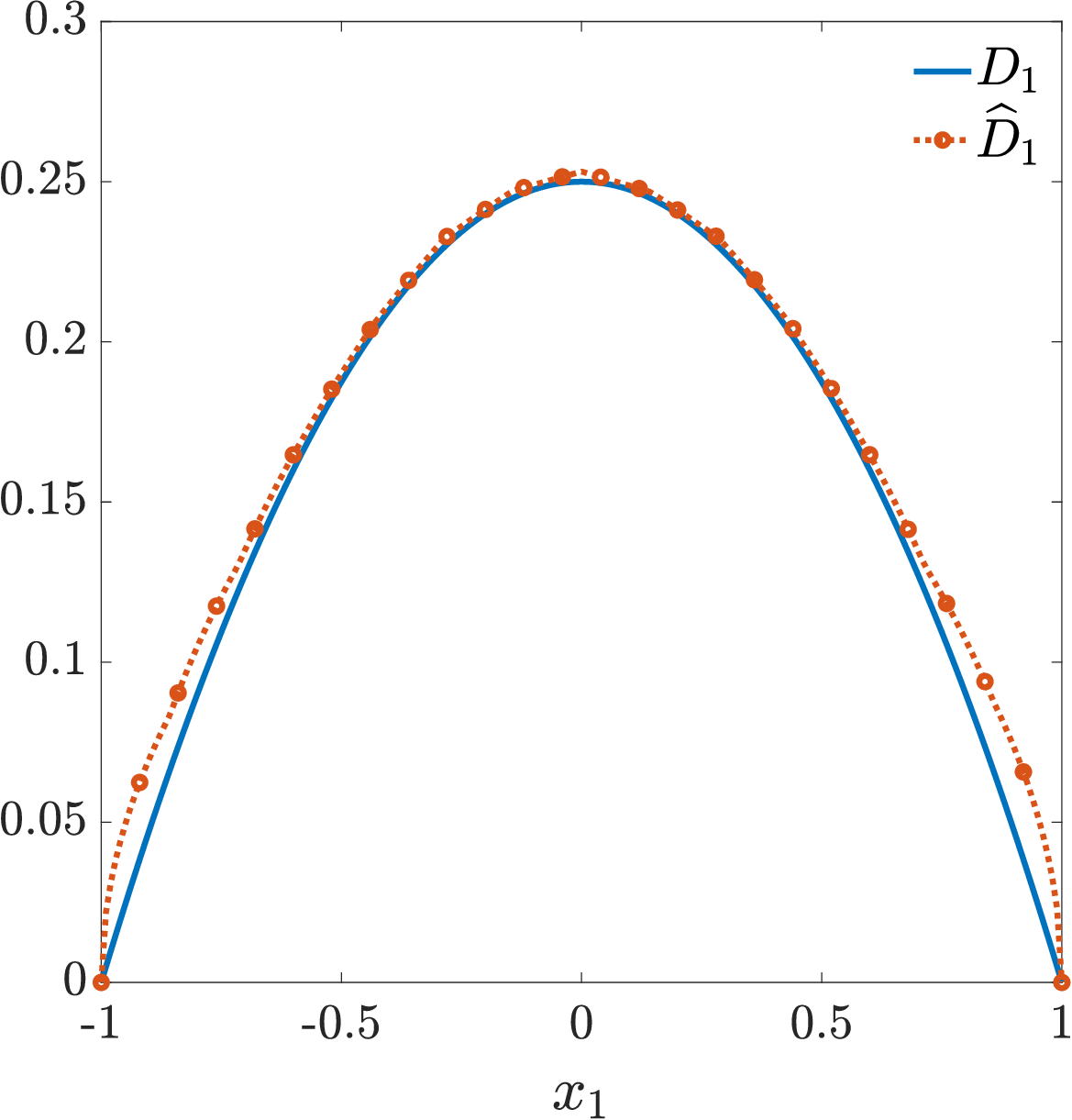}\quad\includegraphics[width=0.305\linewidth]{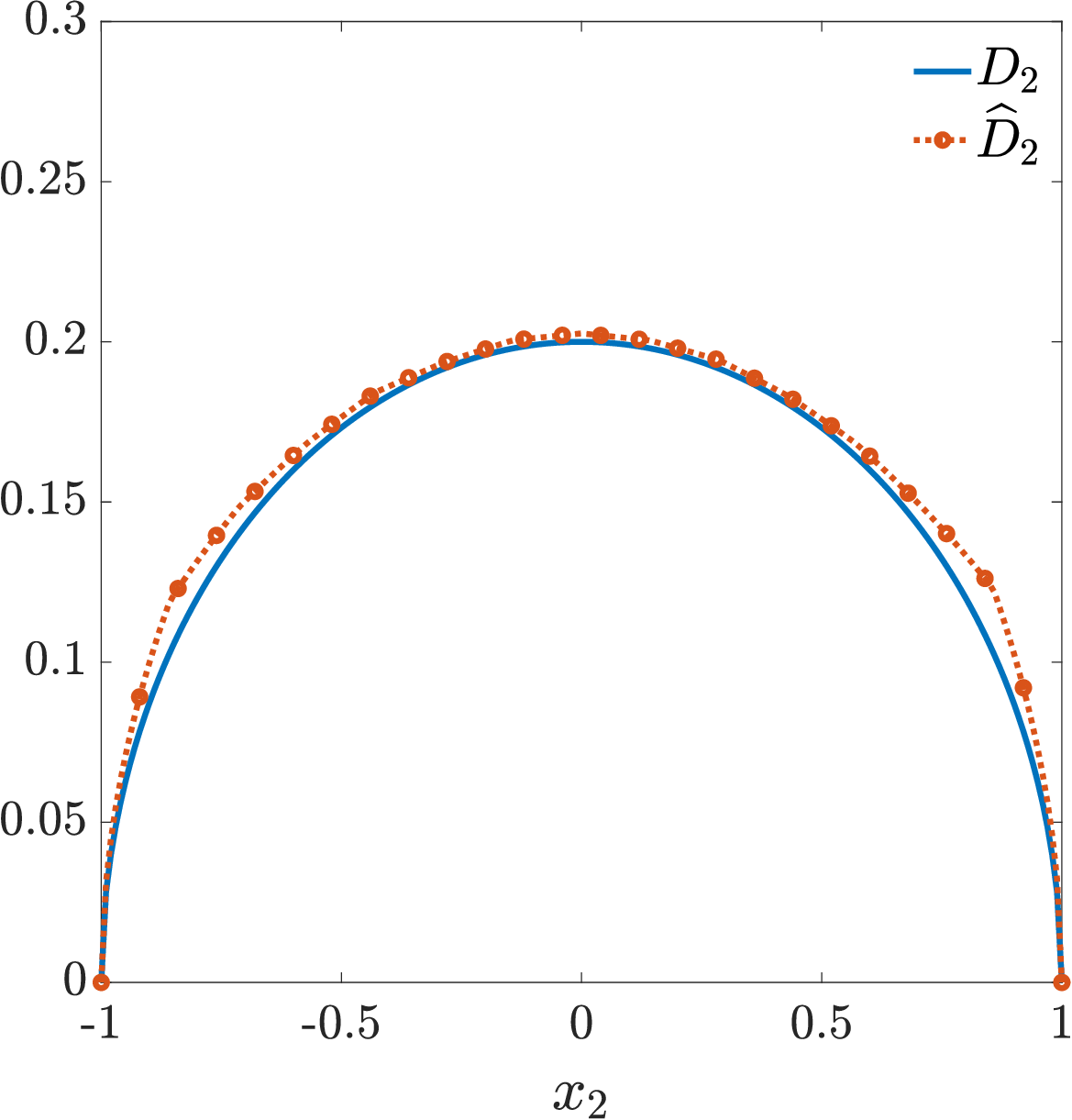}
\caption{Test 4. (Test 2D) Reconstruction of kernel $P = (-(0.1+r)^{-1.15} + r^2)/2$ using $10$ basis functions, $\ell=2$ and $M_P=50$ (left), reconstruction of the diffusion function $D_1(x_1)$ using $15$ basis functions $\ell=1$ and $M_{D_1} = 20$ (center), reconstruction of the diffusion function $D_2(x_2)$ using $15$ basis functions $\ell=1$ and $M_{D_2} = 20$ (right).}
\label{fig:2dcase_mfa1}
\end{figure}

\begin{figure}[htbp]
\centering
\includegraphics[width=0.42\linewidth]{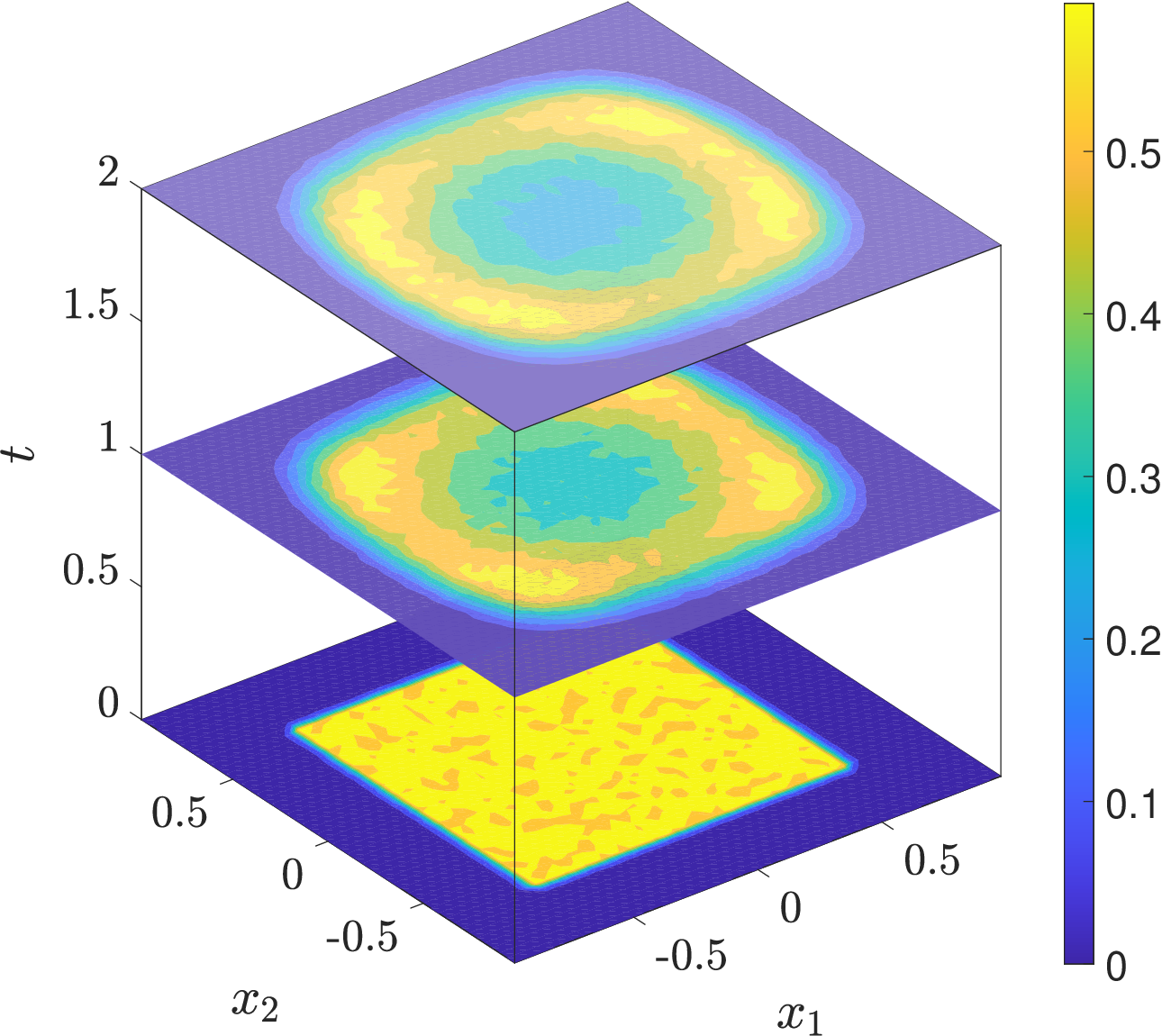}\quad\quad\includegraphics[width=0.35\linewidth]{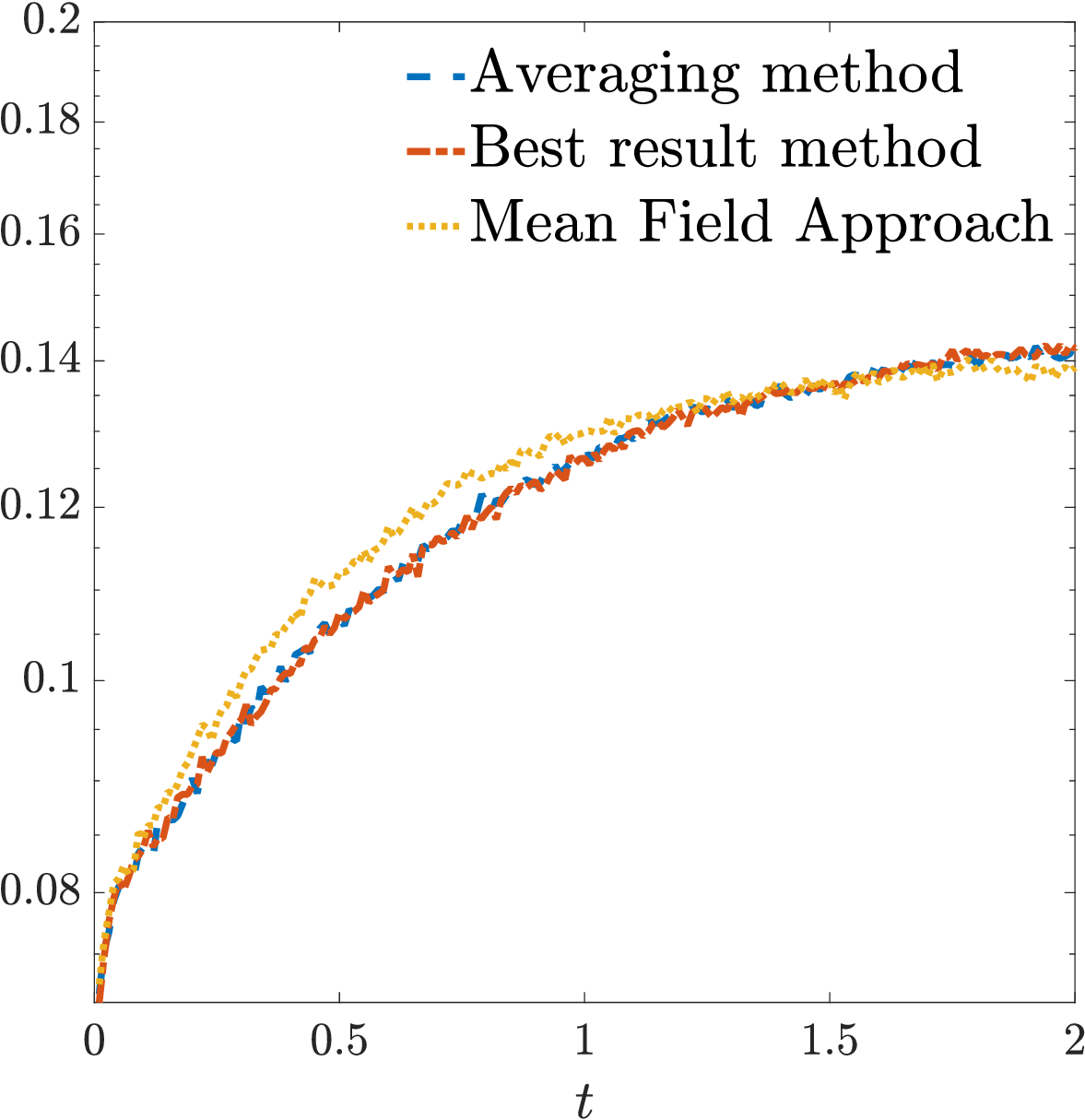}
\caption{Test 4. Evolution of the reconstructed density (left) and trend of the normalized $L^1$ errors over time for the different reconstruction strategies (right), setting $\mathscr{S}=3$. 
}
\label{fig:2dcase_density}
\end{figure}

Figure \ref{fig:2dcase_mfa1} shows a visual comparison between the true kernel and the reconstructed ones using the different approaches (left), the reconstructed diffusion along the first dimension $D_1$ (center), and the reconstructed diffusion along the second dimension $D_2$ (right). Figure \ref{fig:2dcase_density}, instead, shows a comparison between the reconstructed density at the final time and plots of the errors at the final time: the top left picture shows the density at the final time using the averaging approach, the top center one using the best case approach, and the top right one shows the reconstructed density using the mean field approach. The bottom row shows the absolute value of the error on the reconstructed density (compared to the data) for the averaging (left), best case (center) and mean field approaches (right). Finally, Table \ref{tab:tabella_2d} shows a comparison between the errors for different choices of setting $\mathscr{S}$ in the kernel reconstruction, for the three approaches. Also in this example, we can see that the discovered diffusion functions present a larger error close to the boundary but, as already mentioned in the previous tests, this is a consequence of the lack of information when $x_1=\pm 1$ or $x_2=\pm 1$. The kernel for larger $r$ also presents an error but, again, this is expected since we do not have information when the particles are not close.

\begin{table}[htbp]
\small
\centering
\begin{tabular}{|c|c|c|c|c|}
\hline
\multicolumn{5}{|c|}{Averaging method (Alg. \ref{alg:1})}\\
\hline
$\mathscr{S}$ & $E^1_P$ & $E^\infty_P$& $E^1_f$ & $E^1_{f_T}$ \\
\hline
$2$ & $1.83\cdot 10^{-1}$ & $2.92\cdot 10^{-1}$ & $1.21\cdot 10^{-1}$ & $1.42\cdot 10^{-1}$  \\
$3$ & $2.30\cdot 10^{-1}$ & $2.52\cdot 10^{-1}$ & $1.19\cdot 10^{-1}$ & $1.41\cdot 10^{-1}$ \\
\hline
\multicolumn{5}{|c|}{Best result method (Alg. \ref{alg:1})}\\
\hline
$\mathscr{S}$ & $E^1_P$ & $E^\infty_P$ & $E^1_f$ & $E^1_{f_T}$ \\
\hline
$1$ & $4.06\cdot 10^{-1}$ & $3.19\cdot 10^{-1}$ & $8.83\cdot 10^{-2}$ & $9.22\cdot 10^{-2}$ \\
$2$ & $2.06\cdot 10^{-1}$ & $2.97\cdot 10^{-1}$ & $1.20\cdot 10^{-1}$ & $1.43\cdot 10^{-1}$  \\
$3$ & $2.13\cdot 10^{-1}$ & $3.11\cdot 10^{-1}$ & $1.19\cdot 10^{-1}$ & $1.42\cdot 10^{-1}$ \\
\hline
\multicolumn{5}{|c|}{Mean Field Approach (Alg. \ref{alg:2})}\\
\hline
$\mathscr{S}$ & $E^1_P$ & $E^\infty_P$ & $E^1_f$ & $E^1_{f_T}$\\
\hline
$1$ & $1.67\cdot 10^{-1}$ & $1.47\cdot 10^{-1}$ & $1.24\cdot 10^{-1}$ & $1.41\cdot 10^{-1}$ \\
$2$ & $1.66\cdot 10^{-1}$ & $1.47\cdot 10^{-1}$ & $1.23\cdot 10^{-1}$ & $1.40\cdot 10^{-1}$  \\
$3$ & $1.62\cdot 10^{-1}$ & $1.47\cdot 10^{-1}$ & $1.21\cdot 10^{-1}$ & $1.37\cdot 10^{-1}$ \\
\hline
\end{tabular}
\caption{Test 4. Errors for the identification of kernel $P = (-(0.1+r)^{-1.15} + r^2)/2$ using $10$ basis functions. The errors on the diffusion along the first variable are $E^1_{D_1} = 4.40 \cdot 10^{-2}$ and $E^\infty_{D_1} = 1.11\cdot 10^{-2}$, while the errors on the diffusion along the second variable are $E^1_{D_2} = 3.47 \cdot 10^{-2}$ and $E^\infty_{D_2} = 9.82\cdot 10^{-2}$.}
\label{tab:tabella_2d}
\end{table}

\begin{figure}[htbp]
\centering    \includegraphics[width=0.3\linewidth]{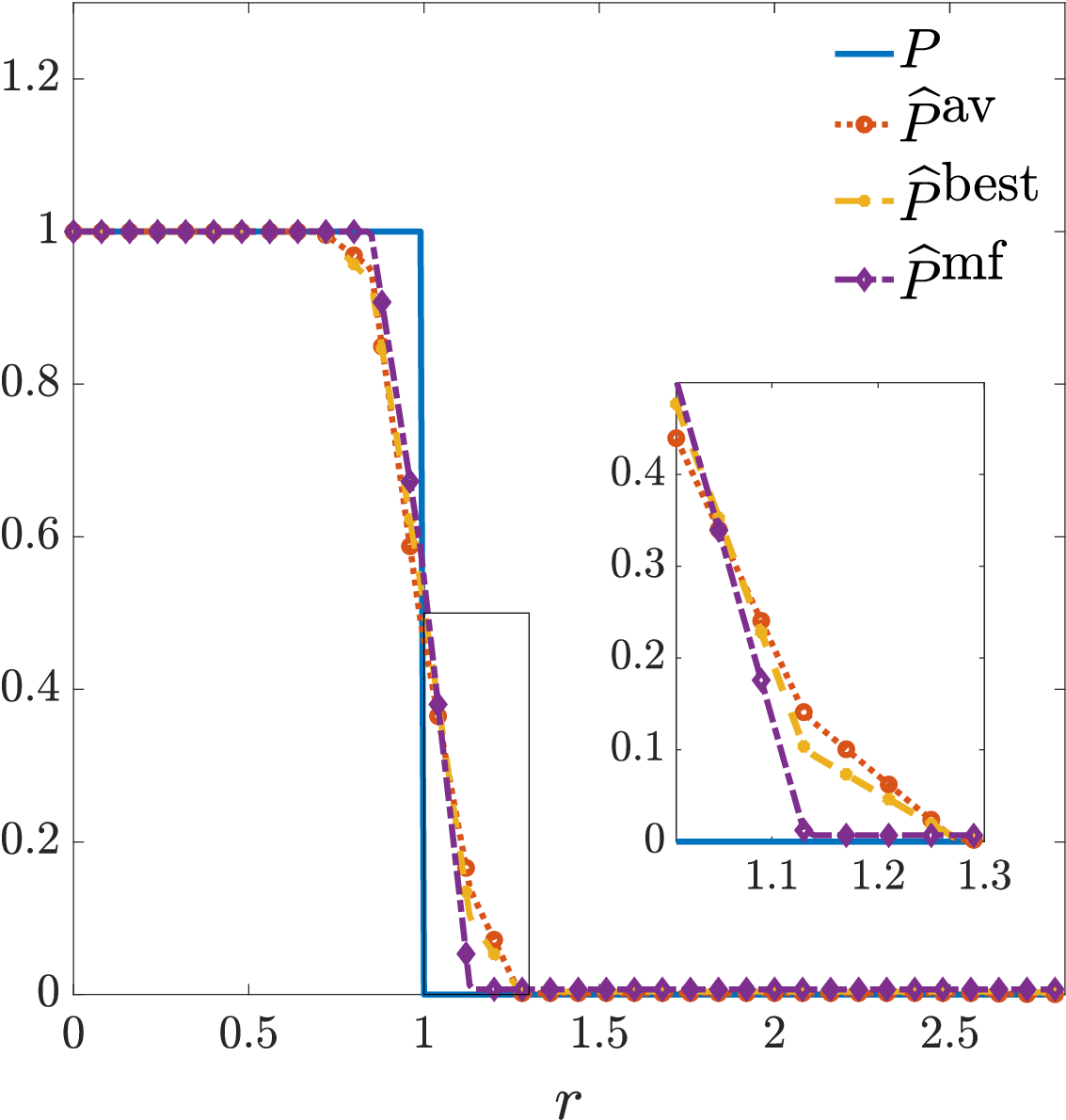}\quad\quad\includegraphics[width=0.3\linewidth]{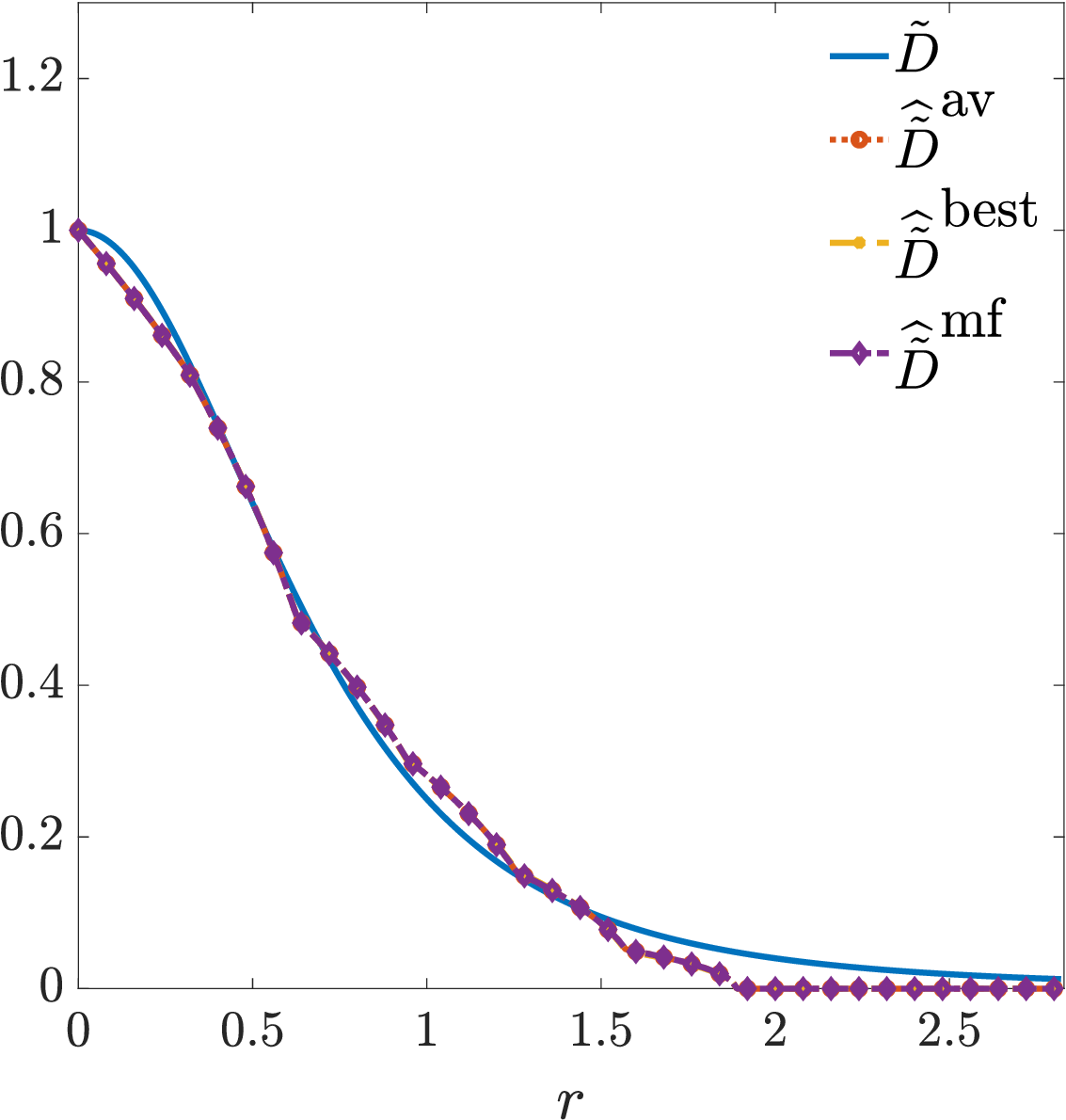}
\caption{Test 5. (Test 2D) Reconstruction of kernel $P = \chi(r<1)$ using $N_b^P=21$, $\ell=4$ and $M_P=25$ (left), reconstruction of the diffusion function $\tilde{D}(r)$ using $N_b^D=10$, $\ell=4$ and $M_{D} = 25$ (right).}
\label{fig:2dcase_test6}
\end{figure}

\begin{figure}[htbp]
\centering
\includegraphics[width=0.4\linewidth]{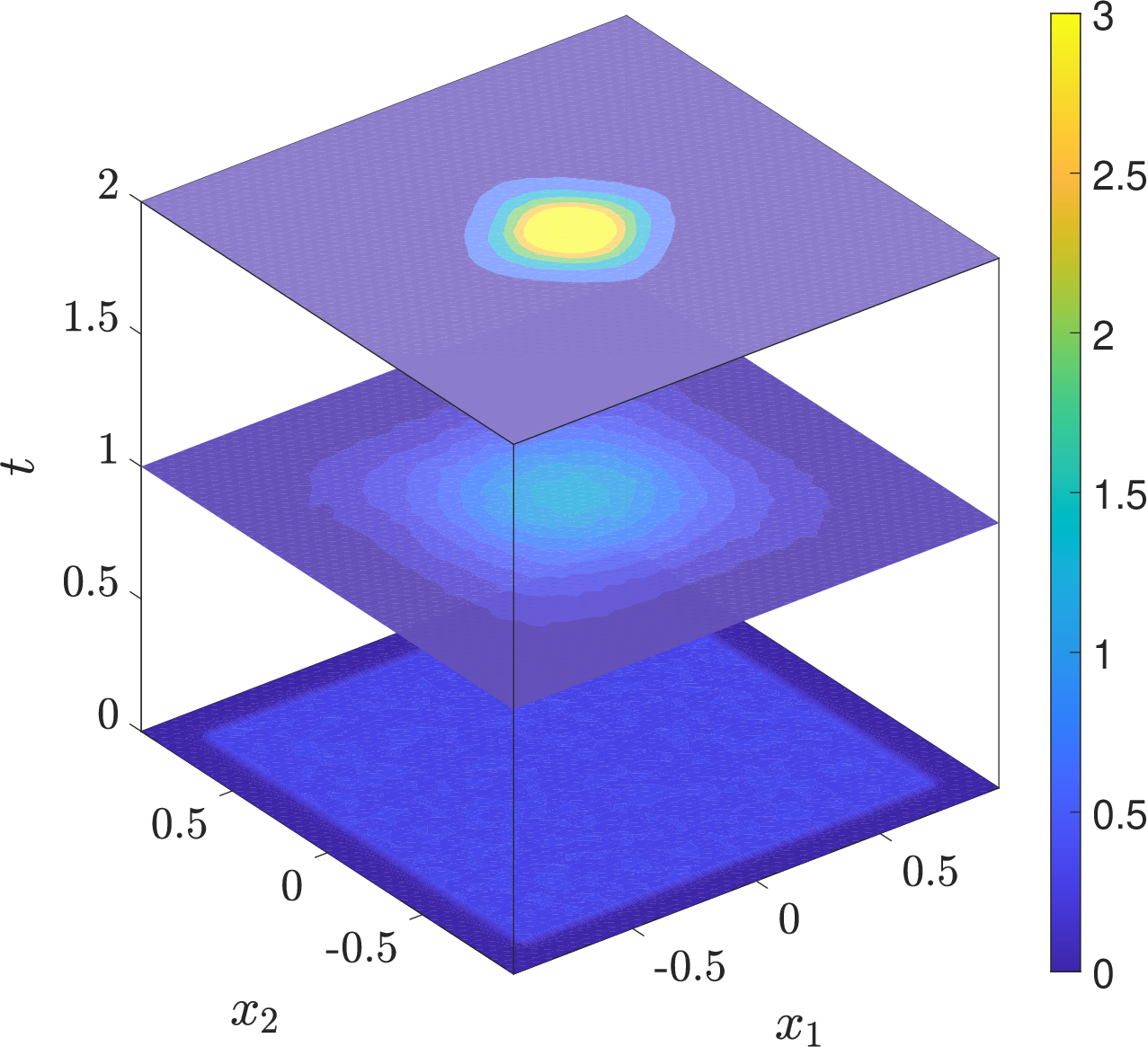}\quad\quad\includegraphics[width=0.33\linewidth]{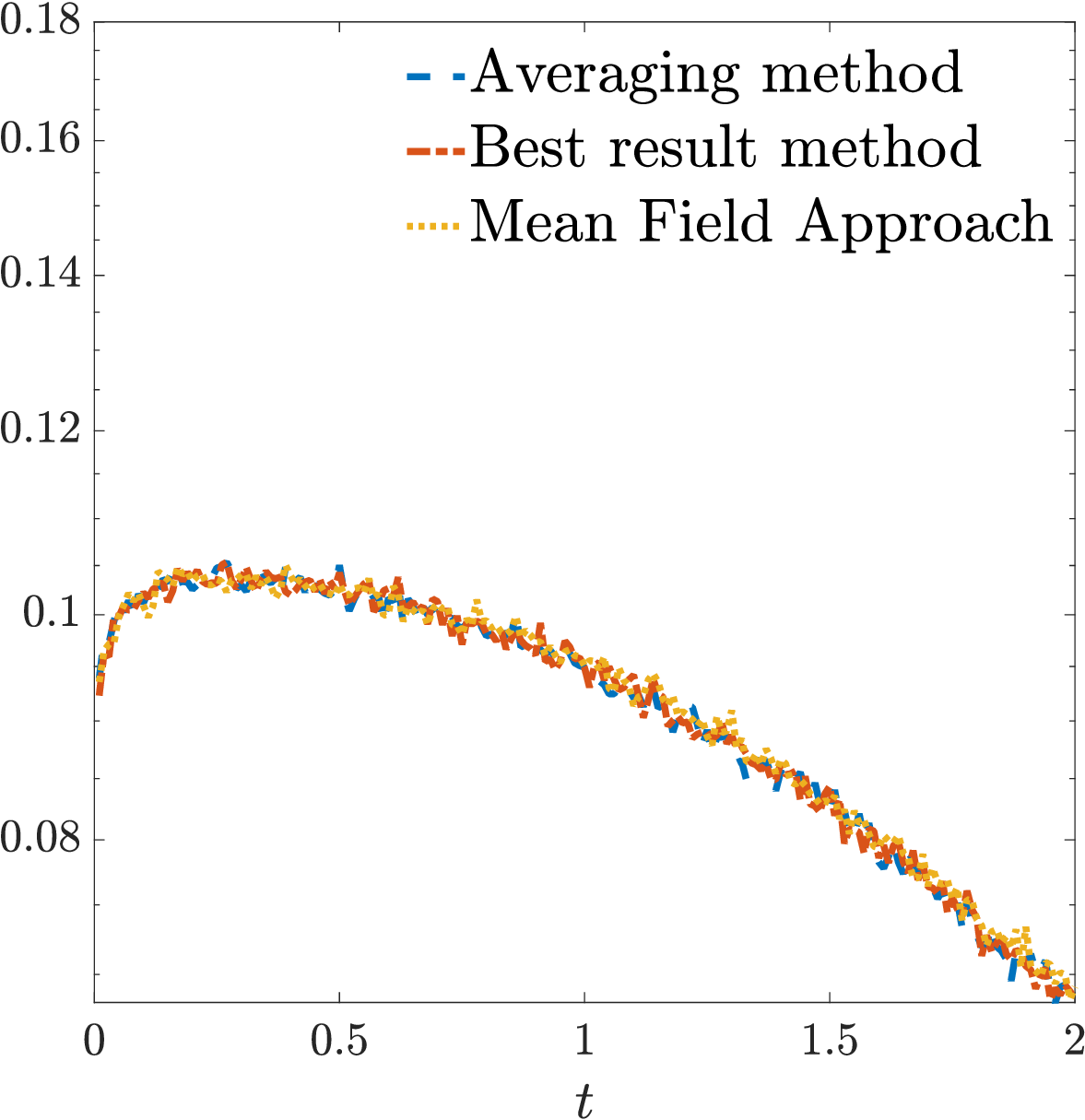}
\caption{Test 5. Evolution of the reconstructed density (left) and trend of the normalized $L^1$ errors over time for the different reconstruction strategies (right), setting $\mathscr{S}=3$.}
\label{fig:2dcase_test6_2}
\end{figure}

\paragraph{Test 5: non-local radial diffusion in the two-dimensional setting} This last simulation is in a two-dimensional setting, with a bounded confidence attractive radial kernel given by $P(r) = \chi(r <1)$ and a nonlocal diffusion $D(x,y) = \tilde D(x,y)(y-x)$, with $\tilde D(x,y) = \tilde D(\lvert x - y \rvert) = \tilde D(r) = (1+r^2)^{-2}$. With this choice of interaction and diffusion functions the dynamics is attractive, so starting from a uniform distribution of the agents we witness their concentration around $(0,0)$ over time. The objective is to reconstruct both $P$ and $\tilde D$, as in Test 3. The available dataset consists in the collection of two opinions of $N = 10^5$ agents, with $201$ available data frames for each of the two opinions, and a time step $\Delta t = 0.01$ between two consecutive time frames. Unfortunately, with the hardware at our disposal, we were unable to perform simulations using the averaging approach in the case $\ell=1$, $M_P=M_D=100$ and $\ell=2$, $M_P=M_D=50$.  We report the qualitative results for the case $\ell = 4$ and $M_P = 25$ in Figures \ref{fig:2dcase_test6} and \ref{fig:2dcase_test6_2}. For the optimization problem, we set $\kappa_P =-1$ and $P_0=1$ in \eqref{eq:AP}, whereas in \eqref{eq:AD} we set $\kappa_D=-1$ and $\overline{\zeta}_0=1$.

\section{Conclusions}
\label{sec:conclusions}
In this work, we proposed a data-driven framework for the identification of interaction kernels in stochastic multi-agent systems from trajectory data, addressing the challenging setting in which the pairwise interactions between agents are not directly observable. Starting from a microscopic binary-interaction model, we formulated the reconstruction of both drift and diffusion terms as regression problems on suitable finite-dimensional approximation spaces. 
The lack of direct information on the interaction structure is handled through two complementary strategies. The first approach combines binary interactions with random batch sampling, providing an efficient approximation of the microscopic dynamics.  The second approach relies on a mean-field description of the system, where the interaction kernels are inferred from the empirical particle density. 
In addition, we derived an a-priori error estimate for the reconstructed trajectories, showing stability of the learned dynamics and consistency of the reconstruction under suitable assumptions. Numerical experiments in one and two spatial dimensions confirm the effectiveness of the proposed methodology. 
In particular, both approaches yield reconstructions with comparable accuracy, even when the interaction structure is only partially observed. 
This highlights the robustness of the framework with respect to incomplete information on the underlying interactions. 
Future research directions include the validation of the proposed methodology on real-world datasets, such as biological or social collective dynamics, as well as the use of neural-network-based representations for the interaction kernels, which may provide a more flexible approximation framework when the interaction laws depend on higher-dimensional variables or exhibit complex structures that are difficult to capture with fixed basis expansions.

\section*{Acknowledgments}
The authors GA, AA, EC are members of the INdAM-GNCS activity group. The authors have developed this work within the activities of the project “Data driven discovery and control of multi-scale interacting artificial agent systems” (code
P2022JC95T), funded by the European Union - NextGenerationEU, National Recovery
and Resilience Plan (PNRR) - Mission 4 component 2, investment 1.1 “Fondo per il
Programma Nazionale di Ricerca e Progetti di Rilevante Interesse Nazionale” (PRIN). Elisa Calzola acknowledges the support by the project titled ``Tackling complexity: advanced numerical approaches for multiscale systems with uncertainties - ADAMUS'' funded by the Ministry of Universities and Research in accordance with Ministerial Decree No. 1236 of August 1, 2023 - Fondo Italiano per la Scienza (Italian Science Fund) – FIS2 Call, CUP F53C25000210001, project code FIS-2023-01334, and by the GNCS 2026 project ``Metodi numerici e data-driven per dinamiche socio-economiche ed epidemiologiche'', CUP E53C25002010001.

\appendix
\section{Proof of Theorem \ref{thm:1}}\label{app:conti}
Let us first observe that $\delta_P,\,\delta_D$, $P_{\max},\widehat{P}_{\max}$ and $D_{\max},\widehat{D}_{\max}$ are all finite positive constants due to the fact that $\mathbf{P},\,\mathbf{D},\,\widehat{\mathbf{P}}$ and $\widehat{\mathbf{D}}$ are all continuous functions defined on a compact subset of $\mathbb{R}_+$.
Moreover, in what follows we will make use of the elementary bounds 
\begin{subequations}\label{eq:lemma}
\begin{align}
& \|S-S'\|_2\le 2,\qquad S,S'\qquad\textrm{permuation matrices}\label{eq:lemma_1}
\\
&  \| a\odot b \|_2^2 \leq \|a\|_{\infty}^2\|b\|^2_2,\qquad a,b\in\mathbb{R}^{N_d},\label{eq:lemma_2}
\end{align}
\end{subequations}

\paragraph{Estimate for $\bar e^n$}
Subtracting \eqref{eq:xbar} from \eqref{eq:og_dynamics}, we obtain
\[
\bar e^{n+1}=\bar e^n+\Delta t\,A^n+\sqrt{\Delta t}\,B^n\odot \Xi^n,
\]
where
\begin{align}
A^n &:= {\mathbf P}(R^n) \odot \bigl( \mathsf{S}^n  X^n -  X^n \bigr) - \widehat{\mathbf{P}}(\bar R^n) \odot \bigl( \mathsf{S}^n  \bar X^n - \bar X^n \bigr),
\label{eq:A}\\
B^n &:= \mathbf{D}(R^n) - \widehat{\mathbf{D}}(\bar R^n).\label{eq:B}
\end{align}
Therefore, we have    
\begin{equation}\label{eq:conto1_ga}\begin{aligned}
\mathbb{E}\left[\|\bar e^{n+1}\|^2_2\right]
=&\mathbb{E}\left[\| \bar e^n \|^2_2\right]+ \Delta t^2\mathbb{E}\left[\| A^n\|^2_2\right] + \Delta t\mathbb{E}\left[\|B^n\odot \Xi^n\|_2^2\right]+\\
&\qquad2\Delta t \mathbb{E}\left[\langle \bar e^n,A^n\rangle\right] + 2 \Delta t^{3/2}\mathbb{E}\left[\langle \bar e^n + \Delta t A^n, B^n\odot \Xi^n\rangle \right],
\end{aligned}
\end{equation}
where the last term vanishes since $\Xi^n$ is centered.
We first estimate the drift contribution. Adding and subtracting ${\mathbf P}(R^n)\odot (\mathsf S^n\bar X^n - \bar X^n)$, we write $A^n$ as
\begin{align}
A^n  = 
{\mathbf P}(R^n) \odot \bigl(( \mathsf{S}^n - \mathsf{I}_{N_d}) \bar e^n\bigl)      + \bigl(\mathbf{P}(R^n) - \widehat{\mathbf{P}}(\bar R^n)\bigr) \odot \bigl( \mathsf{S}^n  \bar X^n - \bar X^n \bigr).
\end{align}
Hence, by using \eqref{eq:lemma_1}  and the bounds \eqref{eq:bounds} we have
\begin{align}
\| A^n\|^2_2 &\leq2 {P}^2_{\textrm{max}}\|( \mathsf{S}^n - \mathsf{I}_{N_d}) \bar e^n\|_2^2+2\|\bigl(\mathbf{P}(R^n) - \widehat{\mathbf{P}}(\bar R^n)\bigr)\|_{\infty}^2\| \bigl( \mathsf{S}^n  \bar X^n - \bar X^n \bigr)\|^2_2
\cr
&\qquad\leq8{P}^2_{\textrm{max}}\|\bar e^n\|_2^2
+8L^2\|\bigl(\mathbf{P}(R^n) - \widehat{\mathbf{P}}(\bar R^n)\bigr)\|_{\infty}^2\cr
&\qquad\quad\leq 8{P}^2_{\textrm{max}}\|\bar e^n\|_2^2
+16L^2N_d\delta^2_P+64L^2L^2_PN_d\|\bar e^n\|^2_2,
\end{align}
with $\mathsf{I}_{N_d}$ the $N_d\times N_d$ identity matrix, and where in the last inequality we added and subtracted
$\widehat{\mathbf{P}}(R^n)$  and used \eqref{eq:bounds}, along with Lipschitz bound for $P$.
For the third term in \eqref{eq:conto1_ga}, we use \eqref{eq:lemma} and observe that $\|\Xi^n\|_2^2 \sim \chi^2_{N_d}$, which implies that $\mathbb{E}\!\left[\|\Xi^n\|_2^2\right] = N_d$. Then we have
\begin{equation}\label{eq:Bdotxi}
\begin{aligned}
\mathbb{E}\left[\|B^n\odot \Xi^n\|^2_2\right]  &\leq\mathbb{E}\left[\| \Xi^n\|_2^2\right]\|(\mathbf{D}(R^n) - \widehat{\mathbf{D}}(\bar R^n))\|^2_\infty
\cr
&\qquad\leq 2N_d\left(4L_{D}^2\|\bar e^n\|_2^2 + \delta_D^2 \right),
\end{aligned}
\end{equation}
where the last inequality is obtained by summing and subtracting ${\mathbf{D}}(\bar R^n)$, and recalling the usual bounds.
Finally, we estimate the fourth term on the right-hand side of \eqref{eq:conto1_ga}. Using the definition of $A^n$ in \eqref{eq:A}, we decompose the inner product $2\langle \bar e^n, A^n \rangle$ into two terms. The first contribution is estimated as follows
\[
\begin{aligned}
2\langle \bar e^n,\mathbf{P}(R^n) \odot \bigl( (\mathsf{S}^n - \mathsf{I}_{N_d})\bar e^n\bigr)\rangle 
\leq  2\|\bar e^n\|_2 \| (\mathsf{S}^n - \mathsf{I}_{N_d})\bar e^n\|_2\leq 4 P_{\max}\|\bar e^n\|_2^2,
\end{aligned}
\]
where the last inequality follows from Cauchy-Schwarz.
The second contribution can be bounded as follows, again using Cauchy-Schwarz and Young's inequalities
\begin{align}
&2\langle \bar e^n, \bigl( \mathbf{P}(R^n) - \mathbf{\widehat{P}}(\bar R^n)\bigr) \odot \bigl((\mathsf{S}^n - \mathsf{I}_{N_d})\bar X^n\bigl) \rangle 
\cr
&\quad\leq2\| \bar e^n\|_2 \| \bigl( \mathbf{P}(R^n) - \mathbf{\widehat{P}}(\bar R^n)\bigr)\|_2 \| (\mathsf{S}^n - \mathsf{I}_{N_d})\bar X^n\|_\infty\cr
&\qquad \leq 4L\|\bar e^n \|_2\left(\delta_P\sqrt{N_d} + 2L_{P}\|\bar e^n\|_2\right)
\leq 4L^2\delta_P^2N_d + \left(1+4LL_P\right)\| \bar e^n\|_2^2
\end{align}
Then, from \eqref{eq:conto1_ga}, setting \(
\bar E^2_{n}:=\mathbb{E}[\|\bar e^n\|^2_2]
\), and combining the previous estimates, we have
\[
\begin{aligned}
\bar E^2_{n+1}
\leq&\bar E^2_n
+8\Delta t^2\left[\bigl(P_{\max}^2+8L^2L_P^2\bigr)\bar E^2_n+2L^2N_d\delta_P^2\right]
+\\&\Delta t\left[(1+4LL_P + 4P_{\max}+8N_dL^2_{D})\bar E^2_n + 2L^2N_d\delta_P^2  + 2N_d\delta_D^2\right],
\end{aligned}
\]
Furthermore, since $\Delta t\le1$, $\Delta t^2$ terms can be absorbed into the $\Delta t$ contribution, as follows
\[
\bar E^2_{n+1}
\le (1+C_1\Delta t)\bar E^2_n
+ C_2(\delta_P^2+\delta_D^2)\Delta t,
\]
where
$C_1 := 8P_{\max}^2+64L^2L_P^2+4P_{\max}+4LL_P+8N_dL_D^2+1$, and
$C_2 := 18L^2N_d+2N_d.$
Iterating this process, since we assumed $\bar E^2_0=0$, it follows that
\begin{align}\label{eq:Ebar}
\bar E^2_n
\le \frac{C_2}{C_1}(\delta_P^2+\delta_D^2)
\Bigl((1+C_1\Delta t)^n-1\Bigr)
\le
\frac{C_2}{C_1}(\delta_P^2+\delta_D^2)
\left(e^{C_1T}-1\right).
\end{align}
\paragraph{ Estimate for $\widetilde{ e}^n$}
We consider now the error term given by
\[
\widetilde{e}^{n+1 }=  \widetilde{e}^n+\Delta t\, \widetilde{A}^n+\sqrt{\Delta t}\, \widetilde{B}^n\odot \Xi^n,
\]
where now
\begin{align*}
\widetilde{A}^n &:= \widehat{{\mathbf P}}(\bar R^n) \odot \bigl( {\mathsf{S}}^n  \bar X^n -  \bar X^n \bigr) - \widehat{\mathbf{P}}(\widetilde{R}^n) \odot \bigl( \widehat{\mathsf{S}}^n  \widetilde{X}^n - \widetilde{X}^n\bigr) \\
\widetilde{B}^n &:= \widehat{\mathbf{D}}(\bar R^n)\odot \Xi^n - \widehat{\mathbf{D}}(\widetilde{R}^n)\odot \widehat{\Xi}^n.
\end{align*}
Analogously to \eqref{eq:conto1_ga} we can expand the norm $\|\widetilde{e}^{n+1}\|^2_2$ and control the different terms. Following the previous computation, for $\|\widetilde{A}^n\|^2_2$ we have 
\begin{equation}
\begin{aligned}\|\widetilde{A}^n\|_2^2 \leq &2\|\bigl(\widehat{\mathbf{P}}(\bar R^n) - \widehat{\mathbf{P}}(\widetilde R^n)\bigr) \odot \bigl(\mathsf{S}^n \bar X^n - \bar X^n \bigr)\|^2_2 
\cr&\qquad+ 2\|\widehat{\mathbf{P}}(\widetilde{R}^n)\odot \bigl(\mathsf{S}^n \bar X^n - \widehat{\mathsf{S}}^n \widetilde X^n - \widetilde{e}^n\bigr)\|_2^2,
\end{aligned}
\end{equation}
where, for the first term, using Lipschitz continuity of $\widehat{\mathbf{P}}$, the bounds \eqref{eq:bounds} and \eqref{eq:lemma_2} we have
\begin{align*}
&2\|\bigl(\widehat{\mathbf{P}}(\bar R^n) - \widehat{\mathbf{P}}(\widetilde R^n)\bigr) \odot \bigl(\mathsf{S}^n \bar X^n - \bar X^n \bigr)\|^2_2 \leq 2\|\widehat{\mathbf{P}}(\bar R^n) - \widehat{\mathbf{P}}(\widetilde R^n)\|_{\infty}^2\| \bigl(\mathsf{S}^n - I_{N_d}\bigr)X^n \|_2^2\cr
&\qquad\leq 8L^2N_dL^2_{\widehat{P}}\|(\mathsf{S}^n -\widehat{\mathsf{S}}^n)\bar{X}^n+(\widehat{\mathsf{S}}^n-I_{N_d})\widetilde{e}^n)\|_2^2\leq 16L^2L^2_{\widehat{P}}N_d(\eta_S^2N_dL^2+4\|\widetilde{e}^n\|^2_2).
\end{align*}
The estimate for the second term follows similarly retaining the bound on $\widehat{P}$ as follows
\[
2\|\widehat{\mathbf{P}}(\widetilde{R}^n)\odot \bigl(\mathsf{S}^n \bar X^n - \widehat{\mathsf{S}}^n \widetilde X^n - \widetilde{e}^n\bigr)\|_2^2\leq 4\widehat{P}_{\textrm{max}}^2\left(\eta^2_SN_dL^2+4\|\widetilde{e}^n\|_2^2\right)
\]
For the term $\widetilde{B}^n$, proceeding as in \eqref{eq:Bdotxi} and using the Lipschitz continuity of $\widehat{D}$ together with the bounds \eqref{eq:bounds}, we obtain
\begin{equation}
\begin{aligned}
\mathbb{E}\left[\|\widetilde{B}^n\|^2_2 \right] \leq& \mathbb{E}\left[\|\Xi^n\|_2^2\right]\|\bigl(\widehat{\mathbf{D}}(\bar R^n) -\widehat{\mathbf{D}}(\widetilde{R}^n) \bigr)\|_\infty^2
\cr& 
\quad\leq2N_dL^2_{\widehat{D}}(\eta_S^2N_dL^2+4\|\widetilde{e}^n\|^2_2)
\end{aligned}
\end{equation}
Finally, the mixed term appearing in the expansion of the error $\|\widetilde{e}^{n+1}\|^2_2$ reads
\begin{equation}\label{eq:termineagg2}
\begin{aligned}
2\langle \widetilde{e}^n, \widetilde{A}^n\rangle= &2\langle \widetilde{e}^n,\bigl(\widehat{\mathbf{P}}(\bar R^n) - \widehat{\mathbf{P}}(\widetilde R^n)\bigr) \odot \bigl(\mathsf{S}^n \bar X^n - \bar X^n \bigr)\rangle +\cr
&\qquad2\langle\widetilde{e}^n,\widehat{\mathbf{P}}(\widetilde{R}^n)\odot \bigl(\mathsf{S}^n \bar X^n - \widehat{\mathsf{S}}^n \widetilde X^n - \widetilde{e}^n\bigr)\rangle.
\end{aligned}
\end{equation}
Here, using \eqref{eq:lemma_2} and Cauchy-Schwarz in the first term of the right hand side we obtain 
\begin{align*}
&2\langle \widetilde{e}^n,\bigl(\widehat{\mathbf{P}}(\bar R^n) - \widehat{\mathbf{P}}(\widetilde R^n)\bigr) \odot \bigl(\mathsf{S}^n \bar X^n - \bar X^n \bigr)\rangle 
\cr &\quad \leq2\| \widetilde{e}^n\|_2 \|\widehat{\mathbf{P}}(\bar R^n) - \widehat{\mathbf{P}}(\widetilde R^n)\|_2\| \mathsf{S}^n \bar X^n - \bar X^n \|_\infty\leq4LL_{\widehat{P}}\|\widetilde{e}^n\|_2(\eta_SL\sqrt{N_d}+2\|\widetilde{e}^n\|_2),
\end{align*}
and similarly for the second term
\begin{align*}
& 2\langle\widetilde{e}^n,\widehat{\mathbf{P}}(\widetilde{R}^n)\odot \bigl(\mathsf{S}^n \bar X^n - \widehat{\mathsf{S}}^n \widetilde X^n - \widetilde{e}^n\bigr)\rangle
\cr
&\quad
\leq
2\|\widetilde{e}^n\|_2 \|\widehat{\mathbf{P}}(\widetilde{R}^n)\|_\infty\|\bigl(\mathsf{S}^n \bar X^n - \widehat{\mathsf{S}}^n \widetilde X^n - \widetilde{e}^n\bigr)\|_2\leq2\widehat{P}_{\max}\|\widetilde{e}^n\|_2(\eta_SL\sqrt{N_d}+2\|\widetilde{e}^n\|_2).
\end{align*}
Then combining this two estimate and using Young's inequality we have
\begin{align*}
2\langle \widetilde{e}^n, \widetilde{A}^n\rangle
&\leq 4(\widehat{P}_{\max}+2LL_{\widehat{P}})\|\widetilde{e}^n\|^2_2 + 2(\widehat{P}_{\max}+2LL_{\widehat{P}})\eta_S\sqrt{N_d}L\|\widetilde{e}^n\|_2
\cr&\quad\leq(1+4(\widehat{P}_{\max}+2LL_{\widehat{P}}))\|\widetilde{e}^n\|^2_2 + (\widehat{P}_{\max}+2LL_{\widehat{P}})^2L^2N_d\eta_S^2.
\end{align*}
Setting $\widetilde{E}^2_n := \mathbb{E}\!\left[ \|\widetilde{e}^n\|_2^2\right]$, and combining the previous estimates, we obtain a recursive bound for $\widetilde{E}^2_n$ as
\begin{equation*}
\widetilde{E}^2_{n+1} \leq \bigl(1 + \Delta t C_3\bigr)\widetilde{E}^2_{n}  + \Delta t C_4\eta^2_S, 
\end{equation*}
where the constants are $C_3 := 1+4\widehat{P}_{\max}+2LL_{\widehat{P}}+8N_dL^2_{\widehat{D}}+8\widehat{P}^2_{\max}+64L^2L^2_{\widehat{P}}N_d$ and $C_4 := (\widehat{P}_{\max}+2LL_{\widehat{P}})^2N_d+2N_d^2L_{\widehat{D}}^2L^2+4\widehat{P}_{\max}^2N_dL^2+16L^4L^2_{\widehat{P}}N^2_d$.
Here we used the assumption $\Delta t \le 1$, which implies $\Delta t^2 \le \Delta t$; consequently, all terms of order $\Delta t^2$ can be absorbed into the constants multiplying $\Delta t$. Eventually,
recalling that $\widetilde{E}^2_{0} = 0$ this leads to 
\begin{equation}\label{eq:stima2}
\widetilde{E}^2_{n+1}\leq \frac{C_4\eta_S^2}{C_3}\left((1+\Delta t C_3)^n - 1\right) \leq \frac{C_4\eta_S^2}{C_3}\left(e^{C_3T}-1\right)
\end{equation}
Combining \eqref{eq:Ebar} and \eqref{eq:stima2} we finally get
\[
\mathbb{E}\left[\|e^{n+1}\|_2^2 \right]\leq \widehat{C}_2(\delta_P^2+\delta_D^2+\eta_S^2)\left(\exp\left\{\widehat{C}_1T \right\} - 1\right),
\]
for all $n = 0, \dots, M-1$, with 
$\widehat{C_2} = \max\left\{C_2/{C_1},{C_4}/{C_3}\right\}$ and $\widehat{C}_1 = \max\left\{C_1,\,C_3\right\}$.

\section{Proof of Remark \ref{stima_inversa}}\label{app:conti2}
As in our a priori estimate (see Theorem \ref{thm:1}), we assume $X^n\in[-L,L]^{dN}$, so that all pairwise distances lie in $[0,R]$ with $R:=2L\sqrt d$, and that $\lvert P\rvert\le P_{\max}$ and $\lvert D\rvert\le D_{\max}$ on $[0,R]$. The basis $\{\phi_k\}_{k=1}^{N_b^P}$ are piecewise linear hat functions on a mesh of $[0,R]$ of width $h$. 
We denote by $\bar\varrho$ the nodal values of $P$,
by $P_h=\sum_k\bar\varrho_k\phi_k$ its piecewise
linear interpolant, and by
$\varepsilon_h:=\|P-P_h\|_{L^\infty(0,R)}$
the interpolation error. For sufficiently smooth
kernels
one has $\varepsilon_h=\mathcal{O}(h^2)$.
Then the observed increment in the data satisfies
\begin{equation}\label{eq:data}
X^{n+1}-X^n=\Delta t\,\mathbf{\Theta}^n\bar\varrho+\Delta t\,\varepsilon^n+\sqrt{\Delta t}\,\mathbf{D}^n\odot\Xi^n,
\quad \|\varepsilon^n\|_2\le 2L\varepsilon_h\sqrt{dN}, \quad n=1,\ldots,M_P,
\end{equation}
where $\mathbf{\Theta}^n$ is built with the true pairing $\mathsf S^n$ and $\varepsilon^n$ is the modeling error.
For the random-batch case, assume that, conditionally on $X^n$, the $N_p$ auxiliary pairing matrices at time $n$ are mutually independent, each distributed as the actual pairing, and independent of the actual pairing and of the noise $\Xi^n$. Let us introduce the notation $\hat{\mathbf{\Theta}}$ for the matrices in the regressions, with  $\hat{ \mathbf{\Theta}}^n=\mathbf{\Theta}^n$ for \eqref{eq:regression_kern} and $\hat{\mathbf{\Theta}}^n=\mathbf{\Theta}^n_{RB}=\frac{1}{N_p}\sum_{q=1}^{N_p}\Theta^{(n,q)}$ for \eqref{eq:regression_kern_batch}. Moreover, we define the stacked matrices
\[
\hat{\mathbf{\Theta}}
=
\begin{bmatrix}
\hat{\mathbf{\Theta}}^{1}\\
\vdots\\
\hat{\mathbf{\Theta}}^{M_P}
\end{bmatrix}\in\mathbb{R}^{dNM_P\times N_b^P}.
\]
Let $\varrho^*$ be the solution of \eqref{eq:regression_kern_batch}
and define the matrix
\[
\qquad
A:=\frac{1}{M_P\,dN}\sum_{n=1}^{M_P}(\hat{\mathbf{\Theta}}^n)^\top\hat{\mathbf{\Theta}}^n=\frac{\hat{\mathbf{\Theta}}^\top\hat{\mathbf{\Theta}}}{M_P\,dN}.
\]
and assume that its smaller eigenvalue is positive and $\lambda_{\min}(A) \geq c > 0$. From the normal equations and \eqref{eq:data} we get 
\begin{equation}\label{eq:err}
M_PdN\,A\,(\varrho^*-\bar\varrho)=\sum_{n=1}^{M_P}(\hat{\mathbf{\Theta}}^n)^\top\Bigl[(\mathbf{\Theta}^n-\hat{\mathbf{\Theta}}^n)\bar\varrho+\varepsilon^n+\Delta t^{-1/2}\,\mathbf{D}^n\odot\Xi^n\Bigr].
\end{equation}
The least-squares problem has a unique solution 
\begin{equation}\label{eq:lsq_sol}
\varrho^*-\bar{\varrho}
=
\frac{1}{M_PdN}
A^{-1}
\sum_{n=1}^{M_P}
(\hat{\mathbf{\Theta}}^n)^\top
\Bigl[(\mathbf{\Theta}^n-\hat{\mathbf{\Theta}}^n)\bar\varrho+\varepsilon^n+\Delta t^{-1/2}\,\mathbf{D}^n\odot\Xi^n\Bigr].
\end{equation}
Consequently, using the triangle inequality and
$\|A^{-1}\|_2\leq c^{-1}$, we obtain
\begin{equation}\label{eq:normal}
\begin{aligned}
\|\varrho^*-\bar\varrho\|_2
\leq \frac{1}{cM_PdN}
\Bigg(
&\left\|
\sum_{n=1}^{M_P}
(\hat{\mathbf{\Theta}}^n)^\top\varepsilon^n
\right\|_2+
\left\|
\sum_{n=1}^{M_P}
(\hat{\mathbf{\Theta}}^n)^\top
(\mathbf{\Theta}^n-\hat{\mathbf{\Theta}}^n)
\bar\varrho
\right\|_2\\
&+
\frac{1}{\sqrt{\Delta t}}
\left\|
\sum_{n=1}^{M_P}
(\hat{\mathbf{\Theta}}^n)^\top
(\mathbf D^n\odot\Xi^n)
\right\|_2
\Bigg).
\end{aligned}
\end{equation}

Let us now estimate each term of the right-hand side. Each row of $\mathbf{\Theta}^n$, $\mathbf{\Theta}^n_{\rm av}$ and $\hat{\mathbf{\Theta}}^n$ is an average of vectors $\bigl(\phi_k(\lvert x_j-x_i\rvert)(x_j-x_i)_c\bigr)_{k}$, with a single partner $j$ for $\mathbf{\Theta}^n$, all possible partners for $\mathbf{\Theta}^n_{\rm av}$, and $N_p$ sampled partners for $\mathbf{\Theta}^n_{RB}$. Since $\lvert(x_j-x_i)_c\rvert\le2L$, $\phi_k\ge0$ and $\sum_k\phi_k\le1$, each such vector has $\ell^1$-norm at most $2L$, and hence so does each row. Since $\|\bar\varrho\|_\infty\le P_{\max}$, it follows that
\begin{equation}\label{eq:stima1}
\|\hat{\mathbf{\Theta}}^n\|_F^2\le4L^2dN, \qquad \lvert(\mathbf{\Theta}^n\bar\varrho)_{(i,c)}\rvert,\ \lvert(\mathbf{\Theta}^n_{\rm av}\bar\varrho)_{(i,c)}\rvert,\ \lvert(\hat{\mathbf{\Theta}}^n\bar\varrho)_{(i,c)}\rvert\le2LP_{\max}.
\end{equation}
The interpolation term in \eqref{eq:normal} is bounded deterministically by
\begin{equation}\label{eq:errore_interp}
\frac{1}{M_PdN}\left\|\sum_{n=1}^{M_P}(\hat{\mathbf{\Theta}}^n)^\top \varepsilon^n\right\|_2
\le4L^2\varepsilon_h.
\end{equation}
For known pairs the second term of the right-hand side of \eqref{eq:normal} vanishes. For random batching, instead, we introduce the conditional mean design matrix
$\mathbf{\Theta}_{\rm av}^n:=\mathbb{E}[\mathbf{\Theta}^n\mid X^n]$ and use the following decomposition
\[
(\mathbf{\Theta}^n-\hat{\mathbf{\Theta}}^n)\bar\varrho
=(\mathbf{\Theta}^n-\mathbf{\Theta}_{\rm av}^n)\bar\varrho
+(\mathbf{\Theta}_{\rm av}^n-\hat{\mathbf{\Theta}}^n)\bar\varrho.
\]
By the triangle inequality, the second term in \eqref{eq:normal} is bounded by
\[
\left\|\sum_{n=1}^{M_P}(\hat{\mathbf{\Theta}}^n)^\top(\mathbf{\Theta}^n-\mathbf{\Theta}_{\rm av}^n)\bar\varrho\right\|_2 + \left\|\sum_{n=1}^{M_P}(\hat{\mathbf{\Theta}}^n)^\top(\mathbf{\Theta}_{\rm av}^n-\hat{\mathbf{\Theta}}^n)\bar\varrho\right\|_2,
\]
where the first is a sum of martingale differences, which we bound in expectation, while we bound the second by the triangle inequality over $n$.
Starting from the second term, by triangle inequality, independence of the auxiliary batches, and \eqref{eq:stima1},
\[
\mathbb{E}\|(\mathbf{\Theta}_{\rm av}^n-\hat{\mathbf{\Theta}}^n)\bar\varrho\|_2^2
\le \frac{4L^2P_{\max}^2dN}{N_p}.
\]
Therefore, using Cauchy--Schwarz,
\begin{equation}\label{eq:errore_theta1}
\frac1{M_PdN}\mathbb{E}\lVert\sum_{n=1}^{M_P}(\hat{\mathbf{\Theta}}^n)^\top
(\mathbf{\Theta}_{\rm av}^n-\hat{\mathbf{\Theta}}^n)\bar\varrho\rVert_2
\le\frac{4L^2P_{\max}}{\sqrt{N_p}}.
\end{equation}
Next, let us focus on the vectors
$(\hat{\mathbf{\Theta}}^n)^\top(\mathbf{\Theta}^n-\mathbf{\Theta}_{\rm av}^n)\bar\varrho$, that are martingale differences with respect to the natural filtration including the observations and the auxiliary batches at each step. By \eqref{eq:stima1},
\[
\lVert (\hat{\mathbf{\Theta}}^n)^\top(\mathbf{\Theta}^n-\mathbf{\Theta}_{\rm av}^n)\bar\varrho\rVert_2\le 8L^2P_{\max}dN.
\]
Using Cauchy--Scwharz inequality and the fact that 
\[
\mathbb{E}\Bigl[\bigl((\hat{\mathbf{\Theta}}^n)^\top(\mathbf{\Theta}^n-\mathbf{\Theta}_{\rm av}^n)\bar\varrho\bigr)^{\!\top}\bigl((\hat{\mathbf{\Theta}}^m)^\top(\mathbf{\Theta}^m-\mathbf{\Theta}_{\rm av}^m)\bar\varrho\bigr)\Bigr]=0,
\]
for $m\neq n$, we get 
\begin{equation}\label{eq:pairingerror}
\begin{aligned}
\mathbb E\left\|\sum_{n=1}^{M_P}( (\hat{\mathbf{\Theta}}^n)^\top(\mathbf{\Theta}^n-\mathbf{\Theta}_{\rm av}^n)\bar\varrho)\right\|_2
&\le
\left(\mathbb E\left\|\sum_{n=1}^{M_P}( (\hat{\mathbf{\Theta}}^n)^\top(\mathbf{\Theta}^n-\mathbf{\Theta}_{\rm av}^n)\bar\varrho)\right\|_2^2\right)^{1/2} \\
&= \left(
\sum_{n=1}^{M_P}\mathbb E\|(\hat{\mathbf{\Theta}}^n)^\top(\mathbf{\Theta}^n-\mathbf{\Theta}_{\rm av}^n)\bar\varrho\|_2^2\right)^{1/2},
\end{aligned}
\end{equation}
leading to
\begin{equation}\label{eq:errore_theta2}
\frac1{M_PdN}\mathbb E\left\|\sum_{n=1}^{M_P}( (\hat{\mathbf{\Theta}}^n)^\top(\mathbf{\Theta}^n-\mathbf{\Theta}_{\rm av}^n)\bar\varrho)\right\|_2
\le\frac{8L^2P_{\max}}{\sqrt{M_P}}.
\end{equation}
Finally, also the noise vectors
$(\hat{\mathbf{\Theta}}^n)^\top(\mathbf D^n\odot\Xi^n)$
are martingale differences and their second moments satisfy
\[
\mathbb{E}\left\|(\hat{\mathbf{\Theta}}^n)^\top(\mathbf D^n\odot\Xi^n)\right\|_2^2
\le D_{\max}^2\mathbb{E}\left\|\hat{\mathbf{\Theta}}^n\right\|_F^2
\le 4L^2D_{\max}^2dN.
\]
Using again the orthogonality of martingale differences we get
\begin{equation}\label{eq:errore_noise}
\frac1{M_PdN\sqrt{\Delta t}}\mathbb{E}\left\|\sum_{n=1}^{M_P}(\hat{\mathbf{\Theta}}^n)^\top(\mathbf D^n\odot\Xi^n)\right\|_2
\le\frac{2LD_{\max}}{\sqrt{dNM_P\Delta t}}.
\end{equation}
\medskip
Taking expectations in \eqref{eq:normal} and using \eqref{eq:errore_interp}, \eqref{eq:errore_theta1}, \eqref{eq:errore_theta2}, and \eqref{eq:errore_noise} we get the desired estimate \eqref{eq:mainbound} which is recalled below for the reader convenience:
\begin{equation*}
\mathbb{E}\left[\left\|\varrho^*-\bar\varrho\right\|_2\right]
\le \frac1c\left(
4L^2\varepsilon_h
+\left(\frac{4L^2P_{\max}}{\sqrt{N_p}}
+\frac{8L^2P_{\max}}{\sqrt{M_P}}\right)\chi_{\rm RB}
+\frac{2LD_{\max}}{\sqrt{dNM_P\Delta t}}
\right),
\end{equation*}
with $\chi_{\rm RB}$ equal to $0$ for known pairs, and equal to $1$ for the random-batch approach. 
Recall that
\[
P_h(r)=\sum_{k=1}^{N_b^P}\bar\varrho_k\phi_k(r),
\qquad
\widehat P(r)=\sum_{k=1}^{N_b^P}\varrho_k^*\phi_k(r),
\]
then subtracting, we obtain
\[
P_h(r)-\widehat P(r) =\sum_{k=1}^{N_b^P}(\bar\varrho_k-\varrho_k^*)\phi_k(r).
\]
Taking absolute values and using the triangle inequality,
\[
|P_h(r)-\widehat P(r)|\le\sum_{k=1}^{N_b^P}
|\bar\varrho_k-\varrho_k^*|\,|\phi_k(r)|\le\|\bar\varrho-\varrho^*\|_\infty
\sum_{k=1}^{N_b^P}|\phi_k(r)|.
\]
Now using nonnegativity of the basis functions and the fact that they are a partition of the unity, we get 
\[
|P_h(r)-\widehat P(r)|\le\|\bar\varrho-\varrho^*\|_\infty.
\]
Taking the supremum over \(r\in[0,R]\),
\[
\|P_h-\widehat P\|_{L^\infty(0,R)}\le\|\bar\varrho-\varrho^*\|_\infty\le\|\bar\varrho-\varrho^*\|_2,
\]
that leads to
\[
\|P-\widehat P\|_{L^\infty(0,R)} = \left\|(P-P_h)+(P_h-\widehat P)\right\|_{L^\infty(0,R)}\le\varepsilon_h+\|\bar\varrho-\varrho^*\|_2.
\]

\bibliographystyle{plain}
\bibliography{sample}
\end{document}